\pdfoutput=1
\documentclass{article}
\usepackage{iclr2027_conference,times}
\iclrpreprintcopy
\usepackage{amsmath,amssymb,amsfonts,amsthm}
\usepackage{booktabs}
\usepackage{array}
\usepackage{graphicx}
\usepackage{subcaption}
\usepackage{flafter}
\usepackage{url}
\usepackage{hyperref}
\usepackage{microtype}
\usepackage{enumitem}

\hypersetup{
    colorlinks=true,
    linkcolor=blue!70!black,
    citecolor=green!50!black,
    urlcolor=blue!80!black
}

\setlist{leftmargin=*,itemsep=0pt,topsep=1pt,parsep=0pt,partopsep=0pt}

\newcommand{\cFourStrictDD}{0.95\% \pm 0.31\%}

\newcommand{\cOneTangentialDD}{5.44\% \pm 2.03\%}
\newcommand{\cOneRadialDD}{37.36\% \pm 2.53\%}

\newcommand{\sFiveDonorDTwoFiveSixCausalSteps}{6{,}791.7 \pm 489.2}
\newcommand{\sFiveDonorDOneTwoEightCausalSteps}{9{,}120.8 \pm 779.5}
\newcommand{\sFiveDonorDOneTwoEightBidirSteps}{32{,}870.8 \pm 7{,}329.8}

\theoremstyle{plain}
\newtheorem{proposition}{Proposition}
\theoremstyle{definition}
\newtheorem{definition}{Definition}

\title{\raggedright Beyond Embedding Transfer: Architectural Interfaces, Iso-Parameter Scaling, and Preconditioner-Governed Consolidation in Grokking}

\author{Zeyu Jia$^{1,2}$ \\
$^{1}$School of Biomedical Engineering and Technology, Tianjin Medical University, Tianjin, China \\
$^{2}$Medical School, Tianjin University, Tianjin, China \\
\texttt{jiazeyu@tju.edu.cn}
}

\begin{document}

\maketitle

\begin{abstract}
Grokking---delayed generalization after memorization---incurs substantial training steps across algebraic tasks. Cross-operator transfer from a grokked donor accelerates target acquisition, yet post-grokking AdamW continuation triggers test-accuracy relapse as unannealed updates drift across residual interfaces and decaying second moments amplify late gradients. Across depths one to three on modular arithmetic ($\mathbb{Z}_{113}$), non-Abelian $S_5$ conjugation, and real-sensory MNIST modular arithmetic, we separate representational transfer from preconditioner-governed consolidation across thirteen experimental batteries. First, co-transferring internal blocks accelerates one-layer and two-layer modular takeoff beyond boundary interfaces, while graded prefix pinning bounds depth-three drift under unannealed continuation. Second, an iso-parameter $S_5$ scaling grid separates attention head dimension from MLP width: widening two-head attention to dimension sixty-four enables $98.54\%$ cold accuracy at $427\text{k}$ parameters and $85.95\%$ boundary-only transfer, while final-slot representations refine conjugation directly across blocks without an intermediate binary-product bottleneck. Third, Plateau-Gated Leaky Consolidation pairs validation-envelope step damping with leaky second-moment decay, reaching $84.85\%$ zero-leakage validation-gated risk-constrained accuracy across five pre-saturation two-layer $S_5$ suites while cutting step budget relative to hard anchoring. Fourth, transferring a grokked visual trunk and blocks to image-disjoint MNIST targets reaches $92.27\%$ and $90.76\%$ accuracy, prunes odd centroid harmonics, and achieves $90.42\%$ risk-constrained accuracy across $60/60$ branches without hyperparameter retuning.
\end{abstract}

\section{Introduction}
\label{sec:intro}

\begin{figure*}[t]
    \centering
    \includegraphics[width=0.72\linewidth]{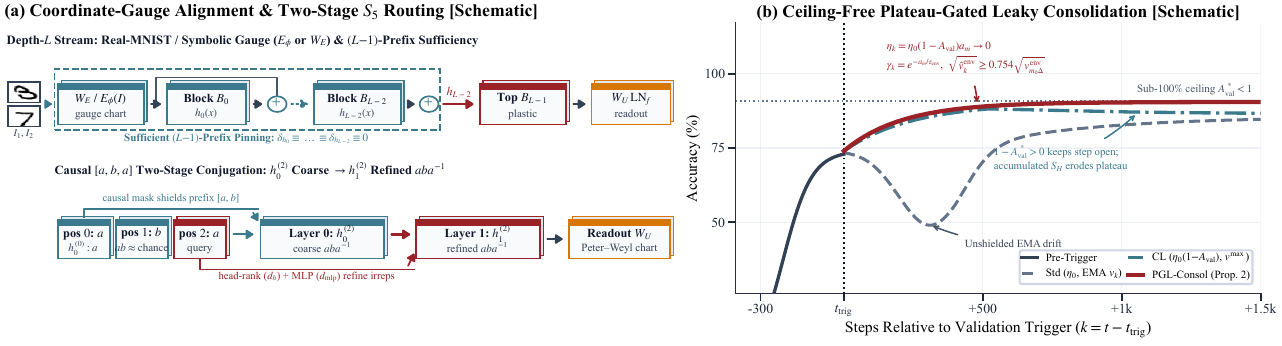}
    \caption{\textbf{Architectural Interfaces and Preconditioner-Governed Consolidation (Schematic).} (a) Depth-$L$ residual stream, $S_5$ causal $[a,b,a]$ routing, and visual trunk $E_\phi$: $(L-1)$-prefix pinning guarantees $\delta_{h_\ell} \equiv 0$ (Definition~\ref{def:interface_drift}), while slot $2$ refines $aba^{-1}$ across blocks. (b) PGL-Consol pairs plateau-gated damping with leaky second-moment decay $v_{k+1,i}^{\rm pgl}$ (Propositions~\ref{prop:precond_path}--\ref{prop:pgl_bounds}).}
    \label{fig:fig0_overview}
\end{figure*}

When trained on algorithmic operations with weight decay, overparameterized networks exhibit \emph{grokking}: training loss converges rapidly via memorization, whereas held-out generalization emerges thousands of steps later \citep{Power2022_220102177,Liu2022_220510343,Liu2022_221001117}. During this transition, structured Fourier or group-representation circuits emerge across embeddings, attention heads, and MLPs \citep{Nanda2023_230105217,Gromov2023_230102679,Chughtai2023_230203025,Morwani2024_231107568}. Recently, \citet{Xu2025_GrokTransfer} showed that transferring pre-trained input embeddings accelerates modular grokking (cf.\ \citealp{Lee2024_240520233}). Yet focusing on symbolic input embeddings leaves four questions open (Figure~\ref{fig:fig0_overview}). Do internal blocks ($B$) accelerate acquisition beyond boundary carriers ($E, U$), and how do depth ($L \ge 2$) and head-versus-MLP width govern non-Abelian composition? Moreover, how do optimizer second moments govern post-grokking consolidation, and do these mechanisms extend to perceptual inputs \citep{Humayun2024_240215555}?

Across thirteen batteries ($>2{,}800$ trajectories; Table~\ref{tab:cohort_provenance}), we establish four findings.
\begin{enumerate}[leftmargin=1.4em,itemsep=0pt,topsep=1pt]
\item \textbf{Component Synergy and Depth-$L$ Prefix Sufficiency.} Under cold Frobenius norm matching, co-transferring internal blocks $B$ accelerates 1L $\mathbb{Z}_{113}$ takeoff by $+5.46$ pp ($-616.7$ steps, $p_{\rm Holm}=0.0039$) and 2L takeoff by $+704.7$ units ($8.1\times$ faster onset) over $E{+}U$, whereas omitting donor $E$ collapses 2L transfer ($-4{,}475.6$ units). Under unannealed $\eta_0=10^{-3}$ continuation, $(L-1)$-prefix pinning suffices for $\delta_{h_\ell} \equiv 0$ (Definition~\ref{def:interface_drift}), cutting 3L drawdown from $75.99\%$ (Std) and $31.26\%$ (Frz $E{+}U$) to $0.32\%$ (Pin $B_0^*$) and $0.04\%$ (Pin $B_{0:1}^*$, $-0.28 \pm 0.19$ pp).
\item \textbf{Plateau-Gated Leaky Consolidation and Zero-Leakage Risk Control.} Guided by coordinate-wise preconditioner bounds (Propositions~\ref{prop:precond_path}--\ref{prop:pgl_bounds}), \textbf{PGL-Consol} pairs validation-loss progress gating ($a_m \in [0,1]$, $\rho_c = e^{-\Delta/\tau_c}$) with leaky second-moment decay ($\gamma_k = e^{-a_m/\tau_{\rm env}}$, plateau-tail floor $\ge 0.7539$). Under zero-leakage validation-gated $\mathrm{ValRC}(1\%)$, PGL-Consol reaches $84.85\% \pm 0.89\%$ across five pre-saturation 2L $S_5$ suites ($58/58$ feasible; $+2.15 \pm 0.51$ pp over Closed-Loop, $+6.56 \pm 0.88$ pp over Matched OL, $p_{{\rm Holm},8}=0.0039$) and $78.46\% \pm 1.10\%$ on 1L $S_5$ ($-79.5\%$ step budget).
\item \textbf{Iso-Parameter $S_5$ Capacity Asymmetry and Direct Final-Slot Refinement.} On $S_5$ conjugation ($aba^{-1}$), bidirectional $[a,b]$ resolves cycle type at $99.38\%$ with partial intra-class disambiguation ($33.90\% \pm 6.21\%$ vs.\ the $5.83\%$ class-uniform floor). Across our $427\text{k}$--$1.64\text{M}$ grid, head dimension and MLP width act as scratch substitutes ($-49.14 \pm 9.56$ pp, $p_{{\rm Holm},18}=0.0088$). Reconfiguring $4{\times}32 \to 2{\times}64$ at $427\text{k}$ lifts Cold to $98.54\% \pm 0.70\%$ and $E{+}U$ to $85.95\% \pm 5.67\%$ ($+60.93 \pm 9.98$ pp over $U{+}B$), while warm Full benefits from wider MLPs ($95.02\% \pm 1.57\%$ at $953\text{k}$). Ridge-calibrated Peter--Weyl probes across 15 suites show near-chance $a\circ b$ at slot $1$ ($0.48\%$--$6.64\%$) and two-stage $aba^{-1}$ refinement at slot $2$ ($68.98\%$--$91.92\% \to 80.94\%$--$97.50\%$).
\item \textbf{Real-Sensory MNIST Transfer, CRT Pruning, and Zero-Retuning Consolidation}. Across five image-disjoint MNIST pools, transferring $E_\phi{+}B{+}U$ from $y_1+y_2\bmod 10$ lifts $2y_1+y_2\bmod 10$ takeoff by $+62.9 \pm 6.6$ $I_{\rm 1k}$ over $E_{\rm only}$ ($92.27\% \pm 0.31\%$). On $y_1-y_2\bmod 10$, Full reaches $90.76\% \pm 0.32\%$ ($+1.99 \pm 0.40$ pp over $E{+}U$, $9.0\times$ faster onset). Unsupervised centroid DFT spectra concentrate $94.58\%$--$96.30\%$ energy into Chinese Remainder Theorem modes $\{k=2,4,5\}$, and zero-retuning PGL-Consol reaches $90.42\% \pm 0.13\%$ $\mathrm{ValRC}(1\%)$ across $60/60$ branches ($+1.67 \pm 0.15$ pp over Closed-Loop).
\end{enumerate}

\section{Problem Formulation \& Zero-Leakage Evaluation Protocol}
\label{sec:protocol}

\subsection{Task Families, Architectures, and Norm-Matched Transfer}
Our study separates three \textbf{core load-bearing mechanisms}---component transfer with $(L-1)$-prefix pinning, causal $[a,b,a]$ routing across head-versus-MLP scaling, and second-moment-governed consolidation---from two \textbf{auxiliary engineering guardrails}: cold Frobenius norm matching (Eq.~\eqref{eq:scale_matching}) and two-step validation gating (Eq.~\eqref{eq:online_trigger_dd}). Hyperparameters and controller constants were locked on $\mathcal{D}_{\rm val}$ on disjoint development seeds: \texttt{500}--\texttt{502} for $S_5$ routing, \texttt{370}--\texttt{375} and \texttt{500}--\texttt{502} for PGL-Consol $(\Delta=50, \tau_c=200, \tau_{\rm env}=400, d_{\rm ref}=0.01)$, and \texttt{400}--\texttt{401} for MNIST.

We evaluate three domains (Table~\ref{tab:cohort_provenance}). First, on \textbf{Abelian $\mathbb{Z}_{113}$}, $a+b \pmod{113}$ donors transfer to $2a+b$, $3a+b$, $a^2+b^2$, and $a-b$ over $12{,}769$ pairs ($20/10/70\%$ splits; $30/10/60\%$ for 2L/3L donors) across pre-LN Transformers \citep{Power2022_220102177,Nanda2023_230105217} at $L\in\{1,2,3\}$ ($228\text{k}$--$623\text{k}$ params). Second, on \textbf{non-Abelian $S_5$} ($|S_5|=120$), $a\circ b$ donors transfer to $aba^{-1}$ and $ab^{-1}$ over $14{,}400$ pairs ($30/20/50\%$ and $45/15/40\%$). Architectures span a $229\text{k}$ 1L baseline and a 2L grid at $427\text{k}$ ($4{\times}32, 2{\times}64, d_{\rm mlp}=512$) and $1.64\text{M}$ ($8{\times}32, d_{\rm mlp}=1{,}024$). An intermediate $953\text{k}$ tier pairs $953{,}344$ parameters on $4{\times}32$ and $2{\times}64$ ($d_{\rm model}=128, d_{\rm mlp}=1{,}536$) with $953{,}028$ on $8{\times}32$ ($d_{\rm model}=256, d_{\rm mlp}=354$). Third, on \textbf{real-sensory MNIST}, a $523{,}136$-parameter Conv2d-trunk and 2L Transformer ($E_\phi, B, U$) transfers from $y_1+y_2\bmod 10$ to $\mathrm{Op1}:\ 2y_1+y_2\bmod 10$ and $\mathrm{Op2}:\ y_1-y_2\bmod 10$ without digit labels. Five image-disjoint pools comprise $15{,}000$ donor-train images ($40{,}000$ pairs), $2{,}000$ donor-val ($5{,}000$ pairs), and $1{,}800$ target-train ($3{,}500$ pairs). Held-out evaluation uses $3{,}000$ target-val ($5{,}000$ pairs) and $10{,}000$ test images ($10{,}000$ pairs). Given $h_{\rm in} = W_E + W_{\rm pos}$ or $[E_\phi(I_1), E_\phi(I_2)] + W_{\rm pos}$, block $B_\ell$ updates $h_\ell = h_{\ell-1} + \mathcal{F}_{B_\ell}$ with readout $W_U\,\mathrm{LN}_f(h_{L-1,-1})$. To separate weight direction from norm inflation \citep{Liu2022_221001117}, donor transfer rescales each tensor $\theta$ to its paired cold norm $n_c = \|\theta^{\rm cold}\|_F$.
\begin{equation}
\label{eq:scale_matching}
\theta^{\rm transfer} = \tfrac{\theta^{\rm donor}}{\|\theta^{\rm donor}\|_F + 10^{-12}} \cdot \|\theta^{\rm cold}\|_F.
\end{equation}
Symbolic runs use AdamW \citep{Loshchilov2019_AdamW} ($\eta_0=10^{-3}, \lambda=1.0, \beta_1=0.9, \beta_2=0.98$, batch $512$ on $\mathbb{Z}_{113}$, $256$ on $S_5$); MNIST uses $\eta_0=1.5\times 10^{-3}, \lambda=0.8$, batch $256$ in pure FP32.

\subsection{Zero-Leakage Dual-Track Metrics, Intent-to-Treat, and Family-Wise Inference}
Throughout our study, $\mathcal{D}_{\rm val}$ governs online gating, checkpoint selection, and risk fallback, while $\mathcal{D}_{\rm test}$ serves as the \textbf{confirmatory} evaluation channel. Test accuracy $A_{\rm test}(t)$ never informs donor qualification, triggers, step sizes, or fallback; the sole retrospective diagnostic is the trough dissection in \S\ref{sec:mechanism} and Appendix~\ref{app:probe_details}. Acquisition is measured by early test integral $I_{\rm 5k}$ ($I_{\rm 1k}$ on MNIST; $I_{\rm full}$ over $30\text{k}$ steps on 2L $S_5$) and validation-selected test accuracy at $\arg\max_t A_{\rm val}(t)$. Latency uses censored validation $\widehat{\mathrm{RMST}}_\tau$ ($A_{\rm val} \ge 0.90$ across $21$ checks at $\Delta=50$, $\tau=30{,}000$; step ratios reflect $\mathrm{RMST}$ or trigger steps rather than wall-clock FLOPs). Consolidation forks from $t_{\rm trig}$ ($k=2$ checks with $A_{\rm val} \ge q_{\rm trig}$) over $H \in \{1{,}500, 2{,}000, 3{,}000\}$. Zero-leakage \textbf{validation-gated risk-constrained accuracy} $\mathrm{ValRC}(\tau)$ reverts to $\theta_{t_{\rm trig}}$ solely when validation drawdown $D_H^{\rm val} > \tau$.
\begin{align}
\label{eq:efficacy_rmst}
I_T &= \textstyle\int_0^T A_{\rm test}(t)\,dt, \qquad
\widehat{\mathrm{RMST}}_\tau = \tfrac{1}{N} \textstyle\sum_{b=1}^N \min(T_b, \tau), \\
\label{eq:online_trigger_dd}
t_{\rm trig} &= \min\bigl\{ t : A_{\rm val}(t{-}\Delta), A_{\rm val}(t) \ge q_{\rm trig} \bigr\}, \quad
D_H^{\rm ch} = \max_{t_{\rm trig} \le s \le t \le t_{\rm trig}+H} \bigl(A_{\rm ch}(s) - A_{\rm ch}(t)\bigr), \\
\label{eq:val_rc}
\mathrm{ValRC}(\tau) &= A_{\rm test}(t_{\rm trig}{+}H)\,\mathbb{I}\bigl[D_H^{\rm val} \le \tau\bigr] + A_{\rm test}(t_{\rm trig})\,\mathbb{I}\bigl[D_H^{\rm val} > \tau\bigr], \quad \mathrm{ch} \in \{\mathrm{val}, \mathrm{test}\}.
\end{align}
For partially triggered suites ($9/12$ on 2L $S_5$ $q_{\rm trig}=0.95$; $8/12$ on 1L $S_5$ $q_{\rm trig}=0.70$), we report triggered-subset metrics alongside full-cohort $12/12$ Intent-to-Treat ($\mathrm{ITT}$) trajectories without zero-imputation (Tables~\ref{tab:protection_battery},~\ref{tab:app_pgl_valrc}). Contrasts across paired seeds \citep{Hagmann2023_230204054} use two-sided exact sign-flip permutation tests ($p_{\rm perm}^{\min} = 2^{-N+1}$: $0.0005$ at $N=12$, $0.0020$ at $N=10$, $0.0078$ at $N=8$, $0.0313$ at $N=6$; $\text{ddof}=1$ SEMs) and paired TOST \citep{Schuirmann1987} ($\pm 500$ steps; $\pm 100/\pm 250$ $I_{\rm 5k}$ units). Multiplicity uses Holm--Bonferroni families \citep{Holm1979} ($M=8$ in Table~\ref{tab:phase_b}; $M=6\text{--}9$ in Tables~\ref{tab:protection_battery},~\ref{tab:app_dissection_1l}--\ref{tab:app_pgl_valrc}; $M\in\{6, 18\}$ in Table~\ref{tab:depth_s5_acquisition}; $M\in\{5, 3, 8\}$ in Table~\ref{tab:mnist_main}).

\begin{figure*}[t]
    \centering
    \includegraphics[width=0.68\linewidth]{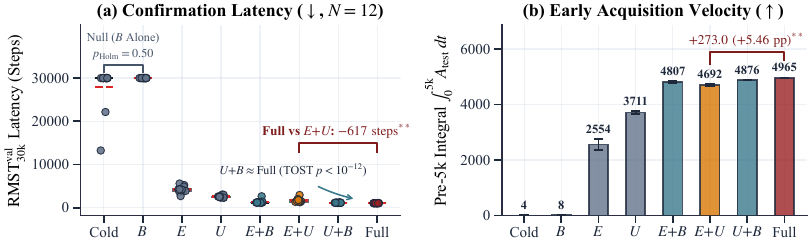}
    \caption{\textbf{Component Localization and Factorial Synergy ($N=12$ Paired Seeds, 108 Runs).} (a) $\widehat{\mathrm{RMST}}_{30\mathrm{k}}^{\rm val}$: Full cuts confirmation steps by $616.7$ vs.\ $E{+}U$ ($p_{\rm Holm} = 0.0039$), while $U{+}B$ lies within $\pm 500$ steps ($p_{\rm TOST} = 2.91\times 10^{-13}$ val; $0.0011$ test). (b) Early integral $I_{\rm 5k}$: Full adds $+273.0$ units ($+5.46$ pp) over $E{+}U$ ($p_{\rm Holm} = 0.0039$).}
    \label{fig:fig3_synergy}
\end{figure*}

\section{Carrier Localization and Trough Attribution}
\label{sec:carrier}
\label{sec:mechanism}

\subsection{Factorial Synergy and 1L vs.\ 2L Omitted-\texorpdfstring{$E$}{E} Asymmetry}
Across 108 runs on 1L $a+b \to 2a+b \pmod{113}$ ($N=12$ paired seeds; Table~\ref{tab:phase_b}, Figure~\ref{fig:fig3_synergy}), Full transfer raises $I_{\rm 5k}$ by $+273.0 \pm 28.4$ units ($\mathbf{+5.46}$ \textbf{pp}, $p_{\rm Holm} = 0.0039$) over $E{+}U$. It also cuts $\widehat{\mathrm{RMST}}_{30\mathrm{k}}^{\rm val}$ by $\mathbf{616.7 \pm 129.0}$ \textbf{steps} ($1{,}058.3 \pm 5.6$ vs.\ $1{,}675.0 \pm 133.5$, $p_{\rm Holm} = 0.0039$). Relative to $\tau=30{,}000$-censored Cold ($27{,}950.0 \pm 1{,}486.7$ steps, $2/12$ confirmed), recipient confirmation uses $26.4\times$ fewer steps, and combined Donor${+}$Full ($6{,}795.8 \pm 275.8$) cuts total steps $4.1\times$. Readout $U$ outperforms $E$, $B$ alone confirms $0/12$, omitting Frobenius norm matching on $E$ adds $+13{,}812.5 \pm 1{,}823.6$ steps ($p_{\rm Holm}=0.0039$), and two-way interactions are positive ($I_{EB} = +2{,}249.5 \pm 207.7$, $I_{UB} = +1{,}161.8 \pm 49.5$). At $L=1$, $U{+}B$ passes coarse $\pm 500$-step $\mathrm{RMST}$ ($+100.0 \pm 10.7$) and $\pm 250$-unit $I_{\rm 5k}$ equivalence ($-88.9 \pm 7.4$, $p_{\rm TOST} = 1.14\times 10^{-10}$), missing the fine $\pm 100$-unit bound ($p_{\rm TOST}=0.0813$). At $L=2$, omitting donor $E$ under $U{+}B$ reduces $I_{\rm 5k}$ by $\mathbf{-4{,}475.6 \pm 76.8}$ units ($-89.51\text{ pp}$, $p_{\rm perm}=0.0005$; $p_{\rm TOST}=1.00$), showing that two-layer residual routing depends on aligned input embeddings.

\begin{table*}[t]
\centering
\scriptsize
\setlength{\tabcolsep}{2.8pt}
\renewcommand{\arraystretch}{0.76}
\caption{\textbf{Factorial Component-Transfer Battery ($p=113$, $N=12$ paired seeds, 108 runs).}
\textbf{Panel A}: Confirmation count, $\widehat{\mathrm{RMST}}_{30\mathrm{k}}^{\rm val/test}$, $I_{\rm 5k}$, early accuracy, and paired $\Delta \mathrm{RMST}_{\rm val}$ vs.\ Full ($\text{Mean} \pm \text{SEM}, \text{ddof}=1$; donors $5{,}737.5 \pm 274.1$, Donor${+}$Full $6{,}795.8 \pm 275.8$; unnormalized $E$: $\mathrm{RMST}_{\rm val} = 18{,}033.3 \pm 1{,}927.0$, $I_{\rm 5k} = 15.36 \pm 5.68$).
\textbf{Panel B}: Sign-flip $p_{\rm perm}$ ($2^{12}=4{,}096$), Holm $p_{\rm Holm}$ ($M=8$), and TOST $p_{\rm TOST}$.}
\label{tab:phase_b}
\begin{tabular*}{\textwidth}{@{\extracolsep{\fill}}lcccccc@{}}
\toprule
\multicolumn{7}{@{}l@{}}{\textbf{Panel A: Factorial Condition Summary ($N=12$ paired seeds: 96 factorial runs plus 12 scale controls)}} \\
\midrule
\textbf{Condition} & \textbf{Conf.} & \textbf{Val RMST} & \textbf{Test RMST} & \textbf{$I_{\rm 5k}$ ($\uparrow$)} & \textbf{Acc\%} & \textbf{$\Delta\mathrm{RMST}_{\rm val}$ [CI]} \\
\midrule
Cold & 2/12 & $27{,}950.0{\pm}1{,}486.7$ & $27{,}950.0{\pm}1{,}486.7$ & $3.9{\pm}0.2$ & $0.08$ & $+26{,}891.7$ $[21308,28929]$ \\
$B$ & 0/12 & $30{,}000.0{\pm}0.0$ & $30{,}000.0{\pm}0.0$ & $7.5{\pm}0.2$ & $0.15$ & $+28{,}941.7$ $[28921,28946]$ \\
$E$ & 12/12 & $4{,}220.8{\pm}219.5$ & $4{,}220.8{\pm}219.5$ & $2{,}553.7{\pm}204.0$ & $51.07$ & $+3{,}162.5$ $[2758,3575]$ \\
$U$ & 12/12 & $2{,}595.8{\pm}67.3$ & $2{,}600.0{\pm}66.0$ & $3{,}710.6{\pm}55.7$ & $74.21$ & $+1{,}537.5$ $[1423,1688]$ \\
$E{+}B$ & 12/12 & $1{,}354.2{\pm}119.7$ & $1{,}354.2{\pm}119.7$ & $4{,}806.9{\pm}32.6$ & $96.14$ & $+295.8$ $[171,752]$ \\
$E{+}U$ & 12/12 & $1{,}675.0{\pm}133.5$ & $1{,}733.3{\pm}153.7$ & $4{,}691.8{\pm}27.4$ & $93.84$ & $+616.7$ $[446,996]$ \\
$U{+}B$ & 12/12 & $1{,}158.3{\pm}8.3$ & $\mathbf{1{,}158.3{\pm}8.3}$ & $4{,}876.0{\pm}8.4$ & $97.52$ & $+100.0$ $[83,121]$ \\
Full & \textbf{12/12} & $\mathbf{1{,}058.3{\pm}5.6}$ & $1{,}175.0{\pm}120.5$ & $\mathbf{4{,}964.8{\pm}2.4}$ & $\mathbf{99.30}$ & Base ($0.0$) \\
\bottomrule
\end{tabular*}
\begin{tabular*}{\textwidth}{@{\extracolsep{\fill}}llcccc@{}}
\multicolumn{6}{@{}l@{}}{\textbf{Panel B: Prospectively Specified Contrasts ($N=12$ paired seeds; exact $p_{\rm perm}$, Holm $p_{\rm Holm}$ with $M=8$, TOST $p_{\rm TOST}$)}} \\
\midrule
\textbf{Contrast} & \textbf{Pair} & \textbf{Metric} & \textbf{Diff $\pm$ SEM} & \textbf{$p_{\rm perm}$} & \textbf{$p_{\rm Holm}$ / $p_{\rm TOST}$} \\
\midrule
1L Efficacy & Full vs $E{+}U$ & $\Delta I_{\rm 5k}$ & $+273.0{\pm}28.4$ & $0.0005$ & $0.0039$ \\
1L Latency & Full vs $E{+}U$ & $\Delta\mathrm{RMST}_{\rm v/t}$ & $+616.7{\pm}129.0$/$+558.3{\pm}177.1$ & $0.0005$/$0.0044$ & $0.0039$/$0.0088$ \\
1L Full & Cold vs Full & $\Delta\mathrm{RMST}_{\rm val}$ & $+26{,}891.7{\pm}1{,}485.7$ & $0.0005$ & $0.0039$ \\
1L $B$ Only & Cold vs $B$ & $\Delta\mathrm{RMST}_{\rm val}$ & $-2{,}050.0{\pm}1{,}486.7$ & $0.5000$ & $0.5000$ \\
1L RMST Eq. & $U{+}B$ vs Full & $\Delta\mathrm{RMST}_{\rm v/t}$ & $+100.0{\pm}10.7$/$-16.7{\pm}121.6$ & --- & $2.91{\times}10^{-13}$/$0.0011$ \\
1L $I_{\rm 5k}$ Eq. & $U{+}B$ vs Full & $\Delta I_{\rm 5k}$ & $-88.9{\pm}7.4$ ($-1.78$ pp) & --- & $0.0813$/$1.14{\times}10^{-10}$ \\
2L No-$E$ & $U{+}B$ vs Full & $\Delta I_{\rm 5k}$ & $-4{,}475.6{\pm}76.8$ ($-89.51$ pp) & $0.0005$ & $1.0000$ \\
\bottomrule
\end{tabular*}
\end{table*}

\subsection{Post-Trigger Relapse Trough Localization}
In our exploratory 12-run scan (Seeds 1--4; Appendix~\ref{app:probe_details}, Figures~\ref{fig:fig2_mechanism}--\ref{fig:fig_probe_events}, Tables~\ref{tab:probe_event_accounting}--\ref{tab:probe_full_results}), three broken Cold runs and one $E{+}U$ run suffer $\ge 10$ pp dips ($0/4$ Full runs dip within $5\text{k}$ steps). Paired $N=12$ state-swap and linear-probe dissections on 1L and 2L $2a+b$ troughs ($H=1{,}500$; Table~\ref{tab:probe_full_results} Panel B) localize post-trigger instability without selection bias. On 2L Full ($97.78\% \pm 0.45\% \to 19.22\% \pm 1.74\%$, tangential block step energy $82.90\% \pm 2.96\%$), restoring trigger $(B_0, B_1)$ recovers $\mathbf{65.54\% \pm 7.01\%}$ ($\mathbf{+46.32 \pm 6.14}$ \textbf{pp}, $12/12$ seeds, $p_{\rm perm}=0.0005$), whereas restoring $E{+}U$ yields $11.93\% \pm 3.13\%$. Refitting an L-BFGS readout on frozen 2L trough features ($N_{\rm calib}=2{,}554$) reaches $50.26\% \pm 6.01\%$ ($-48.82 \pm 6.02$ pp below trigger $99.08\% \pm 0.20\%$), whereas 1L Full troughs show $-8.62 \pm 1.33$ pp deficit ($90.90\% \pm 1.33\%$) and $94.80\% \pm 0.55\%$ early $B$-swap recovery, confirming internal block drift.

\begin{table*}[t]
\centering
\scriptsize
\setlength{\tabcolsep}{2.8pt}
\renewcommand{\arraystretch}{0.68}
\caption{\textbf{Post-Trigger Prefix Pinning and Preconditioner-Governed Consolidation ($L \in \{1,2,3\}$, $\mathbb{Z}_{113}$ and $S_5$, $N=12$ paired seeds).}
$\text{Mean} \pm \text{SEM}$ ($\text{ddof}=1$; sign-flip $p_{\rm perm}$, Holm $p_{\rm Holm}$ over $M=6\text{--}9$, $^{*}p_{\rm Holm}<0.05$, $^{**}p_{\rm Holm}<0.01$; Tables~\ref{tab:app_dissection_1l}--\ref{tab:app_cohort9_3l},~\ref{tab:app_pgl_valrc}). \textbf{Panel A}: $a{+}b \to 2a{+}b$ at $S_H=1.500$ (2L, 3L Std DD: $80.58\%, 75.99\%$; 1L $3a{+}b$ PGL: $0.02\% \pm 0.01\%$ DD). \textbf{Panel B}: 2L $S_5\ q_{\rm trig}=0.70$ macro ($58$ branches; $\mathbb{Z}_{113}$ PGL: $48/48$ feasible, $0.03\%$ DD; 1L $S_5$ PGL: $8/8$ feasible, $78.46\%$ $\mathrm{ValRC}$).}
\label{tab:protection_battery}
\begin{tabular*}{\textwidth}{@{\extracolsep{\fill}}lcccc@{}}
\toprule
\multicolumn{5}{@{}l@{}}{\textbf{Panel A: Graded $(L{-}1)$-Prefix Pinning on $\mathbb{Z}_{113}$ ($a{+}b \to 2a{+}b$, $q_{\rm trig}=0.95, H=1{,}500$, Unannealed $S_H=1.500$)}} \\
\midrule
\textbf{Depth / Policy} & \textbf{$S_H$} & \textbf{Max DD (\%, $\downarrow$)} & \textbf{Min Acc (\%, $\uparrow$)} & \textbf{Term Acc (\%, $\uparrow$)} \\
\midrule
1L Std ($\eta_0{=}10^{-3}, \lambda{=}1$) & $1.500$ & $22.06{\pm}2.40$ & $77.88{\pm}2.42$ & $99.80{\pm}0.13$ \\
1L Frz $E{+}U$ & $1.500$ & $\mathbf{0.60{\pm}0.16}^{*}$ & $\mathbf{98.94{\pm}0.19}$ & $99.36{\pm}0.19$ \\
2L Frz $E{+}U$ & $1.500$ & $65.23{\pm}2.23$ & $34.12{\pm}2.22$ & $98.85{\pm}0.45$ \\
2L Pin $E{+}U{+}B_0^*$ & $1.500$ & $\mathbf{0.95{\pm}0.31}^{*}$ & $\mathbf{96.98{\pm}0.73}$ & $98.76{\pm}0.35$ \\
3L Frz $E{+}U$ & $1.500$ & $31.26{\pm}3.87$ & $68.73{\pm}3.87$ & $96.63{\pm}2.21$ \\
3L Pin $E{+}U{+}B_0^*$ & $1.500$ & $0.32{\pm}0.20$ & $99.63{\pm}0.21$ & $99.92{\pm}0.04$ \\
3L Pin $E{+}U{+}B_{0:1}^*$ & $1.500$ & $\mathbf{0.04{\pm}0.01}^{**}$ & $\mathbf{99.88{\pm}0.05}$ & $\mathbf{99.93{\pm}0.04}$ \\
\bottomrule
\end{tabular*}
\begin{tabular*}{\textwidth}{@{\extracolsep{\fill}}lccccc@{}}
\multicolumn{6}{@{}l@{}}{\textbf{Panel B: Autonomous Consolidation \& Zero-Leakage $\mathrm{ValRC}(1\%)$ on 2L $S_5$ Macro ($q_{\rm trig}=0.70, 58$ branches, $H=3{,}000$)}} \\
\midrule
\textbf{Controller Arm} & \textbf{Feas.} & \textbf{$S_H$} & \textbf{Max DD (\%, $\downarrow$)} & \textbf{Term Acc (\%, $\uparrow$)} & \textbf{$\mathrm{ValRC}(1\%)$ (\%, $\uparrow$)} \\
\midrule
Std ($\eta_0{=}10^{-3}, \lambda{=}1$) & $0/58$ & $3.000{\pm}0.000$ & $20.48{\pm}2.75$ & $84.71{\pm}1.88$ & $73.30{\pm}0.50$ \\
Matched Open-Loop & $16/58$ & $0.257{\pm}0.001$ & $2.75{\pm}0.18$ & $\mathbf{87.14{\pm}0.91}$ & $78.30{\pm}0.75$ \\
Micro-Step ($S_H{=}0.020$) & $\mathbf{58/58}$ & $0.020{\pm}0.000$ & $\mathbf{0.02{\pm}0.01}$ & $79.90{\pm}0.54$ & $79.90{\pm}0.54$ \\
Closed-Loop ($v^{\max}$) & $40/58$ & $0.522{\pm}0.015$ & $1.06{\pm}0.20$ & $85.34{\pm}0.61$ & $82.70{\pm}0.92$ \\
\textbf{PGL-Consol (Ours)} & $\mathbf{58/58}$ & $0.258{\pm}0.016$ & $0.24{\pm}0.02$ & $84.85{\pm}0.89$ & $\mathbf{84.85{\pm}0.89}^{**}$ \\
\bottomrule
\end{tabular*}
\end{table*}

\section{Preconditioner-Governed Consolidation and Prefix Sufficiency}
\label{sec:protection}
\label{sec:online_verification}
\label{sec:cross_operator}

\subsection{From Interface Freezing to Hard Second-Moment Anchoring}
On 1L $a+b \to 2a+b$ ($N=10$; Figure~\ref{fig:fig1_lifecycle}), offline $\{W_E, W_U\}$ freezing from $t=0$ cuts $E{+}U$ running-peak drawdown $D^{\rm max}$ from $20.34\% \pm 2.61\%$ to $0.09\% \pm 0.04\%$ (post-peak dip $\Delta_{\rm relapse}: 19.40\% \pm 2.67\% \to 0.07\% \pm 0.04\%$, $p_{\rm Holm}=0.0059$, differing only by pre-peak dips). Online gating at $q_{\rm trig}=0.95$ cuts 1L $E{+}U$-frozen drawdown to $0.60\%$ ($2a+b$), $5.09\%$ ($3a+b$), and $0.15\%$ ($a^2+b^2$), with norm pinning ($\cOneTangentialDD$) beating direction pinning ($\cOneRadialDD$). Yet $E{+}U$ freezing leaves $65.23\%$ drawdown at $L=2$ and $31.26\%$ at $L=3$ (Table~\ref{tab:protection_battery}), while moment resets, $\lambda=0$, and radial projection ($\perp\text{AdamW}$; \citealp{Prieto2025_250104697}) incur $4.08\%$--$73.97\%$ drawdown ($40.89\% \pm 3.91\%$ vs.\ $86.50\% \pm 3.84\%$ Std terminal accuracy on 2L $S_5$ $q_{\rm trig}=0.95$ macro; Table~\ref{tab:app_pgl_valrc}).

\begin{definition}[Depth-$L$ Prefix-Stream Drift and Strict $(L-1)$-Prefix Pinning]
\label{def:interface_drift}
In a depth-$L$ pre-LN Transformer ($L \ge 2$), stream drift at post-trigger step $k$ on $\mathcal{D}_{\rm val}$ is $\delta_{h_\ell}(k) = \|h_\ell(\mathcal{D}_{\rm val}; t_{\rm trig}+k) - h_\ell(\mathcal{D}_{\rm val}; t_{\rm trig})\|_F / \|h_\ell(\mathcal{D}_{\rm val}; t_{\rm trig})\|_F$ ($0 \le \ell \le L-2$). Freezing the upstream prefix $\{E, W_{\rm pos}, B_0, \dots, B_{L-2}\}$ (including 1D LayerNorm and bias vectors) is sufficient to guarantee $\delta_{h_\ell}(k) \equiv 0$ across $0 \le \ell \le L-2$ (scale/gauge symmetries preclude necessity); \textbf{Strict $(L-1)$-Prefix Pinning} also freezes $\{U, \mathrm{LN}_f\}$.
\end{definition}

Under unannealed $\eta_0=10^{-3}$ continuation at $L=2, 3$ (Figure~\ref{fig:fig4_generality}(a), Tables~\ref{tab:protection_battery},~\ref{tab:app_cohort9_3l}), Pin $E{+}U{+}B_0^*$ cuts 2L drawdown from $65.23\%$ to $\cFourStrictDD$. At $L=3$, drawdown drops from $75.99\% \pm 1.19\%$ (Std) and $31.26\% \pm 3.87\%$ (Frz $E{+}U$) to $0.32\% \pm 0.20\%$ under Pin $E{+}U{+}B_0^*$ ($\delta_{h_1}=0.788 \pm 0.014$) and $0.04\% \pm 0.01\%$ under Pin $E{+}U{+}B_{0:1}^*$ ($-0.28 \pm 0.19$ pp, $p_{\rm Holm}=0.0469$). Confirming sufficiency rather than necessity, Pin $E{+}U{+}B_0^*$ retains $99.92\% \pm 0.04\%$ accuracy despite $\delta_{h_1}=0.788$, while Closed-Loop ($0.01\%$) and PGL-Consol ($0.00\%$) stabilize $L=3$ without freezing.

Separating optimizer preconditioners from Hessian curvature \citep{Reddi2018_190409237,Defazio2024_240515682,Pagliardini2025_240903137,Liang2025_241116085}, we focus on denominator recurrences. Unrolling the decoupled AdamW and AMSGrad recurrences yields elementary coordinate-wise bounds on parameter updates.

\begin{proposition}[Coordinate-Wise Step Bounds under Preserved EMA and Monotonic AMSGrad]
\label{prop:precond_path}
Let decoupled AdamW evolve from $t_{\rm trig} \ge 1$ ($k \ge 0, t_k = t_{\rm trig}+k+1 \ge 2$) with $(m_0, v_0) = (m_{\rm trig}, v_{\rm trig})$, $v_{{\rm trig},i} \ge 0$, $\beta_1, \beta_2 \in (0,1)$, $\epsilon > 0$, and $\lambda \ge 0$.
(i) Under EMA second moments ($\eta_k \ge 0$), every coordinate $i$ obeys $\sqrt{\hat{v}_{t_k,i}} \ge \sqrt{v_{k+1,i}} \ge \beta_2^{(k+1)/2}\sqrt{v_{{\rm trig},i}}$, with strict first inequality if $v_{k+1,i} > 0$ ($0 = 0$ if $v_{k+1,i} = 0$).
(ii) Under AMSGrad $v_{k+1,i}^{\max} = \max(v_{k,i}^{\max}, v_{k+1,i})$ from $v_{0,i}^{\max} = v_{{\rm trig},i}$ and $\eta_k = \eta_0(1 - A_{\rm val}(k))$,
\begin{equation}
\label{eq:prop_precond}
\sqrt{\hat{v}_{t_k,i}^{\max}} = \tfrac{\sqrt{v_{k+1,i}^{\max}}}{\sqrt{1 - \beta_2^{t_k}}} \ge \sqrt{v_{k+1,i}^{\max}} \ge \sqrt{v_{{\rm trig},i}}, \quad
|\Delta \theta_{k,i}| \le \eta_0\bigl(1 - A_{\rm val}(k)\bigr)\Bigl( \tfrac{|\hat{m}_{k+1,i}|}{\sqrt{v_{{\rm trig},i}} + \epsilon} + \lambda |\theta_{k,i}| \Bigr),
\end{equation}
and with $C_{\max} := \max_{0 \le k < H}\bigl(\sum_i (|\hat{m}_{k+1,i}|/(\sqrt{v_{{\rm trig},i}}+\epsilon) + \lambda|\theta_{k,i}|)^2\bigr)^{1/2}$, path length satisfies $\mathcal{P}_H \le C_{\max}\eta_0\sum_{k=0}^{H-1}(1 - A_{\rm val}(k))$.
\end{proposition}

\subsection{Plateau-Gated Leaky Consolidation (PGL-Consol) and the Slingshot Trade-Off}
While Micro-Step ($S_H = 0.02004$ over $201$ steps) halts pre-saturation climb and CL-LR Only suffers $3.0\times$--$5.7\times$ step amplification ($0/8$ feasible on 1L $S_5$), hard Closed-Loop ($v_{k+1,i}^{\max}$) faces two bottlenecks at $q_{\rm trig}=0.70$. First, when $A_{\rm val}(k)$ plateaus at $A_{\inf} < 1$, $\eta_k = \eta_0(1-A_{\rm val}(k))$ accumulates $\mathcal{O}(H(1-A_{\inf}))$ budget ($S_H^{\rm CL} = 0.4682$ on 2L $S_5$ $d=256$; $0.6426$ on 1L $S_5$, $4/8$ feasible). Second, permanent $v_{k,i}^{\max}$ retention locks slingshot gradient spikes \citep{Thilak2022_Slingshot,Lyle2023_230301486,Dohare2024_NaturePlasticity}. \textbf{PGL-Consol} measures validation-loss progress $d_m$ every $\Delta$ steps on $\mathcal{D}_{\rm val}$ to update activity $a_m \in [0,1]$, setting $\eta_k = \eta_0(1 - A_{\rm val}(m))a_m$ and coupling leaky decay $\gamma_k = e^{-a_m/\tau_{\rm env}}$ to $a_m$ so relaxation stops on plateaus (Algorithm 1, Appendix~\ref{app:prop_proof}; locked $\Delta=50, \tau_c=200, \tau_{\rm env}=400, d_{\rm ref}=0.01$).

\begin{proposition}[Bounded Plateau Budget and Leaky Preconditioner Floor of PGL-Consol]
\label{prop:pgl_bounds}
Let PGL-Consol evolve with $\Delta \ge 1$, $\tau_c, \tau_{\rm env}, d_{\rm ref} > 0$, $\rho_c = e^{-\Delta/\tau_c}$, and $\kappa_p = (1 - \rho_c)/d_{\rm ref}$. At checks $k = m\Delta$ from $a_0 = 1, L_0^{\min} = L_{\rm val}(t_{\rm trig})$, let $d_m = \max\bigl(0, \frac{L_{m-1}^{\min} - L_{\rm val}(m)}{L_{m-1}^{\min} + \epsilon_{\rm loss}}\bigr)$, $L_m^{\min} = \min(L_{m-1}^{\min}, L_{\rm val}(m))$, and $a_m = \mathrm{clip}_{[0,1]}(\rho_c a_{m-1} + \kappa_p d_m)$. On $m = \lfloor k/\Delta\rfloor$, set $\eta_k = \eta_0(1 - A_{\rm val}(m)) a_m$, $\gamma_k = e^{-a_m/\tau_{\rm env}}$, and $v_{k+1,i}^{\rm pgl} = \max(\gamma_k v_{k,i}^{\rm pgl}, v_{k+1,i})$ from $v_{0,i}^{\rm pgl} = v_{{\rm trig},i}$. If loss stagnates for $m > m_0$ ($d_m = 0$) with $A_{\inf} := \inf_{m \ge m_0} A_{\rm val}(m)$, then for all $K > k_0 = m_0\Delta$:
(i) $\sum_{k=k_0}^{K-1}\eta_k \le \frac{\eta_0(1 - A_{\inf}) a_{m_0} \Delta}{1 - \rho_c}$ ($\le 0.0497$ at $\eta_0=10^{-3}, \Delta=50, \tau_c=200, A_{\inf}=0.78$).
(ii) $\sqrt{\hat{v}_{t_{K-1},i}^{\rm pgl}} \ge \sqrt{v_{K,i}^{\rm pgl}} \ge \exp\bigl(-\frac{\Delta}{2\tau_{\rm env}}\sum_{m=0}^{m_0-1} a_m\bigr)\exp\bigl(-\frac{\Delta a_{m_0}}{2\tau_{\rm env}(1 - \rho_c)}\bigr)\sqrt{v_{{\rm trig},i}}$ (plateau-tail root factor $\ge 0.7539$ and cumulative floor after $\sum_{m=0}^{m_0-1}a_m = 8$ $\ge 0.4572\sqrt{v_{{\rm trig},i}}$; Appendix~\ref{app:prop_proof}).
\end{proposition}

Across 14 symbolic suites (Tables~\ref{tab:protection_battery},~\ref{tab:app_closed_loop_11suites},~\ref{tab:app_pgl_valrc}), PGL-Consol achieves $0.03\%$ mean DD on saturated $\mathbb{Z}_{113}$ and $99.22\%$ $\mathrm{ValRC}(1\%)$ ($48/48$ feasible). On the 2L $S_5$ $q_{\rm trig}=0.95$ macro, it reaches $98.70\% \pm 0.17\%$ $\mathrm{ValRC}(1\%)$ ($+0.47 \pm 0.07$ pp vs.\ CL; $\mathrm{ITT}_{12}$ DD $2.69\% \pm 0.76\%$ vs.\ $28.20\% \pm 2.96\%$ Std). On the five pre-saturation 2L $S_5$ $q_{\rm trig}=0.70$ suites ($58$ branches), PGL-Consol holds DD to $0.24\% \pm 0.02\%$ ($S_H = 0.258 \pm 0.016$). It reaches $\mathbf{84.85\% \pm 0.89\%}$ $\mathrm{ValRC}(1\%)$ ($58/58$ feasible; $\mathrm{ITT}$ $83.43\% \pm 0.93\%$). It beats Closed-Loop ($\mathbf{+2.15 \pm 0.51}$ \textbf{pp}, $p_{{\rm Holm},8}=0.0039$), Matched OL ($\mathbf{+6.56 \pm 0.88}$ \textbf{pp}, $p_{{\rm Holm},8}=0.0039$), and Micro-Step ($+4.95 \pm 0.52$ pp). It also outperforms PGL Gate + $v^{\max}$ ($+2.39 \pm 0.42$ pp) and Fixed-Leaky ($+5.42 \pm 0.81$ pp). On unconstrained terminal accuracy, anchoring $10^7$--$10^8$ slingshot spikes amid $50$-step $\mathcal{L}_{\rm val}$ fluctuations trades raw climb ($-0.49 \pm 0.39$ pp vs.\ CL, $p_{{\rm Holm},8}=0.6958$; $-2.28 \pm 0.39$ pp vs.\ Matched OL macro; $-6.05$ pp at $1.64\text{M}$). On 1L $S_5$ $q_{\rm trig}=0.70$ ($8/12$ triggered), PGL-Consol cuts $S_H$ by $\mathbf{79.48\%}$ ($0.132 \pm 0.024$ vs.\ $0.643 \pm 0.022$ CL). It reaches $\mathbf{78.46\% \pm 1.10\%}$ $\mathrm{ValRC}(1\%)$, with $8/8$ feasible vs.\ $4/8$ CL at $76.09\% \pm 1.74\%$ ($\mathrm{ITT}_{12}$ $68.91\% \pm 4.92\%$).

\begin{table*}[t]
\centering
\scriptsize
\setlength{\tabcolsep}{2.8pt}
\renewcommand{\arraystretch}{0.68}
\caption{\textbf{Multi-Layer Acquisition ($L=2$ $\mathbb{Z}_{113}$) and Iso-DoF $S_5$ Conjugation ($aba^{-1}$, $N=12$ paired seeds, $\text{Mean} \pm \text{SEM}, \text{ddof}=1$).}
Sign-flip $p_{\rm perm}$, Holm $p_{\rm Holm}$ ($M=6$ crossover, $M=18$ Iso-DoF, $^{*}p_{\rm Holm}<0.05$, $^{**}p_{\rm Holm}<0.01$; Tables~\ref{tab:app_s5_composition}--\ref{tab:app_s5_isodof_pw}). \textbf{Panel A}: $U{+}B - \text{Full} = -4{,}475.6 \pm 76.8^{**}$; $\text{Full} - E{+}U = +704.7 \pm 84.3^{**}$ ($8.1\times$ faster trigger). \textbf{Panel B}: Col.\ 3 lists $q_{\rm trig}=0.70,0.95$ counts (of $12$); isolated $4{\times}32$ $[a,b,a]$ vs.\ $[a,b]$: $+22.74 \pm 10.91$ pp ($p_{\rm perm}=0.0576, p_{\rm Holm}=0.1152$); $2{\times}2$ Cold interaction: $-49.14 \pm 9.56$ pp$^{**}$ ($d=256$ donors $\sFiveDonorDTwoFiveSixCausalSteps$).}
\label{tab:depth_s5_acquisition}
\begin{tabular*}{\textwidth}{@{\extracolsep{\fill}}lcccc@{}}
\toprule
\multicolumn{5}{@{}l@{}}{\textbf{Panel A: 2L $\mathbb{Z}_{113}$ Residual Acquisition ($a{+}b \to 2a{+}b$, $425{,}472$ params, $10{,}000$ steps)}} \\
\midrule
\textbf{Transfer Arm} & \textbf{Trig ($q_{0.95}$)} & \textbf{$t_{\rm trig}^{\rm val}$ (steps, $\downarrow$)} & \textbf{$I_{\rm 5k}$ ($\uparrow$)} & \textbf{$\widehat{\mathrm{RMST}}_{10\mathrm{k}}^{\rm val}$ ($\downarrow$)} \\
\midrule
Cold Scratch & $0/12$ & $10{,}000.0{\pm}0.0$ & $4.7{\pm}0.1$ & $10{,}000.0{\pm}0.0$ \\
$U{+}B$ (No-$E$ Control) & $2/12$ & $9{,}525.0{\pm}427.4$ & $125.3{\pm}73.7$ & $9{,}975.0{\pm}25.0$ \\
$E{+}U$ (Boundary Only) & $12/12$ & $1{,}283.3{\pm}74.2$ & $3{,}896.3{\pm}73.4$ & $9{,}345.8{\pm}654.2$ \\
\textbf{Full ($E{+}B{+}U$)} & $\mathbf{12/12}$ & $\mathbf{158.3{\pm}5.6}$ & $\mathbf{4{,}600.9{\pm}31.3}^{**}$ & $\mathbf{7{,}895.8{\pm}1{,}021.3}$ \\
\bottomrule
\end{tabular*}
\begin{tabular*}{\textwidth}{@{\extracolsep{\fill}}lcccc@{}}
\multicolumn{5}{@{}l@{}}{\textbf{Panel B: 2L Non-Abelian $S_5$ Conjugation ($a{\circ}b \to aba^{-1}$, Crossover \& Iso-DoF Anchors, $30{,}000$ steps FP32)}} \\
\midrule
\textbf{Template / Arm} & \textbf{Params} & \textbf{Trig ($q_{0.70}, q_{0.95}$)} & \textbf{$I_{\rm full}$ ($\uparrow$)} & \textbf{Val-Sel Acc (\%, $\uparrow$)} \\
\midrule
$[a,b]$ Bidir Full ($4{\times}32$) & $427{,}008$ & $0,0$ & $8{,}622.2{\pm}1{,}541.4$ & $33.90{\pm}6.21$ \\
$[a,b,a]$ Bidir Full ($4{\times}32$) & $427{,}008$ & $3,0$ & $9{,}302.9{\pm}2{,}140.1$ & $40.11{\pm}8.99$ \\
$[a,b,a]$ Causal Full ($4{\times}32$) & $427{,}008$ & $\mathbf{9,5}$ & $17{,}745.1{\pm}2{,}141.1$ & $79.98{\pm}7.83^{**}$ \\
$[a,b,a]$ Causal \textbf{Cold} ($2{\times}64$) & $427{,}008$ & $\mathbf{12,11}$ & $15{,}715.5{\pm}846.0$ & $\mathbf{98.54{\pm}0.70}^{**}$ \\
$[a,b,a]$ Causal $E{+}U$ ($2{\times}64$) & $427{,}008$ & $\mathbf{10,5}$ & $16{,}867.0{\pm}1{,}804.1$ & $85.95{\pm}5.67^{*}$ \\
$[a,b,a]$ Causal \textbf{Full} ($2{\times}64$) & $953{,}344$ & $\mathbf{12,6}$ & $\mathbf{21{,}424.8{\pm}934.6}$ & $\mathbf{95.02{\pm}1.57}^{**}$ \\
$[a,b,a]$ Causal \textbf{Full} ($8{\times}32$) & $1{,}640{,}448$ & $\mathbf{12,9}$ & $23{,}784.5{\pm}1{,}019.1$ & $97.89{\pm}1.02^{**}$ \\
\bottomrule
\end{tabular*}
\end{table*}

\section{Iso-Parameter \texorpdfstring{$S_5$}{S5} Scaling and Two-Stage Final-Slot Refinement}
\label{sec:boundaries}
\label{sec:multilayer_governance}
\label{sec:s5_composition}

\begin{figure*}[t]
    \centering
    \includegraphics[width=0.72\linewidth]{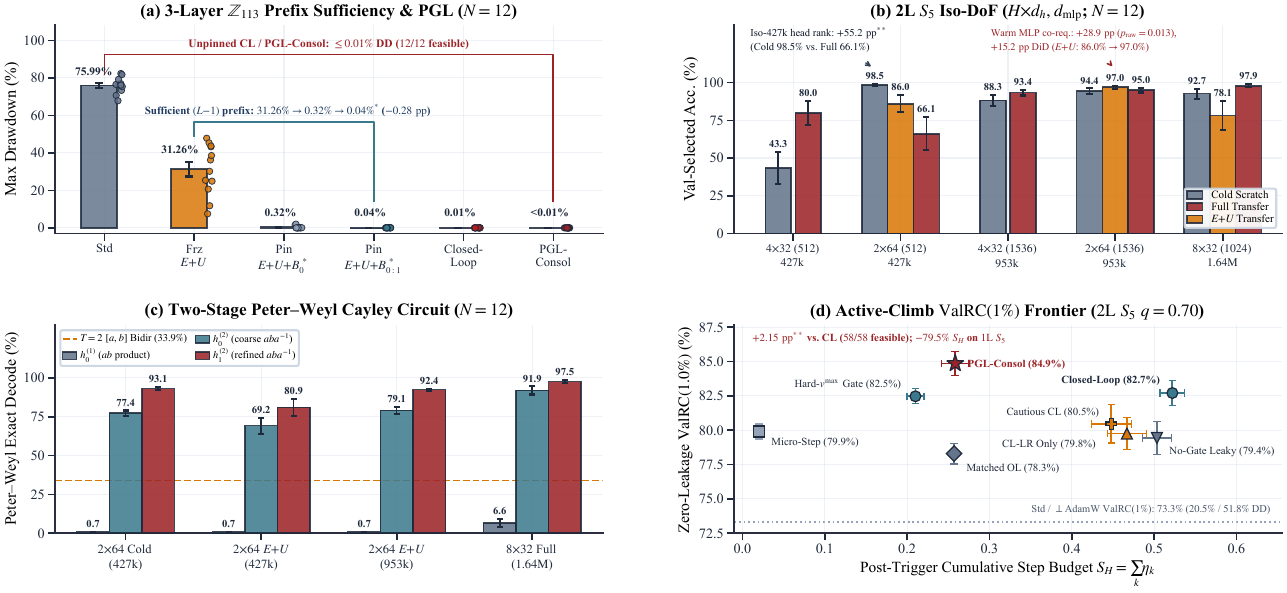}
    \caption{\textbf{Depth-3 Prefix Pinning, Iso-DoF $S_5$ Scaling, Two-Stage Decoding, and PGL-Consol Frontier ($N=12$ Paired Seeds).} (a) 3L $\mathbb{Z}_{113}$: graded drawdown reduction from Std ($75.99\%$) and Frz $E{+}U$ ($31.26\%$) to Pin $B_0^*$ ($0.32\%$), Pin $B_{0:1}^*$ ($0.04\%$), and PGL-Consol ($0.00\%$). (b) 2L $S_5$ ($aba^{-1}$): $4{\times}32 \to 2{\times}64$ at $427\text{k}$ lifts Cold to $98.54\%$ and $E{+}U$ to $85.95\%$. (c) Peter--Weyl decoding: near-chance slot-1 $a\circ b$ vs.\ slot-2 $aba^{-1}$ refinement. (d) 2L $S_5$ macro: PGL-Consol reaches $84.85\%$ $\mathrm{ValRC}(1\%)$ ($58/58$ feasible).}
    \label{fig:fig4_generality}
\end{figure*}

\subsection{Conjugacy-Class Partial Resolution and \texorpdfstring{$d_{\rm model}=128$}{d=128} Orthogonal Crossover}
Transferring from $a\circ b$ to $aba^{-1}$ in $S_5$ ($|S_5|=120$) exposes conjugacy-class partial resolution \citep{Stander2024_231206581,GroupWords2025,Liu2023_221010749}. Because $\mathrm{Cl}(aba^{-1}) = \mathrm{Cl}(b)$ has seven cycle-type classes ($1, 10, 15, 20, 20, 24, 30$), a predictor using $\mathrm{Cl}(b)$ without $a$ achieves at most $7/120 \approx 5.83\%$ (Eq.~\eqref{eq:s5_unary_floor}). Under Full transfer at $4{\times}32$ ($427\text{k}$), 2L $[a,b]$ bidirectional models reach $99.38\% \pm 0.16\%$ cycle-type accuracy and $33.90\% \pm 6.21\%$ exact accuracy ($58.58\%$ on transpositions; Table~\ref{tab:depth_s5_acquisition}). Scrambling $a \mapsto a'$ drops exact accuracy by $-28.52 \pm 6.12$ pp ($p_{\rm perm}=0.0015$) to $5.38\% \pm 0.09\%$ while preserving $99.56\% \pm 0.12\%$ cycle-type accuracy and $87.90\% \pm 7.56\%$ fixed-$b$ diversity.

In our $N=12$ crossover at $4{\times}32$ ($427\text{k}$; Table~\ref{tab:depth_s5_acquisition}), donors qualify in $\sFiveDonorDOneTwoEightCausalSteps$ causal vs.\ $\sFiveDonorDOneTwoEightBidirSteps$ bidirectional steps. Recipient causal masking adds $\mathbf{+39.87 \pm 8.00}$ \textbf{pp} ($79.98\% \pm 7.83\%$ vs.\ $40.11\% \pm 8.99\%$, $p_{\rm Holm}=0.0030$), and causal donor features add $\mathbf{+29.55 \pm 8.74}$ \textbf{pp} ($p_{\rm Holm}=0.0137$; $+16.53 \pm 3.17$ pp for $[a,b,a]$ causal vs.\ $[a,b]$ bidir under fixed bidir donor, $p_{\rm perm}=0.0020$). With causal donor and mask fixed at $4{\times}32$, the isolated $[a,b,a]$ vs.\ $[a,b]$ contrast is $+22.74 \pm 10.91$ pp ($79.98\% \pm 7.83\%$ vs.\ $57.24\% \pm 7.41\%$), which is marginal ($p_{\rm perm}=0.0576, p_{\rm Holm}=0.1152$).

\subsection{Iso-DoF Attention-vs.-MLP Asymmetry and Two-Stage Final-Slot Refinement}
Whereas $4{\times}32$ ($427\text{k}$) Cold stopped at $43.33\%$, $8{\times}32$ ($1.64\text{M}$, donors $\sFiveDonorDTwoFiveSixCausalSteps$) reached $92.68\% \pm 3.36\%$ Cold and $\mathbf{97.89\% \pm 1.02\%}$ Full ($97.97\% \pm 1.01\%$ max). Full achieves $3.2\times$ faster $q_{\rm trig}=0.70$ onset than Cold ($6{,}154.2 \pm 1{,}124.8$ vs.\ $19{,}504.2 \pm 1{,}280.5$ steps) and $+19.75$ pp over $E{+}U$ ($78.14\% \pm 9.55\%$; $3/12$ seeds lag on one class). To separate head dimension from MLP width \citep{Sanford2023_230602896,Sanford2024_240209268}, we evaluate an iso-parameter $427\text{k}$--$953\text{k}$ grid ($M=18$; Figure~\ref{fig:fig4_generality}(b--c), Tables~\ref{tab:depth_s5_acquisition},~\ref{tab:app_s5_isodof_pw}).
\begin{itemize}[leftmargin=1.4em,itemsep=0pt,topsep=1pt]
\item \textbf{Substitutive $2\times 2$ Scratch Capacity.} At fixed $427{,}008$ parameters ($d_{\rm mlp}=512$), reconfiguring $4{\times}32 \to 2{\times}64$ lifts Cold accuracy by $\mathbf{+55.21 \pm 11.00}$ \textbf{pp} to $\mathbf{98.54\% \pm 0.70\%}$ ($12/12$ triggered, $p_{{\rm Holm},18}=0.0088$). Across the $2\times 2$ head-by-MLP grid, Cold exhibits a negative substitutive interaction ($\mathbf{-49.14 \pm 9.56}$ \textbf{pp}, $p_{{\rm Holm},18}=0.0088$): widening $d_{\rm mlp}=1{,}536$ ($953\text{k}$) lifts $4{\times}32$ Cold to $88.30\% \pm 3.77\%$, whereas shrinking $d_{\rm mlp}=354$ at $8{\times}32$ ($953\text{k}$) drops Cold to $46.89\% \pm 10.37\%$.
\item \textbf{Boundary Transfer vs.\ Warm-Block MLP Co-Requirement.} At $2{\times}64, d_{\rm mlp}=512$ ($427\text{k}$), $E{+}U$ alone reaches $\mathbf{85.95\% \pm 5.67\%}$ ($\mathbf{+60.93 \pm 9.98}$ \textbf{pp} over $U{+}B$ at $25.02\% \pm 9.57\%$, $p_{{\rm Holm},18}=0.0146$), adding $+367.2 \pm 132.2$ $I_{\rm 5k}$ over Cold. In contrast, warm donor blocks ($2{\times}64$ Full) slow reconfiguration at $d_{\rm mlp}=512$ ($66.08\% \pm 10.97\%$). Widening $d_{\rm mlp}=1{,}536$ at $953\text{k}$ lifts $2{\times}64$ $E{+}U$ by $+11.02 \pm 6.11$ pp to $\mathbf{96.97\% \pm 0.95\%}$ ($p_{\rm raw}=0.0942$; $+15.18 \pm 7.56$ pp interaction, $p_{\rm raw}=0.0459, p_{{\rm Holm},18}=0.4590$) and $2{\times}64$ Full by $+28.94 \pm 11.27$ pp to $\mathbf{95.02\% \pm 1.57\%}$ ($p_{\rm raw}=0.0132, p_{{\rm Holm},18}=0.1714$). Whereas scratch $2{\times}64$ co-adapts $(W_E, B_{0:1}, W_U)$ jointly ($\mathrm{LN}_f$ $R^2$: $\mathrm{std}_4\ 0.285, \mathrm{sgn}\ 0.972$), fixed donor $E{+}U$ requires MLP capacity to align onto the preset chart ($\mathrm{std}_4\ 0.322, \mathrm{std{\cdot}sgn}_4\ 0.248$).
\item \textbf{Direct Two-Stage Final-Slot Refinement.} Across all 15 suites in Table~\ref{tab:app_s5_isodof_pw}, ridge-calibrated Peter--Weyl probes at slot $1$ ($h_0^{\rm slot1}$) decode $a\circ b$ near chance ($0.48\%$--$6.64\%$ vs.\ $0.83\%$). Across the ten grokked suites, slot $2$ refines $aba^{-1}$ directly from $h_0^{\rm slot2}$ ($68.98\%$--$91.92\%$) to $h_1^{\rm slot2}$ ($80.94\%$--$97.50\%$), lifting $2{\times}64$ Cold from $77.37\% \pm 1.65\%$ to $93.12\% \pm 1.03\%$ and $1.64\text{M}$ Full from $91.92\% \pm 2.62\%$ to $97.50\% \pm 1.08\%$.
\end{itemize}

\begin{table*}[t]
\centering
\scriptsize
\setlength{\tabcolsep}{2.8pt}
\renewcommand{\arraystretch}{0.68}
\caption{\textbf{Real-Sensory MNIST Modular Arithmetic, Norm-Scale Deconfounding, and Zero-Retuning PGL-Consol ($N=12$ paired seeds, $523{,}136$ params).}
Five image-disjoint pools ($15{,}000$ donor-train images, $40{,}000$ pairs; sign-flip $p_{\rm perm}$, Holm $p_{\rm Holm}$ over $M=5,3,8$, $^{**}p_{\rm Holm}<0.01$; Table~\ref{tab:app_mnist_full}). \textbf{Panel A}: donor arms rescale to $1.0\times n_c$ vs.\ RandOrth $1.0\times n_d$; CRT $\{k{=}2,4,5\}$ DFT energy is $94.6\%$ (Full) and $94.7\%$ (Cold), with $k{=}5$ at $18.03\% \pm 0.72\%$ (PGL) vs.\ $8.84\% \pm 0.54\%$ (Std). \textbf{Panel B}: PGL vs.\ Closed-Loop $\mathrm{ValRC}(1\%)$: $+1.67 \pm 0.15$ pp$^{**}$.}
\label{tab:mnist_main}
\begin{tabular*}{\textwidth}{@{\extracolsep{\fill}}lcccc@{}}
\toprule
\multicolumn{5}{@{}l@{}}{\textbf{Panel A: Real-Sensory MNIST Acquisition ($\mathrm{Op1}: 2y_1{+}y_2$; $\mathrm{Op2}: y_1{-}y_2$, $\bmod 10$; $12/12$ triggered at $q_{\rm trig}{=}0.82$)}} \\
\midrule
\textbf{Target / Arm} & \textbf{Norm Scale} & \textbf{$I_{\rm 1k}$ ($\uparrow$)} & \textbf{$\widehat{\mathrm{RMST}}_{\rm val}$ ($\downarrow$)} & \textbf{Val-Sel Acc (\%, $\uparrow$)} \\
\midrule
Op1: Cold Scratch & $1.0\times n_c$ & $640.1{\pm}7.9$ & $575.0{\pm}17.9$ & $90.52{\pm}0.18$ \\
Op1: RandOrth ($n_d$) & $1.0\times n_d$ & $713.9{\pm}3.0$ & $475.0{\pm}34.5$ & $90.44{\pm}0.24$ \\
Op1: $U{+}B$ (No-$E_\phi$) & $1.0\times n_c$ & $450.1{\pm}37.6$ & $1{,}012.5{\pm}107.3$ & $89.50{\pm}0.18$ \\
Op1: $E_\phi$ Only & $1.0\times n_c$ & $802.4{\pm}4.9$ & $237.5{\pm}9.0$ & $91.94{\pm}0.15$ \\
Op1: $E_\phi{+}U$ & $1.0\times n_c$ & $810.5{\pm}4.5$ & $216.7{\pm}7.1$ & $91.73{\pm}0.21$ \\
Op1: \textbf{Full ($E_\phi{+}B{+}U$)} & $1.0\times n_c$ & $\mathbf{865.4{\pm}5.7}^{**}$ & $\mathbf{158.3{\pm}8.3}$ & $\mathbf{92.27{\pm}0.31}$ \\
Op2: $E_\phi{+}U$ & $1.0\times n_c$ & $721.1{\pm}17.4$ & $400.0{\pm}41.3$ & $88.77{\pm}0.32$ \\
Op2: \textbf{Full ($E_\phi{+}B{+}U$)} & $1.0\times n_c$ & $\mathbf{852.5{\pm}4.8}^{**}$ & $\mathbf{158.3{\pm}8.3}$ & $\mathbf{90.76{\pm}0.32}^{**}$ \\
\bottomrule
\end{tabular*}
\begin{tabular*}{\textwidth}{@{\extracolsep{\fill}}lccccc@{}}
\multicolumn{6}{@{}l@{}}{\textbf{Panel B: Zero-Retuning 5-Suite Consolidation Macro ($60$ branches across $5$ suites, $q_{\rm trig}{=}0.82, H{=}2{,}000, M{=}8$)}} \\
\midrule
\textbf{Controller Arm} & \textbf{Feas.} & \textbf{$S_H$} & \textbf{Max DD (\%, $\downarrow$)} & \textbf{Term Acc (\%, $\uparrow$)} & \textbf{$\mathrm{ValRC}(1\%)$ (\%, $\uparrow$)} \\
\midrule
Std ($\eta_0{=}1.5{\times}10^{-3}, \lambda{=}0.8$) & $0/60$ & $3.000{\pm}0.000$ & $25.02{\pm}3.69$ & $86.96{\pm}0.27$ & $87.55{\pm}0.18$ \\
Matched Open-Loop & $59/60$ & $0.061{\pm}0.000$ & $0.23{\pm}0.02$ & $\mathbf{91.30{\pm}0.12}$ & $91.26{\pm}0.15$ \\
Micro-Step ($S_H{=}0.020$) & $\mathbf{60/60}$ & $0.020{\pm}0.000$ & $\mathbf{0.02{\pm}0.00}$ & $90.47{\pm}0.12$ & $90.47{\pm}0.12$ \\
Closed-Loop ($v^{\max}$) & $32/60$ & $0.323{\pm}0.007$ & $1.03{\pm}0.06$ & $89.44{\pm}0.17$ & $88.76{\pm}0.26$ \\
\textbf{PGL-Consol (Ours)} & $\mathbf{60/60}$ & $0.061{\pm}0.001$ & $0.09{\pm}0.01$ & $90.42{\pm}0.13$ & $\mathbf{90.42{\pm}0.13}^{**}$ \\
\bottomrule
\end{tabular*}
\end{table*}

\section{Real-Sensory MNIST Transfer, CRT Pruning, and Consolidation}
\label{sec:mnist}

\begin{figure*}[t]
    \centering
    \includegraphics[width=0.76\linewidth]{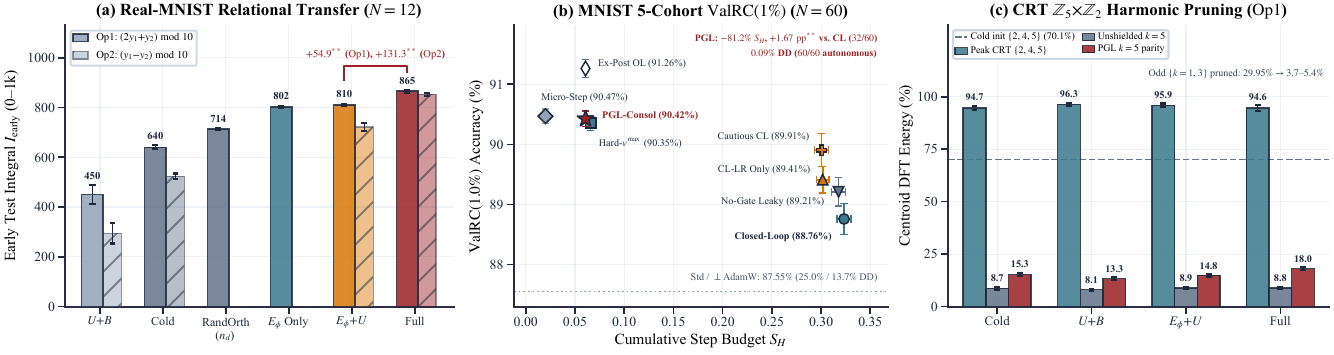}
    \caption{\textbf{Real-Sensory MNIST Transfer, Zero-Retuning PGL-Consol, and CRT Pruning ($N=12$ Paired Seeds).} (a) Image-disjoint $\mathrm{Op1}$ and $\mathrm{Op2}$: Full ($1.0\times n_c$) raises $I_{\rm 1k}$ to $865.4$ on $\mathrm{Op1}$ and reaches $90.76\%$ on $\mathrm{Op2}$ ($+1.99$ pp over $E{+}U$, $9.0\times$ faster RMST). (b) 5-suite macro ($60$ branches): PGL-Consol cuts $S_H$ by $81.2\%$ and reaches $90.42\%$ $\mathrm{ValRC}(1\%)$ ($60/60$ feasible, $+1.67$ pp vs.\ CL). (c) Centroid DFT prunes odd $k\in\{1,3\}$ onto CRT $\{k=2,4,5\}$, and PGL preserves $k=5$ parity ($18.03\%$ vs.\ $8.84\%$ Std).}
    \label{fig:fig5_mnist}
\end{figure*}

To test perceptual inputs beyond lookup tables, we train our $523{,}136$-parameter Conv2d-trunk and 2L Transformer ($E_\phi, B, U$; Appendix~\ref{app:arch_details}) across five image-disjoint MNIST pools ($N=12$ paired seeds; Table~\ref{tab:mnist_main}, Figure~\ref{fig:fig5_mnist}). Donors on $y_1+y_2\bmod 10$ ($15{,}000$ images / $40{,}000$ pairs) qualify in $4{,}933.3 \pm 203.5$ steps ($95.49\% \pm 0.17\%$) without digit labels.

\paragraph{Norm-Scale Deconfounding and Target-Dependent Block Utility.}
Donor arms ($E_{\rm only}$, $E{+}U$, $U{+}B$, Full) rescale transferred tensors to $1.0\times$ cold norm $n_c$ (Eq.~\eqref{eq:scale_matching}), whereas \textbf{RandOrth} matches trained donor norms $n_d$ ($1.2\times$--$5.4\times n_c$). On $\mathrm{Op1}:\ 2y_1+y_2\bmod 10$ ($1{,}800$ images / $3{,}500$ pairs), RandOrth slightly lifts $I_{\rm 1k}$ over Cold ($713.9 \pm 3.0$ vs.\ $640.1 \pm 7.9$) without peak gain ($90.44\% \pm 0.24\%$ vs.\ $90.52\% \pm 0.18\%$). At $1.0\times n_c$, Full reaches $I_{\rm 1k} = \mathbf{865.4 \pm 5.7}$ and $\mathrm{RMST}_{\rm val} = 158.3 \pm 8.3$ ($3.6\times$ faster than Cold $575.0 \pm 17.9$; $+151.5 \pm 7.3$ $I_{\rm 1k}$ over RandOrth, $p_{{\rm Holm},5}=0.0024$). Co-transferring $B$ accelerates takeoff over $E_{\rm only}$ ($\mathbf{+62.9 \pm 6.6}$ $I_{\rm 1k}$, $p_{{\rm Holm},5}=0.0024$) and $E{+}U$ ($+54.9 \pm 7.3$). Peak accuracy matches $E_{\rm only}$ ($92.27\% \pm 0.31\%$ vs.\ $91.94\% \pm 0.15\%$, $+0.33 \pm 0.27$ pp, $p_{\rm raw}=0.2656$), whereas omitting $E_\phi$ ($U{+}B$) lags Cold ($I_{\rm 1k} = 450.1 \pm 37.6$; $-360.4 \pm 39.6$ vs.\ $E{+}U$). On $\mathrm{Op2}:\ y_1-y_2\bmod 10$, Full achieves $9.0\times$ faster onset than Cold ($158.3 \pm 8.3$ vs.\ $1{,}420.8 \pm 141.9$) and gains $\mathbf{+1.99 \pm 0.40}$ \textbf{pp} over $E{+}U$ ($\mathbf{90.76\% \pm 0.32\%}$ vs.\ $88.77\% \pm 0.32\%$, $p_{{\rm Holm},3}=0.0015$; $+3.91 \pm 0.35$ pp over Cold).

\paragraph{Cluster Separability vs.\ Unsupervised CRT Pruning and Zero-Retuning PGL-Consol.}
On $E_\phi(I) \in \mathbb{R}^{128}$ (Table~\ref{tab:app_mnist_full} Panel A), a supervised 10-class ridge probe and 9-mode Fourier $R^2$ ($S_{10}$-permutation-invariant \emph{cluster separability}) show donor $E_\phi$ initializes at $97.26\% \pm 0.07\%$ and $0.808 \pm 0.002$ without digit training labels (vs.\ $88.57\% \pm 0.25\%$ and $0.570 \pm 0.003$ cold). Conversely, the unsupervised DFT over class centroids $\{\bar{z}_d\}_{d=0}^9$ tracks cyclic $\mathbb{Z}_{10}$ order: under $\mathbb{Z}_{10} \cong \mathbb{Z}_5 \times \mathbb{Z}_2$, $2y_1 \bmod 10$ annihilates $\mathbb{Z}_2$ parity on slot $0$ while retaining $k\in\{2,4\}$ and slot-$1$ $k=5$ parity \citep{Wu2025_241007476}. Training on $\mathrm{Op1}$ prunes odd modes $k\in\{1,3\}$ from $29.95\% \pm 0.50\%$ (Cold/$U{+}B$) and $13.59\% \pm 1.67\%$ (donor $E_\phi$) to $3.70\%$--$5.42\%$, concentrating $94.58\%$--$96.30\%$ into $\{k=2,4,5\}$. Post-trigger ($q_{\rm trig}=0.82, H=2{,}000$), Std AdamW suffers $25.02\% \pm 3.69\%$ DD across $60$ branches ($\perp\text{AdamW}$ $13.66\% \pm 0.86\%$) and erodes Full $k=5$ energy to $8.84\% \pm 0.54\%$. With \textbf{zero hyperparameter retuning}, PGL-Consol preserves $k=5$ parity at $\mathbf{18.03\% \pm 0.72\%}$ ($+9.20 \pm 0.78$ pp) and holds DD to $\mathbf{0.09\% \pm 0.01\%}$ ($S_H = 0.061 \pm 0.001$, $-81.20\%$ vs.\ CL). It reaches $\mathbf{90.42\% \pm 0.13\%}$ $\mathrm{ValRC}(1\%)$ across $60/60$ feasible branches ($\mathbf{+1.67 \pm 0.15}$ \textbf{pp} over CL $88.76\% \pm 0.26\%$, $32/60$ feasible, $p_{{\rm Holm},8}=0.0039$).

\section{Related Work}
\label{sec:related}

\paragraph{Grokking Circuits, Compositional Depth, and Perceptual Extensions.}
Grokking studies \citep{Power2022_220102177,Liu2022_220510343,Liu2022_221001117} trace delayed generalization to Fourier/group circuits \citep{Nanda2023_230105217,Gromov2023_230102679,Chughtai2023_230203025} shaped by margin maximization \citep{Morwani2024_231107568,Wu2025_241007476}, circuit competition \citep{Merrill2023_230311873,Varma2023_230902390}, lazy-to-rich dynamics \citep{Kumar2023_231006110,Mohamadi2024_240712332}, scaling geometry \citep{Tian2025_ModularScaling,Beck2024_241004489,Han2025_250917738}, and perceptual manifolds \citep{Humayun2024_240215555}. Group-word \citep{Stander2024_231206581,GroupWords2025} and Transformer width/head analyses \citep{Sanford2023_230602896,Sanford2024_240209268,Wang2024_240515071,Liu2023_221010749} study scratch models; we isolate cross-operator head-versus-MLP substitutivity, two-stage $S_5$ refinement, and MNIST CRT pruning (Table~\ref{tab:related_comparison}).

\paragraph{Feature Transfer, Plasticity, and Preconditioner-Governed Stabilization.}
Whereas \citet{Xu2025_GrokTransfer} transfer symbolic embeddings (cf.\ \citealp{Lee2024_240520233}) and prior studies address weight-norm inflation \citep{Liu2022_221001117} or two-stage schedules (\citealp{Kumar2022_LPFT,Wen2026_NN_Curriculum}), we localize multi-layer block synergy, warm-transfer MLP co-requirements, and perceptual-trunk alignment. Late AdamW instabilities \citep{Loshchilov2019_AdamW} relate to slingshot cycles \citep{Thilak2022_Slingshot}, softmax collapse ($\perp\text{AdamW}$; \citealp{Prieto2025_250104697}), readout masking \citep{Janati2026_260807436}, and plasticity loss \citep{Lyle2023_230301486,Dohare2024_NaturePlasticity}. Extending second-moment bounds \citep{Reddi2018_190409237}, validation damping \citep{Defazio2024_240515682}, multi-timescale momentum \citep{Pagliardini2025_240903137}, and cautious updates \citep{Liang2025_241116085}, PGL-Consol pairs plateau-gated damping with leaky second-moment floors under zero-leakage multi-objective risk control \citep{Wang2026_NN_GANN}.

\section{Discussion, Limitations, and Conclusion}
\label{sec:conclusion}
Across $\mathbb{Z}_{113}$, $S_5$, and image-disjoint MNIST, cross-operator transfer relies on aligned upstream interfaces ($E/E_\phi, B$) and substitutive head-versus-MLP capacity ($98.54\%$ $2{\times}64$ Cold at $427\text{k}$; $95.02\%$--$97.89\%$ warm Full with wide MLPs), while PGL-Consol governs post-grokking retention ($84.85\%$ $\mathrm{ValRC}(1\%)$ on 2L $S_5$; $90.42\%$ across $60/60$ MNIST branches without retuning). Scope boundaries: experiments span $L \in \{1,2,3\}$ on algebraic and two-digit MNIST tasks under decoupled AdamW ($\lambda \in \{0.8, 1.0\}$) rather than open-ended generation, and Propositions~\ref{prop:precond_path}--\ref{prop:pgl_bounds} bound step denominators rather than Hessian curvature. Moreover, second-moment anchoring trades some unconstrained climb ($-0.49$ to $-2.28$ pp macro; $-6.05$ pp at $1.64\text{M}$) for $\le 1\%$ validation risk, and $3/12$ seeds at $1.64\text{M}$ $S_5$ lag on one class within $30\text{k}$ steps.

\paragraph{Reproducibility Statement.} Architectures, splits, disjoint locks (\texttt{370}--\texttt{375}, \texttt{500}--\texttt{502}, \texttt{400}--\texttt{401}), and statistical families appear in \S\ref{sec:protocol} and Appendices~\ref{app:probe_details}--\ref{app:arch_details} (Table~\ref{tab:cohort_provenance}). The supplementary archive stages \texttt{src/}, execution and sub-2s verification \texttt{scripts/}, \texttt{tests/}, \texttt{figures/}, and CSV/JSON \texttt{results/}, omitting binary \texttt{.pt} checkpoints for archive size. \textbf{AI Use Disclosure:} LLM assistants (Gemini and Claude) aided coding and editing; the author designed the study, audited all claims, and takes full responsibility.

\clearpage
{\sloppy\hbadness=10000
\bibliography{refs}
\bibliographystyle{iclr2027_conference}
}
\clearpage
\appendix
\section{Diagnostic Linear Probing Specification, Optimization Convergence Checks, and Full Accounting}
\label{app:probe_details}

This appendix provides the technical specification, optimization convergence checks, trajectory event accounting, and numerical results for the diagnostic linear probing investigation, together with a structured paradigm comparison (Table~\ref{tab:related_comparison}).

\begin{table*}[!htbp]
\centering
\scriptsize
\setlength{\tabcolsep}{2.8pt}
\renewcommand{\arraystretch}{0.78}
\caption{\textbf{Comparison of Grokking, Feature-Transfer, and Post-Grokking Stabilization Paradigms} (\citealp{Power2022_220102177,Nanda2023_230105217}; entries reflect the evaluated scope in each respective study).}
\label{tab:related_comparison}
\begin{tabular*}{\textwidth}{@{\extracolsep{\fill}}lccccc@{}}
\toprule
\textbf{Study} & \textbf{Transfer} & \textbf{Depth / DoF} & \textbf{Domain} & \textbf{Precond.} & \textbf{Risk Gate} \\
\midrule
\citet{Nanda2023_230105217} & Scratch & $1\text{L}$ scratch & $\mathbb{Z}_p$ sym. & Std EMA & None \\
\citet{Kumar2022_LPFT} & Head+FT & Unpinned FT & Vision OOD & SGD/EMA & 2-stage \\
\citet{Xu2025_GrokTransfer} & Input $E$ & Shallow $E$ (1L) & $\mathbb{Z}_p$ sym. & Std EMA & Static $t{=}0$ \\
\citet{Prieto2025_250104697} & Scratch & Unpinned (1L) & $\mathbb{Z}_p$ sym. & Radial $g_\perp$ & Cont. \\
\citet{Janati2026_260807436} & Scratch & Readout (1L) & $\mathbb{Z}_p$ sym. & Diag. & Offline \\
\textbf{Ours (PGL-Consol)} & \textbf{$\{E/E_\phi,B,U\}$} & \textbf{$(L{-}1)$ \& Iso-DoF} & \textbf{$\mathbb{Z}_{113}, S_5$, MNIST} & \textbf{$v_{k}^{\rm pgl}$ / $v_k^{\max}$} & \textbf{$\mathrm{ValRC}(\tau)$} \\
\bottomrule
\end{tabular*}
\end{table*}

\subsection{Diagnostic Figures}
\label{app:mechanism}
\begin{figure*}[!htbp]
    \centering
    \includegraphics[width=0.76\linewidth]{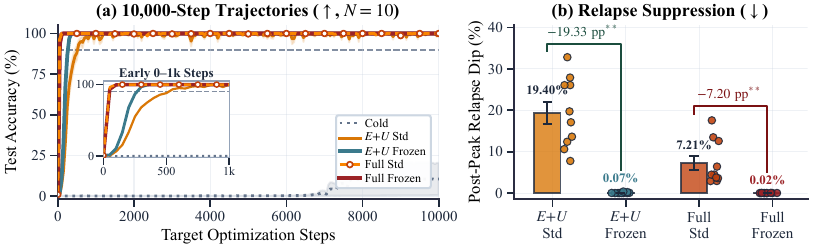}
    \caption{\textbf{Takeoff Velocity vs.\ Trajectory Stability ($N=10$ Paired Seeds, 70 Runs).} (a) Test-accuracy trajectories ($\pm$ SEM): Full reaches $90\%$ by step 50 vs.\ 250--500 for $E{+}U$, whereas unshielded continuation (Std) relapses. (b) Offline $\{W_E, W_U\}$ freezing cuts $E{+}U$ post-peak dip $\Delta_{\rm relapse} = A_{\rm peak} - \min_{t>t_{\rm peak}} A(t)$ from $19.40\%$ to $0.07\%$ (corresponding to $20.34\% \to 0.09\%$ running-peak drawdown $D^{\rm max}$, $p_{\rm Holm}=0.0059$), while Full Frozen retains 50-step takeoff with $0.02\%$ dip.}
    \label{fig:fig1_lifecycle}
\end{figure*}
\begin{figure*}[!htbp]
    \centering
    \includegraphics[width=0.95\textwidth]{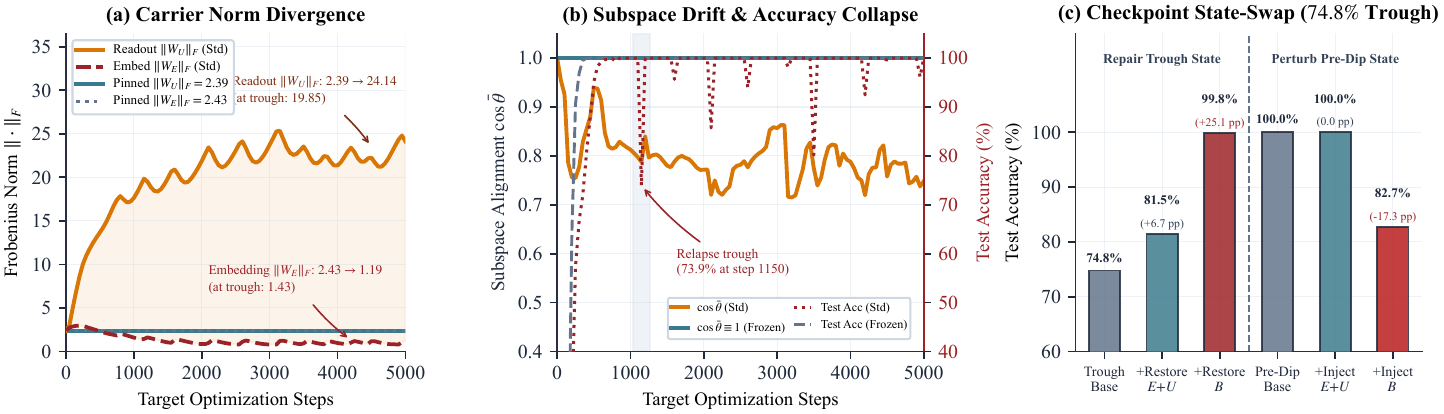}
    \caption{\textbf{Exploratory Trajectory and State-Swap Diagnostics.} (a--b) Embedding/readout norms and embedding-subspace alignment evolve along a 5,000-step unshielded trajectory ($73.9\%$ relapse trough at step 1,150); alignment is not monotone in accuracy. (c) On the saved pre-relapse/trough checkpoint pair ($74.76\%$ trough), restoring internal blocks recovers $99.85\%$ accuracy, whereas restoring $E{+}U$ reaches $81.45\%$. These single-trajectory observations localize the contemporaneous functional deficit within the probed checkpoint pair.}
    \label{fig:fig2_mechanism}
\end{figure*}

\paragraph{Grassmann Subspace Alignment Metric.}
In Figure~\ref{fig:fig2_mechanism}(b), we measure alignment between token embeddings $W_E(t) \in \mathbb{R}^{p \times d_{\rm model}}$ and donor initialization $W_E(0)$. Specifically, we compute the mean principal angle cosine across their top $k=12$ left singular subspaces:
\begin{equation}
\label{eq:grassmann_alignment}
\cos\bar{\theta}(t) = \frac{1}{k} \sum_{i=1}^k \sigma_i(Q(t)^\top Q^{\rm donor}) = \frac{1}{k} \sum_{i=1}^k \cos\theta_i.
\end{equation}
Here $Q(t), Q^{\rm donor} \in \mathbb{R}^{p \times k}$ are orthonormal bases for the top $k$ left singular vectors of $W_E(t)$ and $W_E(0)$, with singular values $\sigma_i$.

\begin{figure*}[!htbp]
    \centering
    \includegraphics[width=0.95\textwidth]{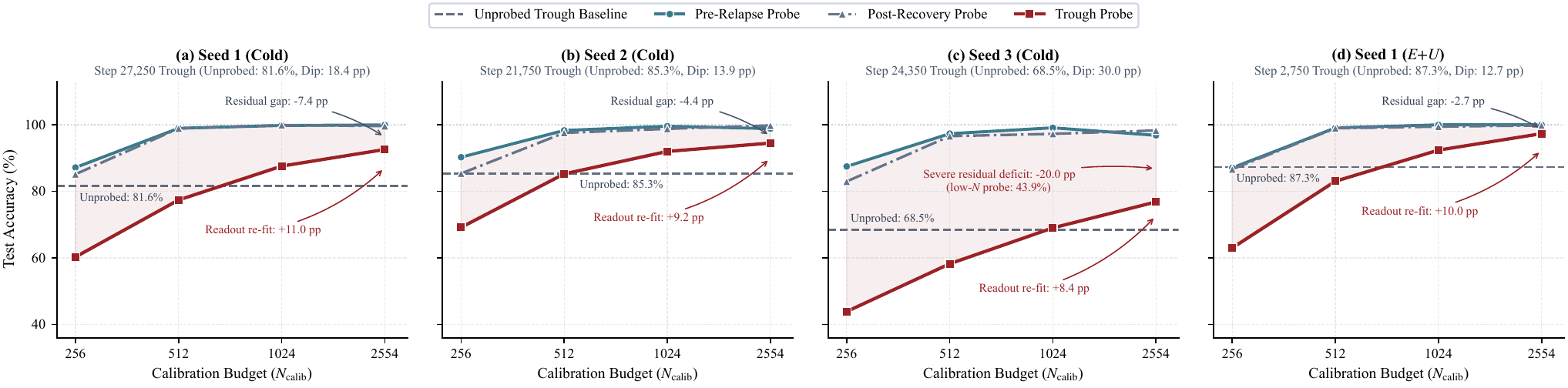}
    \caption{\textbf{Linear Probes on Four Test-Conditioned Relapse Events.} Test accuracy for probes fitted to pre-relapse, trough, and recovery representations across calibration budgets $N \in \{256, 512, 1024, 2554\}$ (Seeds 1--4 map to raw trajectory seed IDs \texttt{210}--\texttt{213} for Cold, \texttt{220}--\texttt{223} for $E{+}U$, and \texttt{230}--\texttt{233} for Full; three qualified events are Cold and one is $E{+}U$). Trough Probe accuracy improves with larger sample budgets but remains below paired Pre-Relapse Probe accuracy under the tested protocol. This finite-budget deficit neither establishes irreversible information loss nor excludes a nonlinear decoder.}
    \label{fig:fig_probe_events}
\end{figure*}

\subsection{Protocol Specification}

\paragraph{Motivation and Scope.}
Following recent analyses distinguishing circuit failure from circuit masking \citep{Janati2026_260807436}, we test whether post-grokking relapse reflects intrinsic feature loss or geometric misalignment with the readout head. We evaluate an exploratory development battery of 12 training trajectories across Seeds 1--4 spanning Cold, $E{+}U$, and Full transfer. Under $E{+}U$ and Full transfer, transferred 2D matrices use per-tensor Frobenius scale matching via Eq.~\eqref{eq:scale_matching}, while 1D LayerNorm parameters and positional embeddings remain cold.

\paragraph{Representation Extraction and Consistency Verification.}
At the final token position, representations $z(x) = \mathrm{LN}_f(h_{L-1}(x)_{-1}) \in \mathbb{R}^{d_{\rm model}}$ ($L=1$, $d_{\rm model}=128$) are extracted after terminal LayerNorm and prior to readout $W_U \in \mathbb{R}^{p \times d_{\rm model}}$. For every evaluated checkpoint, applying the model readout head to $z(x)$ reproduces the forward logits $f(x)$ within numerical precision:
\begin{equation}
\max_{x \in \mathcal{D}_{\rm test}} \| W_U z(x) - f(x) \|_\infty < 10^{-5},
\end{equation}
confirming that probed features match the operational input to the classification head.

\paragraph{Event Detection via Running Peak Drawdown.}
To avoid censoring early relapses occurring before the global trajectory maximum, events are tracked via running peak drawdown. Let $M(t) = \max_{s \le t} A_{\rm test}(s)$ denote running peak accuracy after validation breakthrough ($A_{\rm val}(t) \ge 95\%$ for two checks at cadence $\Delta=50$). We track three stages:
\begin{enumerate}[leftmargin=*,itemsep=0pt,topsep=1pt]
    \item \textbf{Pre-Relapse Peak ($t_{\rm pre}$)}: the running peak immediately preceding a drop.
    \item \textbf{Relapse Trough ($t_{\rm trough}$)}: the local minimum where drawdown $\Delta_{\rm dip}(t) = M(t) - A_{\rm test}(t) \ge 10.0$ pp and $A_{\rm test}(t) < 90.0\%$.
    \item \textbf{Post-Recovery ($t_{\rm recov}$)}: the first subsequent evaluation where $A_{\rm test}(t) \ge 95.0\%$.
\end{enumerate}
Events are searched within 2,000 steps after breakthrough across a 30,000-step budget for Cold runs and a 5,000-step budget for transfer runs. Runs without a qualifying drop are recorded as \emph{No Relapse}, and runs without breakthrough as \emph{Unbroken}.

\paragraph{Linear Probe Objective and Subsetting.}
At each checkpoint stage, an independent linear probe $U \in \mathbb{R}^{p \times d_{\rm model}}$ minimizes regularized cross-entropy over a calibration subset $\mathcal{S}_N \subset \mathcal{D}_{\rm train}$:
\begin{equation}
\min_{U \in \mathbb{R}^{p \times d_{\rm model}}} \mathcal{L}(U) = \frac{1}{N} \sum_{i=1}^N \mathcal{L}_{\rm CE}(U z_i, y_i) + \frac{\lambda}{2} \|U\|_F^2,
\end{equation}
across sample budgets $N \in \{256, 512, 1024, 2554\}$, where $N=2554$ is the full training partition. Subsets $\mathcal{S}_{256} \subset \mathcal{S}_{512} \subset \mathcal{S}_{1024} \subset \mathcal{S}_{2554}$ are nested leading prefixes ($\mathcal{D}_{\rm train}[:N]$) under the deterministic split for each paired seed. Candidate penalties $\lambda \in \{0,10^{-4},10^{-3},10^{-2},10^{-1},1\}$ are selected by validation cross-entropy and evaluated on $\mathcal{D}_{\rm test}$. Because event times use $A_{\rm test}(t)$, this probe serves as a test-conditioned diagnostic.

\subsection{Optimization Convergence Checks}

\paragraph{Random Stream Isolation.}
To prevent coupling between trajectory execution and probe fitting, each training run used a dedicated CUDA generator for mini-batch sequencing. Probe initialization weights were drawn from an independent CPU generator with seed $42$.

\paragraph{Solver and Gate.}
The 288 candidate fits use L-BFGS with strong Wolfe line search (maximum 500 iterations). Of these, 286 satisfy $\|\nabla\mathcal{L}\|_F<10^{-3}$; the two exceptions occur at $\lambda=1, N=2554$ and are never selected. All 48 validation-selected probes pass the gradient gate. A Seed-1 check comparing Adam and L-BFGS at one trough configuration found a 0.03-point test-accuracy difference. For $\lambda=0$, we report the finite-budget unregularized L-BFGS iterate.

\subsection{Full Trajectory Event Accounting}
Table~\ref{tab:probe_event_accounting} reports the trajectory event census across all 12 runs. Three broken Cold trajectories and one $E{+}U$ trajectory underwent qualified relapses (dips of $18.37$, $13.86$, $29.95$, and $12.70$ pp), whereas Full transfer had $0/4$ qualified relapses within $5{,}000$ steps, motivating the $N=12$ paired post-trigger trough dissections in Panel B of Table~\ref{tab:probe_full_results}.

\begin{table*}[!htbp]
\centering
\scriptsize
\setlength{\tabcolsep}{3.0pt}
\renewcommand{\arraystretch}{0.88}
\caption{\textbf{Event Accounting and Trajectory Census Across 12 Runs (Seeds 1--4).} Here Seeds 1--4 correspond to raw trajectory seed IDs \texttt{210}--\texttt{213} (Cold), \texttt{220}--\texttt{223} ($E{+}U$), and \texttt{230}--\texttt{233} (Full) in the saved CSV/JSON logs. Breakthrough is the first step with $A_{\rm val}(t) \ge 95\%$ for two consecutive evaluations. Pre-relapse peak, relapse trough, and recovery steps are tracked via running peak drawdown.}
\label{tab:probe_event_accounting}
\begin{tabular*}{\textwidth}{@{\extracolsep{\fill}}lccccccc@{}}
\toprule
\textbf{Seed (Cond.)} & \textbf{Break.} & \textbf{$t_{\rm pre}$ (Acc)} & \textbf{$t_{\rm trough}$} & \textbf{Trgh Acc} & \textbf{Dip} & \textbf{$t_{\rm recov}$ (Acc)} & \textbf{Status} \\
\midrule
Seed 1 (Cold) & $26{,}600$ & $27{,}100$ ($99.92\%$) & $27{,}250$ & $81.55\%$ & $18.37$ pp & $27{,}300$ ($99.71\%$) & \textbf{Qualified} \\
Seed 2 (Cold) & $21{,}300$ & $21{,}550$ ($99.19\%$) & $21{,}750$ & $85.33\%$ & $13.86$ pp & $21{,}800$ ($99.07\%$) & \textbf{Qualified} \\
Seed 3 (Cold) & $23{,}850$ & $24{,}150$ ($98.41\%$) & $24{,}350$ & $68.46\%$ & $29.95$ pp & $24{,}400$ ($97.61\%$) & \textbf{Qualified} \\
Seed 4 (Cold) & $>30\mathrm{k}$ & --- & --- & --- & --- & --- & Unbroken \\
\midrule
Seed 1 ($E{+}U$) & $850$ & $1{,}200$ ($100.00\%$) & $2{,}750$ & $87.30\%$ & $12.70$ pp & $2{,}800$ ($100.00\%$) & \textbf{Qualified} \\
Seed 2 ($E{+}U$) & $850$ & --- & --- & --- & $6.42$ pp & --- & No Relapse \\
Seed 3 ($E{+}U$) & $500$ & --- & --- & --- & $9.57$ pp & --- & No Relapse \\
Seed 4 ($E{+}U$) & $500$ & --- & --- & --- & $1.20$ pp & --- & No Relapse \\
\midrule
Seed 1 (Full) & $100$ & --- & --- & --- & $0.94$ pp & --- & No Relapse \\
Seed 2 (Full) & $100$ & --- & --- & --- & $5.43$ pp & --- & No Relapse \\
Seed 3 (Full) & $100$ & --- & --- & --- & $4.52$ pp & --- & No Relapse \\
Seed 4 (Full) & $100$ & --- & --- & --- & $6.89$ pp & --- & No Relapse \\
\bottomrule
\end{tabular*}
\end{table*}

\subsection{Full Linear Probe Decodability, Paired Deficits, and \texorpdfstring{$N=12$}{N=12} Full-Transfer Trough Dissection}
Table~\ref{tab:probe_full_results} reports test accuracies across calibration budgets $N \in \{256, 512, 1024, 2554\}$ for the four exploratory Cold/$E{+}U$ relapse events and the $N=12$ paired Full-transfer trough state-swap and linear-probe dissections on Batteries 1 and 4.

\begin{table*}[!htbp]
\centering
\scriptsize
\setlength{\tabcolsep}{2.8pt}
\renewcommand{\arraystretch}{0.86}
\caption{\textbf{Linear Probe Decodability and Paired $N=12$ Full-Transfer Trough State-Swap Dissection.} \textbf{Panel A}: Test accuracy (\%) of probes trained on frozen penultimate features across four Cold/$E{+}U$ relapse events (Cold sample SDs across $N \in \{256, 512, 1024, 2554\}$ are $11.68, 13.27, 11.88, 8.30$ pp). \textbf{Panel B}: Paired $N=12$ Full-transfer post-trigger relapse trough state swaps and linear probes on Battery 1 (1L $2a+b$, seeds \texttt{250}--\texttt{261}, $t_{\rm trig} = 479.2 \pm 46.7$, trough offset $\Delta k = 828.3 \pm 152.6$; all-12 swaps: Restore $B$ $39.87 \pm 12.02\%$ vs.\ Restore $E{+}U$ $35.02 \pm 8.19\%$; late troughs $k \ge 760$, $8/12$: $12.41 \pm 4.15\%$ vs.\ $16.25 \pm 2.18\%$) and Battery 4 (2L $2a+b$, seeds \texttt{310}--\texttt{321}, $t_{\rm trig} = 1{,}508.3 \pm 255.0$, $\Delta k = 720.8 \pm 137.3$; single-block swaps: Restore $B_0$ $47.96 \pm 6.28\%$ vs.\ Restore $B_1$ $17.08 \pm 3.08\%$, $\Delta = +30.88 \pm 4.56$ pp; $\text{Mean} \pm \text{SEM}, \text{ddof}=1$).}
\label{tab:probe_full_results}\label{tab:probe_full_transfer_n12}
\begin{tabular*}{\textwidth}{@{\extracolsep{\fill}}llccccccc@{}}
\toprule
\multicolumn{9}{@{}l@{}}{\textbf{Panel A: Exploratory Single-Seed Cold / $E{+}U$ Relapse Probe Curves and Paired Trough Deficits $\Delta(N)$}} \\
\midrule
\textbf{Seed (Cond.)} & \textbf{Stage ($t$)} & \textbf{Orig} & \textbf{$N{=}256$} & \textbf{$N{=}512$} & \textbf{$N{=}1024$} & \textbf{$N{=}2554$} & \textbf{Gain} & \textbf{$\Delta_{2554}$} \\
\midrule
Seed 1 (Cold) & Pre ($27{,}100$) & 99.92 & 87.17 & 98.96 & 99.79 & 99.98 & --- & --- \\
           & Trgh ($27{,}250$) & 81.55 & 60.30 & 77.41 & 87.59 & 92.58 & \textbf{+11.03} & \textbf{-7.40} \\
           & Recv ($27{,}300$) & 99.71 & 85.15 & 98.90 & 99.89 & 99.62 & --- & --- \\
\midrule
Seed 2 (Cold) & Pre ($21{,}550$) & 99.19 & 90.26 & 98.31 & 99.57 & 98.85 & --- & --- \\
           & Trgh ($21{,}750$) & 85.33 & 69.18 & 85.25 & 92.00 & 94.50 & \textbf{+9.17} & \textbf{-4.35} \\
           & Recv ($21{,}800$) & 99.07 & 85.37 & 97.57 & 98.78 & 99.80 & --- & --- \\
\midrule
Seed 3 (Cold) & Pre ($24{,}150$) & 98.41 & 87.48 & 97.34 & 99.10 & 96.83 & --- & --- \\
           & Trgh ($24{,}350$) & 68.46 & 43.90 & 58.26 & 69.03 & 76.83 & \textbf{+8.37} & \textbf{-20.00} \\
           & Recv ($24{,}400$) & 97.61 & 82.98 & 96.59 & 97.28 & 98.30 & --- & --- \\
\midrule
Seed 1 ($E{+}U$) & Pre ($1{,}200$) & 100.00 & 87.07 & 99.07 & 100.00 & 100.00 & --- & --- \\
              & Trgh ($2{,}750$) & 87.30 & 63.01 & 83.04 & 92.39 & 97.34 & \textbf{+10.04} & \textbf{-2.66} \\
              & Recv ($2{,}800$) & 100.00 & 86.66 & 98.95 & 99.45 & 99.81 & --- & --- \\
\midrule
\multicolumn{2}{@{}l}{Deficit Cold ($N{=}3$)} & --- & $-30.51 \pm 6.75$ & $-24.56 \pm 7.66$ & $-16.61 \pm 6.86$ & $-10.58 \pm 4.79$ & --- & --- \\
\multicolumn{2}{@{}l}{Deficit $E{+}U$ ($N{=}1$)} & --- & $-24.05$ & $-16.03$ & $-7.61$ & $-2.66$ & --- & --- \\
\bottomrule
\end{tabular*}
\vspace{2pt}
\begin{tabular*}{\textwidth}{@{\extracolsep{\fill}}llccccc@{}}
\multicolumn{7}{@{}l@{}}{\textbf{Panel B: Paired $N=12$ Full-Transfer Post-Trigger Trough State Swaps and Linear Probes}} \\
\midrule
\textbf{Battery} & \textbf{Stage} & \textbf{Orig (\%)} & \textbf{$N{=}256$ (\%)} & \textbf{$N{=}2554$ (\%)} & \textbf{Gain / $B$} & \textbf{$\Delta_{2554}$ / $E{+}U$} \\
\midrule
1L Full ($2a{+}b$) & Trigger ($t_{\rm trig}$) & $99.21 \pm 0.14$ & $88.63 \pm 0.48$ & $99.52 \pm 0.12$ & --- & --- \\
                   & Trough (Probes, $12/12$) & $77.88 \pm 2.42$ & $52.44 \pm 1.59$ & $90.90 \pm 1.33$ & $\mathbf{+13.01 \pm 1.44}$ & $\mathbf{-8.62 \pm 1.33}$ \\
                   & Early Swaps ($k{\le}360, 4/12$) & $73.36 \pm 4.36$ & --- & --- & $\mathbf{94.80 \pm 0.55}$ & $72.57 \pm 3.33$ \\
\midrule
2L Full ($2a{+}b$) & Trigger ($t_{\rm trig}$) & $97.78 \pm 0.45$ & $87.80 \pm 0.85$ & $99.08 \pm 0.20$ & --- & --- \\
                   & Trough (Probes, $12/12$) & $19.22 \pm 1.74$ & $24.15 \pm 3.86$ & $50.26 \pm 6.01$ & $\mathbf{+31.04 \pm 5.28}$ & $\mathbf{-48.82 \pm 6.02}$ \\
                   & State Swaps ($12/12$ all) & $19.22 \pm 1.74$ & --- & --- & $\mathbf{65.54 \pm 7.01}$ & $11.93 \pm 3.13$ \\
\bottomrule
\end{tabular*}
\end{table*}

\clearpage
\section{Proofs of Propositions 1--2, Provenance, and Extended Batteries}
\label{app:consolidation_details}

This appendix presents the algebraic proofs of Propositions~\ref{prop:precond_path}--\ref{prop:pgl_bounds} (\S\ref{app:prop_proof}), the provenance manifest across Batteries 1--13 (Table~\ref{tab:cohort_provenance}), and the post-trigger dissections across $\mathbb{Z}_{113}$ and $S_5$ (Tables~\ref{tab:app_dissection_1l}--\ref{tab:app_cohort9_3l}). It also reports the $S_5$ compositional and Iso-DoF Peter--Weyl tables (\S\ref{app:s5_composition_details}, Tables~\ref{tab:app_s5_composition}--\ref{tab:app_s5_isodof_pw}), the 14-suite PGL-Consol $\mathrm{ValRC}(\tau)$ ledger (Table~\ref{tab:app_pgl_valrc}), and the real-sensory MNIST CRT and consolidation ledger (Table~\ref{tab:app_mnist_full}).

\subsection{Complete Proofs of Propositions 1 and 2}
\label{app:prop_proof}
\begin{proof}[Proof of Proposition~\ref{prop:precond_path}]
Consider decoupled AdamW \citep{Loshchilov2019_AdamW} indexed by post-trigger step $k = 0, 1, \dots, H-1$, where $k=0$ corresponds to global step $t_{\rm trig} \ge 1$, preserving $(m_0, v_0) = (m_{\rm trig}, v_{\rm trig})$ with $v_{{\rm trig},i} \ge 0$. At global step $t_k = t_{\rm trig} + k + 1 \ge 2$ with gradient $g_{k+1}$, moments update via $m_{k+1,i} = \beta_1 m_{k,i} + (1-\beta_1) g_{k+1,i}$ and $v_{k+1,i} = \beta_2 v_{k,i} + (1-\beta_2) g_{k+1,i}^2$. Bias-corrected moments satisfy $\hat{m}_{k+1,i} = m_{k+1,i}/(1 - \beta_1^{t_k})$ and $\hat{v}_{t_k,i} = v_{k+1,i}/(1 - \beta_2^{t_k})$.

\emph{Part (i) (Preserved EMA Second Moments).} Since $(1-\beta_2)g_{k+1,i}^2 \ge 0$, unrolling $v_{k+1,i}$ from $v_{0,i} = v_{{\rm trig},i}$ gives $v_{k+1,i} = \beta_2^{k+1} v_{{\rm trig},i} + (1-\beta_2) \sum_{j=0}^{k} \beta_2^{k-j} g_{j+1,i}^2 \ge \beta_2^{k+1} v_{{\rm trig},i}$. Because $0 < 1 - \beta_2^{t_k} < 1$ for $t_k \ge 2$, every coordinate $i$ satisfies $\sqrt{\hat{v}_{t_k,i}} = \sqrt{v_{k+1,i}} / \sqrt{1 - \beta_2^{t_k}} \ge \sqrt{v_{k+1,i}} \ge \beta_2^{(k+1)/2}\sqrt{v_{{\rm trig},i}}$, with strict first inequality $\sqrt{\hat{v}_{t_k,i}} > \sqrt{v_{k+1,i}}$ whenever $v_{k+1,i} > 0$ (and $0 = 0$ when $v_{k+1,i} = 0$). Under zero post-trigger gradients ($g_{k+1}=0$), $(\beta_1/\sqrt{\beta_2})^{k+1} \approx 0.909^{k+1} < 1$ decays, whereas non-zero late mini-batch gradients arriving after $v_{k+1,i} \ll v_{{\rm trig},i}$ experience amplified preconditioned step sizes.

\emph{Part (ii) (Monotonic AMSGrad Anchoring).} With $v_{0,i}^{\max} = v_{{\rm trig},i}$ and $v_{k+1,i}^{\max} = \max(v_{k,i}^{\max}, v_{k+1,i})$, induction over $k \ge 0$ gives $v_{k+1,i}^{\max} \ge v_{{\rm trig},i}$, so $\sqrt{\hat{v}_{t_k,i}^{\max}} = \sqrt{v_{k+1,i}^{\max}}/\sqrt{1 - \beta_2^{t_k}} \ge \sqrt{v_{k+1,i}^{\max}} \ge \sqrt{v_{{\rm trig},i}}$ for all $i$. Substituting $\eta_k = \eta_0(1 - A_{\rm val}(k))$ into $\Delta \theta_{k,i} = -\eta_k(\hat{m}_{k+1,i}/(\sqrt{\hat{v}_{t_k,i}^{\max}}+\epsilon) + \lambda\theta_{k,i})$ yields Eq.~\eqref{eq:prop_precond}. Taking the Euclidean norm across coordinates with $C_{\max} := \max_{0 \le k < H}(\sum_i (|\hat{m}_{k+1,i}|/(\sqrt{v_{{\rm trig},i}}+\epsilon) + \lambda|\theta_{k,i}|)^2)^{1/2} < \infty$ gives $\|\Delta\theta_k\|_2 \le C_{\max}\eta_0(1 - A_{\rm val}(k))$, and summing over $0 \le k < H$ yields $\mathcal{P}_H \le C_{\max}\eta_0\sum_{k=0}^{H-1}(1 - A_{\rm val}(k))$.
\end{proof}

\begin{proof}[Proof of Proposition~\ref{prop:pgl_bounds}]
\emph{Part (i) (Bounded Plateau Step Budget).} Partition post-onset steps $k \ge k_0 = m_0\Delta$ into cadence blocks $m = m_0 + r$ ($r \ge 0$), over which $A_{\rm val}(m)$ and $a_m \in [0,1]$ are piecewise constant across $m\Delta \le k < (m+1)\Delta$. Because $d_m = 0$ for all $m > m_0$ and $\rho_c = e^{-\Delta/\tau_c} \in (0,1)$, induction from $a_{m_0} \in [0,1]$ yields $a_{m_0+r} = \rho_c^r a_{m_0}$ for every $r \ge 0$. Using $1 - A_{\rm val}(m_0+r) \le 1 - A_{\inf}$, summing $\eta_k$ over $k_0 \le k < K$ gives:
\begin{equation}
\sum_{k=k_0}^{K-1} \eta_k \le \eta_0(1 - A_{\inf})\Delta \sum_{r=0}^{\lfloor(K-1)/\Delta\rfloor - m_0} \rho_c^r a_{m_0} \le \frac{\eta_0(1 - A_{\inf}) a_{m_0} \Delta}{1 - \rho_c}.
\end{equation}
At locked $\eta_0=10^{-3}, \Delta=50,$ and $\tau_c=200$, we have $\rho_c = e^{-0.25} \approx 0.77880$ and $\frac{\Delta}{1-\rho_c} \approx 226.0406$. For $A_{\inf}=0.78$ and $a_{m_0} \le 1$, the bound evaluates to $10^{-3}\cdot 0.22\cdot 226.0406 \approx 0.04973 \approx 0.0497$.

\emph{Part (ii) (Positive Leaky Preconditioner Floor).} At every step $k \ge 0$, the leaky anchor $v_{k+1,i}^{\rm pgl} = \max(\gamma_k v_{k,i}^{\rm pgl}, v_{k+1,i})$ satisfies $v_{k+1,i}^{\rm pgl} \ge \gamma_k v_{k,i}^{\rm pgl}$ under $\gamma_k = e^{-a_{\lfloor k/\Delta\rfloor}/\tau_{\rm env}}$. Unrolling across $0 \le k < K$ from $v_{0,i}^{\rm pgl} = v_{{\rm trig},i}$ and splitting at $k_0 = m_0\Delta$ yields $\prod_{k=0}^{k_0-1}\gamma_k = \allowbreak \exp\bigl(-\tfrac{\Delta}{\tau_{\rm env}}\sum_{m=0}^{m_0-1} a_m\bigr)$ and $\prod_{k=k_0}^{K-1}\gamma_k \ge \allowbreak \exp\bigl(-\tfrac{\Delta a_{m_0}}{\tau_{\rm env}(1-\rho_c)}\bigr)$. Taking square roots and applying $1 - \beta_2^{t_{K-1}} \in (0,1)$ gives:
\begin{equation}
\sqrt{\hat{v}_{t_{K-1},i}^{\rm pgl}} \ge \sqrt{v_{K,i}^{\rm pgl}} \ge \exp\Bigl(-\tfrac{\Delta}{2\tau_{\rm env}}\textstyle\sum_{m=0}^{m_0-1}a_m\Bigr)\exp\Bigl(-\tfrac{\Delta a_{m_0}}{2\tau_{\rm env}(1-\rho_c)}\Bigr)\sqrt{v_{{\rm trig},i}}.
\end{equation}
At $\Delta=50$, $\tau_c=200$, $\tau_{\rm env}=400$, and $a_{m_0}\le 1$, the plateau-tail factor exceeds $e^{-226.0406/800} \approx 0.7539$, and after $\sum_{m=0}^{m_0-1}a_m = 8$ the cumulative floor exceeds $e^{-0.5 - 226.0406/800} \approx 0.4572$.
\end{proof}

\begin{table}[!htbp]
\centering
\scriptsize
\setlength{\tabcolsep}{3.0pt}
\renewcommand{\arraystretch}{0.80}
\begin{tabular}{@{}r p{0.93\textwidth}@{}}
\toprule
\multicolumn{2}{@{}l@{}}{\textbf{Algorithm 1: Online Validation-Gated Consolidation (Closed-Loop \& PGL-Consol) and Zero-Leakage $\mathrm{ValRC}(\tau)$}} \\
\midrule
\multicolumn{2}{@{}p{\textwidth}@{}}{\textbf{Input:} Pre-trigger $\mathrm{AdamW}(\eta_0, \lambda, \beta_1, \beta_2)$, cap $T_{\max}$, cadence $\Delta=50$, threshold $q_{\rm trig}$, horizon $H$, locked PGL vector $(\tau_c=200, \tau_{\rm env}=400, d_{\rm ref}=0.01)$, risk tolerance $\tau=0.01$. \textbf{Output:} Consolidated $\theta$, $D_H^{\rm val}, D_H^{\rm test}$, zero-leakage $\mathrm{ValRC}(\tau)$.} \\
\midrule
\texttt{1:} & Run pre-trigger AdamW until two consecutive $\Delta$-spaced checks satisfy $A_{\rm val}(t) \ge q_{\rm trig}$ ($t_{\rm trig}\leftarrow t$). If unreached by $T_{\max}$, record full-run trajectory metrics directly in $\mathrm{ITT}_{12}$ without zero-imputation and return $\theta_{T_{\max}}$. \\
\texttt{2:} & \textbf{Trigger ($k=0$):} Save snapshot $\theta_{t_{\rm trig}}$; preserve $(m_0, v_0) \leftarrow (m_{\rm trig}, v_{\rm trig})$, $v_0^{\rm pgl} \leftarrow v_{\rm trig}$, $L_{\rm val}^{\min}(0) \leftarrow L_{\rm val}(t_{\rm trig})$, $A_{\rm val}(0) \leftarrow A_{\rm val}(t_{\rm trig})$, $a_0 \leftarrow 1$. \\
\texttt{3:} & \textbf{for} $k = 0, \dots, H-1$ \textbf{do} \\
\texttt{4:} & \hspace{0.5em} If $k \bmod \Delta = 0$ ($k>0$, check $m=k/\Delta$), evaluate $(L_{\rm val}(m), A_{\rm val}(m))$, $d_m = \max(0, (L_{\rm val}^{\min}(m{-}1)-L_{\rm val}(m))/(L_{\rm val}^{\min}(m{-}1)+\epsilon_{\rm loss}))$, $L_{\rm val}^{\min}(m) = \min(L_{\rm val}^{\min}(m{-}1), L_{\rm val}(m))$, and $a_m = \mathrm{clip}_{[0,1]}(\rho_c a_{m-1} + \kappa_p d_m)$. \\
\texttt{5:} & \hspace{0.5em} With $m = \lfloor k/\Delta\rfloor$, set $\eta_k = \eta_0(1 - A_{\rm val}(m)) a_m$, $\gamma_k = e^{-a_m/\tau_{\rm env}}$, and update $v_{k+1,i}^{\rm pgl} = \max(\gamma_k v_{k,i}^{\rm pgl}, v_{k+1,i})$ before the AdamW step (Prop.~\ref{prop:pgl_bounds}). \\
\texttt{6:} & \textbf{return} $\theta_{\rm out} \leftarrow \theta_{t_{\rm trig}+H}$ if $D_H^{\rm val} \le \tau$ else $\theta_{t_{\rm trig}}$ (evaluating $A_{\rm test}(\theta_{\rm out})$ once for $\mathrm{ValRC}(\tau)$, Eq.~\eqref{eq:val_rc}). \\
\bottomrule
\end{tabular}
\end{table}

\begin{table*}[!htbp]
\centering
\scriptsize
\setlength{\tabcolsep}{2.8pt}
\renewcommand{\arraystretch}{0.84}
\caption{\textbf{Experimental Battery Protocol, Parameter Census, Precision, and Provenance Manifest (Batteries 1--13).} All metrics and tables are programmatically verified against raw per-seed trajectory logs and JSON receipts. Source-task donor qualification uses only the upstream donor task's split and never accesses any recipient target task's locked test split $\mathcal{D}_{\rm test}$.}
\label{tab:cohort_provenance}
\begin{tabular*}{\textwidth}{@{\extracolsep{\fill}}>{\raggedright\arraybackslash}p{0.108\textwidth}>{\raggedright\arraybackslash}p{0.118\textwidth}>{\raggedright\arraybackslash}p{0.088\textwidth}>{\raggedright\arraybackslash}p{0.088\textwidth}>{\raggedright\arraybackslash}p{0.136\textwidth}>{\raggedright\arraybackslash}p{0.100\textwidth}>{\raggedright\arraybackslash}p{0.094\textwidth}>{\raggedright\arraybackslash}p{0.126\textwidth}@{}}
\toprule
\textbf{Battery / Task} & \textbf{Source $\to$ Target ($N$ Seeds)} & \textbf{Arch / Params} & \textbf{Train / Val / Test} & \textbf{Donor Rule / Trigger $q_{\rm trig}$} & \textbf{Prec.\ / Batch} & \textbf{Eval Cadence / $H$} & \textbf{Epistemic Status} \\
\midrule
Phase B Factorial & $a{+}b \to 2a{+}b$ ($N{=}12$ paired seeds) & 1L ($227{,}712$) & 20\% / 10\% / 70\% & Val $\ge 0.90$ ($k{=}21$) / Confirm $0.90$ & Pure FP32 ($512$) & $\Delta{=}50$, $\tau{=}30{,}000$ & Prospective Confirmation \\
Phase C Offline & $a{+}b \to 2a{+}b$ ($N{=}10$ paired seeds) & 1L ($227{,}712$) & 20\% / 10\% / 70\% & Val $\ge 0.90$ ($k{=}21$) / Offline $t{=}0$ & Pure FP32 ($512$) & $\Delta{=}50$, $T{=}10{,}000$ & Prospective Confirmation \\
Battery 1 Verification & $a{+}b \to 2a{+}b$ ($N{=}12$ paired seeds) & 1L ($227{,}712$) & 20\% / 10\% / 70\% & Val $\ge 0.90$ ($k{=}21$) / $q_{\rm trig}{=}0.95$ ($k{=}2$) & AMP $\to$ FP32 ($512$) & $\Delta{=}50$ pre / $10$ post, $H{=}1{,}500$ & Prospective (Frz) / State-Isolated (12-arm) \\
Battery 2 Affine & $a{+}b \to 3a{+}b$ ($N{=}12$ paired seeds) & 1L ($227{,}712$) & 20\% / 10\% / 70\% & Val $\ge 0.90$ ($k{=}21$) / $q_{\rm trig}{=}0.95$ ($k{=}2$) & AMP $\to$ FP32 ($512$) & $\Delta{=}50$ pre / $10$ post, $H{=}1{,}500$ & Prospective (Frz) / State-Isolated (12-arm) \\
Battery 3 Quadratic & $a{+}b \to a^2{+}b^2$ ($N{=}12$ paired seeds) & 1L ($227{,}712$) & 20\% / 10\% / 70\% & Val $\ge 0.90$ ($k{=}21$) / $q_{\rm trig}{=}0.95$ ($k{=}2$) & AMP $\to$ FP32 ($512$) & $\Delta{=}50$ pre / $10$ post, $H{=}1{,}500$ & Prospective Confirmation \\
Battery 4 2L Inter-Block & 1L $a{+}b \to$ 2L $2a{+}b$ ($N{=}12$ paired seeds) & 2L ($425{,}472$) & 20\% / 10\% / 70\% & Val $\ge 0.90$ ($k{=}21$) / $q_{\rm trig}{=}0.95$ ($k{=}2$) & AMP $\to$ FP32 ($512$) & $\Delta{=}50$ pre / $10$ post, $H{=}1{,}500$ & Prospective ($E{+}U$) / State-Isolated (13-arm) \\
Battery 5 2L Acquisition & 2L $a{+}b \to$ 2L $2a{+}b$ ($N{=}12$ paired seeds) & 2L ($425{,}472$) & 30/10/60 $\to$ 20/10/70 & Val $\ge 0.95$ ($k{=}3$) + SrcHoldout $\ge 0.95$ (source $a{+}b$ split) & AMP ($1024 \to 512$) & $\Delta{=}50$, $T{=}10{,}000$ & Prospective Confirmation \\
Battery 6 1L $S_5$ & $a{\circ}b \to a b a^{-1}$ ($N{=}12$ paired seeds) & 1L ($229{,}248$) & 30\% / 20\% / 50\% & Val $\ge 0.99$ ($k{=}1$) / $q_{\rm trig}{=}0.70$ ($k{=}2$) & AMP $\to$ FP32 ($256$) & $\Delta{=}50$ pre / $20$ post, $H{=}3{,}000$ & Prospective Confirmation \\
$S_5$ High-Data / $a b^{-1}$ & $a b a^{-1}$ (45\%) / $a b^{-1}$ ($N{=}6$ paired seeds) & 1L ($229{,}248$) & 45/15/40 \& 30/20/50 & Val $\ge 0.99$ ($k{=}1$) / $q_{\rm trig}{=}0.70$ \& $0.95$ & AMP $\to$ FP32 ($256$) & $\Delta{=}50$ pre / $20$ post, $H{=}3{,}000$ & Prospective Confirmation \\
Battery 7 2L Sub & 1L $a{+}b \to$ 2L $a{-}b$ ($N{=}12$ paired seeds) & 2L ($425{,}472$) & 20\% / 10\% / 70\% & Val $\ge 0.90$ ($k{=}21$) / $q_{\rm trig}{=}0.95$ ($k{=}2$) & \textbf{Pure FP32} ($512$) & $\Delta{=}50$ pre / $10$ post, $H{=}1{,}500$ & Prospective Confirmation \\
Battery 8 2L $S_5$ $T{=}2$ & $a{\circ}b \to a b a^{-1}$ \& $a b^{-1}$ ($N{=}12$ paired seeds) & 2L ($427{,}008$) & 30/20/50 \& 45/15/40 & Val $\ge 0.99$ ($k{=}2$) / $q_{\rm trig}{=}0.70$ \& $0.95$ & \textbf{Pure FP32} ($256$) & $\Delta{=}50$ pre / $20$ post, $H{=}3{,}000$ & Prospective Confirmation \\
Battery 9 3L $\mathbb{Z}_{113}$ & 3L $a{+}b \to$ 3L $2a{+}b$ ($N{=}12$ paired seeds) & 3L ($623{,}232$) & 30/10/60 $\to$ 20/10/70 & Val $\ge 0.95$ ($k{=}3$) / $q_{\rm trig}{=}0.95$ ($k{=}2$) & \textbf{Pure FP32} ($512$) & $\Delta{=}50$ pre / $10$ post, $H{=}1{,}500$ & Prospective Confirmation \\
Battery 10 2L $S_5$ Comp. & $a{\circ}b \to$ $[a,b,a]$ $aba^{-1}$ ($N{=}12$ paired seeds) & 2L ($427{,}008$ \& $1{,}640{,}448$) & 30\% / 20\% / 50\% & Val $\ge 0.99$ ($k{=}2$) / $q_{\rm trig}{=}0.70$ \& $0.95$ & \textbf{Pure FP32} ($256$) & $\Delta{=}50$ pre / $20$ post, $H{=}3{,}000$ & Prospective Confirmation \\
Battery 11 $S_5$ Iso-DoF & $a{\circ}b \to$ $[a,b,a]$ $aba^{-1}$ ($N{=}12$ paired seeds) & 2L ($427{,}008$, $953{,}028$, \& $953{,}344$) & 30\% / 20\% / 50\% & Val $\ge 0.99$ ($k{=}2$) / $q_{\rm trig}{=}0.70$ \& $0.95$ & \textbf{Pure FP32} ($256$) & $\Delta{=}50$, $T{=}30{,}000$ + PW Probes & Prospective Confirmation \\
Battery 12 14-Suite PGL & $\mathbb{Z}_{113}$ (1--3L) \& $S_5$ (1--2L, $14$ suites) & 1--3L ($227\text{k}$--$1.64\text{M}$) & Suite-matched & Dev Lock: $\Delta{=}50$, $\tau_c{=}200$, $\tau_{\rm env}{=}400$, $d_{\rm ref}{=}0.01$ & \textbf{Pure FP32} & $\Delta{=}10$--$20$, $H{=}1{,}500$--$3{,}000$ & Prospective Confirmation \\
Battery 13 Real MNIST & $y_1{+}y_2 \to$ $2y_1{+}y_2$, $y_1{-}y_2$ ($\bmod 10, N{=}12$) & Conv+2L ($523{,}136$) & 5 Disjoint Image Pools & Donor Val Peak ($6\text{k}$) / $q_{\rm trig}{=}0.82$ ($k{=}2$, 0 retune) & \textbf{Pure FP32} ($256$) & $\Delta{=}50$, $T{=}5\text{k}$, $H{=}2\text{k}$ & Prospective Confirmation \\
\bottomrule
\end{tabular*}
\end{table*}

\begin{table*}[!htbp]
\centering
\scriptsize
\setlength{\tabcolsep}{2.8pt}
\renewcommand{\arraystretch}{0.88}
\caption{\textbf{Twelve-Arm Post-Trigger Optimization Dissection on 1-Layer Battery 1 ($2a+b$) and Battery 2 ($3a+b$; $N=12$ paired seeds each, $H=1{,}500$).} Branches fork at $t_{\rm trig}$, reporting budget $S_H = \sum_{k=0}^{H-1} \eta_k$, path length $\mathcal{P}_H = \sum_{k=0}^{H-1} \|\theta_{t_{\rm trig}+k+1} - \theta_{t_{\rm trig}+k}\|_2$, and peak validation-logit RMS drift $\mathrm{Drift} = \max_{0 \le k \le H} \|z_{\rm val}(t_{\rm trig}+k) - z_{\rm val}(t_{\rm trig})\|_F / \sqrt{|\mathcal{D}_{\rm val}|V}$ ($\text{Mean} \pm \text{SEM}, \text{ddof}=1$).}
\label{tab:app_dissection_1l}
\begin{tabular*}{\textwidth}{@{\extracolsep{\fill}}llcccccc@{}}
\toprule
\textbf{Task} & \textbf{Post-Trigger Policy} & \textbf{$S_H$} & \textbf{Max DD (\%)} & \textbf{Min Acc (\%)} & \textbf{Term Acc (\%)} & \textbf{$\mathcal{P}_H$} & \textbf{Drift} \\
\midrule
\textbf{1L $2a{+}b$} & Snapshot ($t_{\rm trig}$) & $0.000$ & $0.00 \pm 0.00$ & $99.21 \pm 0.14$ & $99.21 \pm 0.14$ & $0.00 \pm 0.00$ & $0.00 \pm 0.00$ \\
(Battery 1, & Std ($\eta{=}10^{-3}, \lambda{=}1$) & $1.500$ & $22.06 \pm 2.40$ & $77.88 \pm 2.42$ & $99.80 \pm 0.13$ & $56.28 \pm 1.58$ & $3.25 \pm 0.09$ \\
$N{=}12$) & Frz $E{+}U$ & $1.500$ & $0.60 \pm 0.16$ & $98.94 \pm 0.19$ & $99.36 \pm 0.19$ & $90.93 \pm 0.99$ & $1.04 \pm 0.03$ \\
 & Moment Reset & $1.500$ & $23.97 \pm 2.04$ & $75.95 \pm 2.08$ & $99.81 \pm 0.17$ & $51.82 \pm 0.96$ & $3.33 \pm 0.06$ \\
 & Zero WD ($\eta{=}10^{-3}, \lambda{=}0$) & $1.500$ & $45.40 \pm 2.59$ & $54.47 \pm 2.63$ & $98.09 \pm 0.32$ & $28.45 \pm 1.05$ & $4.61 \pm 0.07$ \\
 & Val-Loss Plateau Halving & Var. & $6.21 \pm 3.24$ & $93.36 \pm 3.20$ & $\mathbf{99.91 \pm 0.06}$ & $22.25 \pm 3.15$ & $2.09 \pm 0.21$ \\
 & Low-LR ($\eta{=}10^{-4}, \lambda{=}0$) & $0.150$ & $1.03 \pm 0.40$ & $98.45 \pm 0.40$ & $98.81 \pm 0.44$ & $13.62 \pm 0.19$ & $2.56 \pm 0.04$ \\
 & Low-LR + WD ($\eta{=}10^{-4}$) & $0.150$ & $0.67 \pm 0.28$ & $98.77 \pm 0.29$ & $99.24 \pm 0.33$ & $14.08 \pm 0.14$ & $1.79 \pm 0.04$ \\
 & Frz $E{+}U$ + Consol.\ $B$ & $0.020$ & $0.11 \pm 0.02$ & $99.20 \pm 0.13$ & $99.48 \pm 0.09$ & $1.43 \pm 0.04$ & $0.40 \pm 0.03$ \\
 & Exp.\ Consol.\ ($\tau{=}200, \lambda{=}1$) & $0.020$ & $0.10 \pm 0.02$ & $99.19 \pm 0.13$ & $99.59 \pm 0.08$ & $1.68 \pm 0.04$ & $0.48 \pm 0.02$ \\
 & Micro-Step ($S_H{=}0.02$) & $0.020$ & $\mathbf{0.08 \pm 0.01}$ & $\mathbf{99.20 \pm 0.13}$ & $99.58 \pm 0.07$ & $\mathbf{1.28 \pm 0.05}$ & $\mathbf{0.47 \pm 0.03}$ \\
 & Cosine ($S_H{=}0.02$) & $0.020$ & $0.12 \pm 0.03$ & $99.18 \pm 0.13$ & $99.58 \pm 0.08$ & $2.24 \pm 0.03$ & $0.55 \pm 0.02$ \\
 & Exp.\ Consol.\ + Reset & $0.020$ & $0.12 \pm 0.03$ & $99.17 \pm 0.14$ & $99.59 \pm 0.09$ & $2.53 \pm 0.02$ & $0.57 \pm 0.02$ \\
\midrule
\textbf{1L $3a{+}b$} & Snapshot ($t_{\rm trig}$) & $0.000$ & $0.00 \pm 0.00$ & $98.86 \pm 0.24$ & $98.86 \pm 0.24$ & $0.00 \pm 0.00$ & $0.00 \pm 0.00$ \\
(Battery 2, & Std ($\eta{=}10^{-3}, \lambda{=}1$) & $1.500$ & $28.56 \pm 2.40$ & $71.18 \pm 2.51$ & $98.89 \pm 1.11$ & $46.98 \pm 0.60$ & $3.13 \pm 0.05$ \\
$N{=}12$) & Frz $E{+}U$ & $1.500$ & $5.09 \pm 1.54$ & $94.28 \pm 1.53$ & $97.79 \pm 0.88$ & $80.05 \pm 1.66$ & $1.44 \pm 0.08$ \\
 & Moment Reset & $1.500$ & $22.88 \pm 2.02$ & $76.49 \pm 2.16$ & $100.00 \pm 0.00$ & $49.72 \pm 0.90$ & $3.21 \pm 0.09$ \\
 & Zero WD ($\eta{=}10^{-3}, \lambda{=}0$) & $1.500$ & $49.98 \pm 2.40$ & $49.51 \pm 2.43$ & $98.63 \pm 0.22$ & $20.72 \pm 0.30$ & $4.57 \pm 0.07$ \\
 & Val-Loss Plateau Halving & Var. & $14.38 \pm 3.88$ & $85.06 \pm 3.75$ & $\mathbf{99.87 \pm 0.08}$ & $14.02 \pm 2.45$ & $2.25 \pm 0.26$ \\
 & Low-LR ($\eta{=}10^{-4}, \lambda{=}0$) & $0.150$ & $3.47 \pm 1.77$ & $95.80 \pm 1.78$ & $96.03 \pm 1.80$ & $12.00 \pm 0.19$ & $2.40 \pm 0.02$ \\
 & Low-LR + WD ($\eta{=}10^{-4}$) & $0.150$ & $2.50 \pm 1.47$ & $96.65 \pm 1.46$ & $97.18 \pm 1.52$ & $12.58 \pm 0.16$ & $1.66 \pm 0.02$ \\
 & Frz $E{+}U$ + Consol.\ $B$ & $0.020$ & $0.17 \pm 0.05$ & $98.78 \pm 0.26$ & $99.04 \pm 0.25$ & $0.95 \pm 0.05$ & $0.38 \pm 0.02$ \\
 & Exp.\ Consol.\ ($\tau{=}200, \lambda{=}1$) & $0.020$ & $0.17 \pm 0.07$ & $98.78 \pm 0.26$ & $99.20 \pm 0.24$ & $1.15 \pm 0.06$ & $0.51 \pm 0.02$ \\
 & Micro-Step ($S_H{=}0.02$) & $0.020$ & $\mathbf{0.08 \pm 0.02}$ & $\mathbf{98.84 \pm 0.25}$ & $99.19 \pm 0.22$ & $\mathbf{0.67 \pm 0.07}$ & $\mathbf{0.37 \pm 0.02}$ \\
 & Cosine ($S_H{=}0.02$) & $0.020$ & $0.38 \pm 0.20$ & $98.60 \pm 0.32$ & $98.98 \pm 0.32$ & $1.89 \pm 0.04$ & $0.67 \pm 0.04$ \\
 & Exp.\ Consol.\ + Reset & $0.020$ & $0.52 \pm 0.30$ & $98.45 \pm 0.38$ & $98.83 \pm 0.40$ & $2.46 \pm 0.01$ & $0.73 \pm 0.04$ \\
\bottomrule
\end{tabular*}
\end{table*}

\begin{table*}[!htbp]
\centering
\scriptsize
\setlength{\tabcolsep}{2.8pt}
\renewcommand{\arraystretch}{0.88}
\caption{\textbf{Thirteen-Arm Inter-Block Interface and Consolidation Dissection on 2-Layer Battery 4 ($1\text{L } a+b \to 2\text{L } 2a+b$, $N=12$ paired seeds, $H=1{,}500$).} Reports budget $S_H = \sum_{k=0}^{H-1} \eta_k$, interface drift $\max_t \delta_{h_0}(t)$, and path length $\mathcal{P}_H = \sum_{k=0}^{H-1} \|\theta_{t_{\rm trig}+k+1} - \theta_{t_{\rm trig}+k}\|_2$ (terminal validation-logit RMS drift: $\|z_{\rm val}(t_{\rm trig}+H) - z_{\rm val}(t_{\rm trig})\|_F / \sqrt{|\mathcal{D}_{\rm val}|V}$; $\text{Mean} \pm \text{SEM}, \text{ddof}=1$).}
\label{tab:app_dissection_2l}
\begin{tabular*}{\textwidth}{@{\extracolsep{\fill}}lccccccc@{}}
\toprule
\textbf{Policy} & \textbf{$S_H$} & \textbf{Max DD (\%)} & \textbf{Min Acc (\%)} & \textbf{Term Acc (\%)} & \textbf{Max $\delta_{h_0}$} & \textbf{$\mathcal{P}_H$} & \textbf{$\le 2\%$} \\
\midrule
Snapshot ($t_{\rm trig}$) & $0.000$ & $0.00 \pm 0.00$ & $97.78 \pm 0.45$ & $97.78 \pm 0.45$ & $0.000 \pm 0.000$ & $0.00 \pm 0.00$ & 12/12 \\
Std ($\eta{=}10^{-3}$) & $1.500$ & $80.58 \pm 1.86$ & $19.22 \pm 1.74$ & $90.23 \pm 6.19$ & $1.160 \pm 0.070$ & $79.11 \pm 1.81$ & 0/12 \\
Frz $E{+}U$ & $1.500$ & $65.23 \pm 2.23$ & $34.12 \pm 2.22$ & $98.85 \pm 0.45$ & $0.927 \pm 0.064$ & $69.18 \pm 0.79$ & 0/12 \\
Frz $E{+}U{+}B_1$ & $1.500$ & $35.54 \pm 4.45$ & $62.69 \pm 4.62$ & $97.98 \pm 0.60$ & $0.717 \pm 0.045$ & $45.56 \pm 2.00$ & 0/12 \\
Frz $B_0{+}B_1$ & $1.500$ & $12.69 \pm 1.98$ & $86.72 \pm 2.09$ & $98.58 \pm 0.44$ & $0.399 \pm 0.028$ & $31.72 \pm 0.60$ & 0/12 \\
Pin $E{+}U{+}B_0$ & $1.500$ & $2.35 \pm 1.32$ & $95.91 \pm 1.33$ & $\mathbf{99.40 \pm 0.20}$ & $0.641 \pm 0.102$ & $70.60 \pm 1.35$ & 9/12 \\
Pin $E{+}U{+}B_0{+}W_{{\rm pos}}$ & $1.500$ & $1.01 \pm 0.33$ & $96.94 \pm 0.75$ & $98.60 \pm 0.42$ & $\mathbf{0.000 \pm 0.000}$ & $74.05 \pm 1.17$ & 10/12 \\
Pin $E{+}U{+}B_0^*$ & $1.500$ & $0.95 \pm 0.31$ & $96.98 \pm 0.73$ & $98.76 \pm 0.35$ & $\mathbf{0.000 \pm 0.000}$ & $77.54 \pm 1.40$ & 10/12 \\
Low-LR ($\eta{=}10^{-4}$) & $0.150$ & $1.23 \pm 0.62$ & $96.98 \pm 0.84$ & $99.07 \pm 0.50$ & $0.760 \pm 0.112$ & $13.70 \pm 0.29$ & 9/12 \\
Pin + Cons.\ $B_1$ & $0.020$ & $0.24 \pm 0.07$ & $97.63 \pm 0.49$ & $97.94 \pm 0.50$ & $0.060 \pm 0.006$ & $1.18 \pm 0.03$ & \textbf{12/12} \\
Exp.\ Consol.\ ($\tau{=}200$) & $0.020$ & $0.20 \pm 0.09$ & $97.72 \pm 0.48$ & $98.61 \pm 0.44$ & $0.228 \pm 0.019$ & $1.51 \pm 0.04$ & \textbf{12/12} \\
Micro-Step ($S_H{=}0.02$) & $0.020$ & $\mathbf{0.13 \pm 0.04}$ & $\mathbf{97.77 \pm 0.46}$ & $98.43 \pm 0.43$ & $0.130 \pm 0.009$ & $\mathbf{0.82 \pm 0.04}$ & \textbf{12/12} \\
Cosine ($S_H{=}0.02$) & $0.020$ & $0.31 \pm 0.17$ & $97.63 \pm 0.52$ & $98.63 \pm 0.52$ & $0.352 \pm 0.037$ & $2.55 \pm 0.03$ & 11/12 \\
Exp.\ Consol.\ + Reset & $0.020$ & $0.35 \pm 0.19$ & $97.58 \pm 0.54$ & $98.60 \pm 0.55$ & $0.387 \pm 0.047$ & $3.13 \pm 0.06$ & 11/12 \\
\bottomrule
\end{tabular*}
\end{table*}

\begin{table*}[!htbp]
\centering
\scriptsize
\setlength{\tabcolsep}{2.8pt}
\renewcommand{\arraystretch}{0.86}
\caption{\textbf{Prospective Confirmation on Non-Abelian $S_5$ (Battery 6 and High-Data / Right-Inverse Suites) and 2-Layer Modular Subtraction (Battery 7, Pure FP32).} Values are $\text{Mean} \pm \text{SEM}$ ($\text{ddof}=1$). On Battery 6 ($n=8$ triggered), exact sign-flip $p_{\rm raw} = 2^{-(8-1)} = 0.0078125$ with $M=4$ Holm step-down yields $p_{\rm Holm} = 0.03125^{*}$; on the $N=6$ diagnostic suites, the discrete sign-flip resolution floor is $p_{\rm raw} = 2^{-(6-1)} = 0.03125$, reported explicitly without asterisk bolding for multi-comparison family-wise significance.}
\label{tab:app_prospective_s5_sub}
\begin{tabular*}{\textwidth}{@{\extracolsep{\fill}}lcccccc@{}}
\toprule
\textbf{Post-Trigger Policy} & \textbf{Max DD (\%, $\downarrow$)} & \textbf{Min Acc (\%, $\uparrow$)} & \textbf{Term Acc (\%, $\uparrow$)} & \textbf{Gain} & \textbf{Persist} & \textbf{Holm $p$ / Note} \\
\midrule
\multicolumn{7}{@{}l@{}}{\textit{Battery 6: $1\text{L } S_5$ Conjugation ($a\circ b \to a b a^{-1}$, 30\% train, $q{=}0.70$, $N{=}12$ paired seeds, $8/12$ trig, $H{=}3{,}000$)}} \\
Snapshot ($8/12$ trig) & $0.00 \pm 0.00$ & $72.01 \pm 0.73$ & $72.01 \pm 0.73$ & $+0.00$ pp & --- & Baseline ($q_{\rm trig}{=}0.70$) \\
Std ($\eta{=}10^{-3}$) & $11.85 \pm 1.44$ & $62.06 \pm 1.66$ & $73.63 \pm 1.67$ & $+1.62$ pp & $3/8$ & --- \\
Frz $E{+}U$ & $10.73 \pm 0.89$ & $66.02 \pm 1.20$ & $73.79 \pm 1.55$ & $+1.78$ pp & $4/8$ & Null ($p{=}0.273$) \\
Micro-Step ($S_H{=}0.02$) & $\mathbf{0.09 \pm 0.03}$ & $\mathbf{72.01 \pm 0.73}$ & $77.34 \pm 0.88$ & $+5.33$ pp & $\mathbf{8/8}$ & $\mathbf{0.03125^{*}}$ \\
Exp.\ Consol.\ ($\tau{=}200$) & $0.18 \pm 0.04$ & $\mathbf{72.01 \pm 0.73}$ & $77.90 \pm 0.93$ & $+5.89$ pp & $\mathbf{8/8}$ & $\mathbf{0.03125^{*}}$ \\
Frz $E{+}U$ + Consol.\ $B$ & $0.18 \pm 0.02$ & $\mathbf{72.01 \pm 0.73}$ & $76.98 \pm 0.83$ & $+4.97$ pp & $\mathbf{8/8}$ & --- \\
Gentle Budget ($S_H{=}0.30$) & $2.10 \pm 0.41$ & $\mathbf{72.01 \pm 0.73}$ & $\mathbf{78.53 \pm 1.19}$ & $+6.52$ pp & $\mathbf{8/8}$ & Gentle diag. \\
\midrule
\multicolumn{7}{@{}l@{}}{\textit{$1\text{L } S_5$ High-Data Conjugation ($a b a^{-1}$, 45\% train, $q{=}0.70$, $N{=}6$ paired seeds, $t_{\rm trig}{=}3{,}558 \pm 935$)}} \\
Std ($6/6$ trig) & $3.72 \pm 0.32$ & $70.98 \pm 0.47$ & $\mathbf{82.05 \pm 2.79}$ & $+10.23$ pp & $6/6$ & Plateau-climb \\
Frz $E{+}U$ & $2.34 \pm 0.24$ & $71.82 \pm 0.42$ & $78.23 \pm 0.59$ & $+6.41$ pp & $6/6$ & Early ($71.82\%$) \\
Micro-Step ($S_H{=}0.02$) & $\mathbf{0.12 \pm 0.02}$ & $\mathbf{71.82 \pm 0.42}$ & $77.85 \pm 0.52$ & $+6.03$ pp & $\mathbf{6/6}$ & $p_{\rm raw}{=}0.03125$ \\
Exp.\ Consol.\ ($\tau{=}200$) & $0.23 \pm 0.03$ & $\mathbf{71.82 \pm 0.42}$ & $78.26 \pm 0.56$ & $+6.44$ pp & $\mathbf{6/6}$ & $p_{\rm raw}{=}0.03125$ \\
\midrule
\multicolumn{7}{@{}l@{}}{\textit{$1\text{L } S_5$ High-Readiness Right-Inverse ($a b^{-1}$, 30\% train, $q{=}0.95$, $N{=}6$ paired seeds, $100.00\%$ trig)}} \\
Std ($6/6$ trig) & $22.65 \pm 3.25$ & $77.35 \pm 3.25$ & $\mathbf{100.00 \pm 0.00}$ & $+0.00$ pp & $\mathbf{6/6}$ & High-readiness \\
Frz $E{+}U$ & $\mathbf{0.00 \pm 0.00}$ & $\mathbf{100.00 \pm 0.00}$ & $\mathbf{100.00 \pm 0.00}$ & $+0.00$ pp & $\mathbf{6/6}$ & $p_{\rm raw}{=}0.03125$ vs Std \\
Micro-Step / Exp.\ Consol. & $\mathbf{0.00 \pm 0.00}$ & $\mathbf{100.00 \pm 0.00}$ & $\mathbf{100.00 \pm 0.00}$ & $+0.00$ pp & $\mathbf{6/6}$ & $p_{\rm raw}{=}0.03125$ vs Std \\
\midrule
\multicolumn{7}{@{}l@{}}{\textit{Battery 7: Zero-Touch Pure-FP32 2L Subtraction ($1\text{L } a{+}b \to 2\text{L } a{-}b$, $N{=}12$ paired seeds, $H{=}1{,}500$, Pure FP32)}} \\
Snapshot ($12/12$ trig) & $0.00 \pm 0.00$ & $99.87 \pm 0.07$ & $99.87 \pm 0.07$ & $+0.00$ pp & --- & Pure-FP32 base. \\
Std ($\eta{=}10^{-3}$) & $66.81 \pm 3.06$ & $33.19 \pm 3.06$ & $\mathbf{100.00 \pm 0.00}$ & $+0.13$ pp & $0/12$ & --- \\
Frz $E{+}U$ & $2.37 \pm 1.56$ & $97.58 \pm 1.55$ & $\mathbf{100.00 \pm 0.00}$ & $+0.13$ pp & $9/12$ & Max $18.56\%$ ($1$ seed) \\
Pin $E{+}U{+}B_0$ & $0.04 \pm 0.02$ & $99.86 \pm 0.07$ & $99.99 \pm 0.01$ & $+0.12$ pp & $\mathbf{12/12}$ & --- \\
Micro-Step ($S_H{=}0.02$) & $\mathbf{0.02 \pm 0.01}$ & $\mathbf{99.87 \pm 0.07}$ & $99.97 \pm 0.02$ & $+0.10$ pp & $\mathbf{12/12}$ & $\mathbf{0.00195^{**}}$ vs Std \\
Exp.\ Consol.\ ($\tau{=}200$) & $\mathbf{0.02 \pm 0.01}$ & $\mathbf{99.87 \pm 0.07}$ & $99.98 \pm 0.01$ & $+0.11$ pp & $\mathbf{12/12}$ & $\mathbf{0.00195^{**}}$ vs Std \\
\bottomrule
\end{tabular*}
\end{table*}

\begin{table*}[!htbp]
\centering
\scriptsize
\setlength{\tabcolsep}{2.8pt}
\renewcommand{\arraystretch}{0.84}
\caption{\textbf{Eleven-Suite $2\times 2$ Controller Factorial, $\perp$AdamW, Matched-Budget Open-Loop, and Risk-Constrained Frontier ($\mathbb{Z}_{113}$ and $S_5$, Pure FP32 Post-Trigger).}
\textbf{Panel A}: Triggered seed counts (plus full-cohort $\mathrm{ITT}_{12}$ on 2L $S_5$ $q_{\rm trig}=0.95$), endogenous budget $S_H^{\rm CL} = \sum_{k=0}^{H-1} \eta_k$, and Max DD ($\text{Mean} \pm \text{SEM}, \text{ddof}=1$) under Std, AMSGrad-Only, and $\perp$AdamW \citep{Prieto2025_250104697} at $\eta_0=10^{-3}$, alongside paired DD/TermAcc (\%) for Micro-Step ($S_H=0.02004$), CL-LR Only, and Closed-Loop.
\textbf{Panel B}: Matched-budget open-loop schedules ($S_H = \bar{S}_H^{\rm CL}$) and Risk-Constrained Terminal Accuracy $\mathrm{TermAcc}_{\rm RC}(\tau_{\rm DD})$ (returning $A_{\rm test}(t_{\rm trig}+H)$ if $\mathrm{MaxDD} \le \tau_{\rm DD}$ and reverting to $A_{\rm test}(t_{\rm trig})$ otherwise).}
\label{tab:app_closed_loop_11suites}
\begin{tabular*}{\textwidth}{@{\extracolsep{\fill}}l@{\hspace{2.0pt}}c@{\hspace{2.0pt}}c@{\hspace{2.0pt}}c@{\hspace{2.0pt}}c@{\hspace{2.0pt}}c@{\hspace{2.0pt}}c@{\hspace{2.0pt}}c@{}}
\toprule
\multicolumn{8}{@{}l@{}}{\textbf{Panel A: Eleven-Suite Controller Factorial, $\perp$AdamW, and Full-Cohort $\mathrm{ITT}_{12}$ Accounting}} \\
\midrule
\textbf{Suite (Trig.)} & \textbf{$S_H^{\rm CL}$} & \textbf{Std} & \textbf{AMSGrad-Only} & \textbf{$\perp$AdamW} & \textbf{Micro-Step} & \textbf{CL-LR Only} & \textbf{Closed-Loop} \\
\midrule
\multicolumn{8}{@{}l@{}}{\textit{High-Readiness Saturated Triggers ($q_{\rm trig} = 0.95$): Stability \& Relapse Suppression}} \\
1L $\mathbb{Z}_{113}$ $2a{+}b$ ($12/12$) & $0.0083$ & $22.06 {\pm} 2.40$ & $7.75 {\pm} 3.70$ & $4.08 {\pm} 1.34$ & $0.08/\mathbf{99.58}$ & $0.08/99.51$ & $\mathbf{0.04}/99.46$ \\
1L $\mathbb{Z}_{113}$ $3a{+}b$ ($12/12$) & $0.0162$ & $28.56 {\pm} 2.40$ & $23.45 {\pm} 8.58$ & $6.48 {\pm} 2.07$ & $0.08/\mathbf{99.19}$ & $0.83/98.52$ & $\mathbf{0.05}/99.09$ \\
2L $\mathbb{Z}_{113}$ $2a{+}b$ ($12/12$) & $0.0291$ & $80.58 {\pm} 1.86$ & $88.16 {\pm} 1.47$ & $59.35 {\pm} 2.97$ & $0.13/98.43$ & $0.50/\mathbf{98.52}$ & $\mathbf{0.12}/98.23$ \\
3L $\mathbb{Z}_{113}$ $2a{+}b$ ($12/12$) & $0.0016$ & $75.99 {\pm} 1.19$ & $82.12 {\pm} 1.03$ & $27.66 {\pm} 4.81$ & $\mathbf{0.01}/\mathbf{99.96}$ & $\mathbf{0.01}/99.94$ & $\mathbf{0.01}/99.92$ \\
2L $S_5$ $ab^{-1}$ ($12/12$) & $0.0079$ & $76.59 {\pm} 5.47$ & $75.19 {\pm} 4.78$ & $73.97 {\pm} 2.12$ & $0.02/99.82$ & $0.02/\mathbf{99.84}$ & $\mathbf{0.01}/99.79$ \\
2L $S_5$ $[a,b,a]$ $aba^{-1}$ ($9/12$) & $0.0510$ & $35.90 {\pm} 11.44$ & $75.28 {\pm} 5.94$ & $17.90 {\pm} 3.53$ & $\mathbf{0.02}/98.63$ & $0.05/\mathbf{99.51}$ & $0.04/98.47$ \\
2L $S_5$ $[a,b,a]$ $\mathrm{ITT}_{12}$ ($12/12$) & $0.0383$ & $26.92 {\pm} 9.66$ & $56.46 {\pm} 10.76$ & $13.42 {\pm} 3.50$ & $\mathbf{0.02}$/$96.58$ & $0.04$/$\mathbf{97.24}$ & $0.03$/$96.46$ \\
\midrule
\multicolumn{8}{@{}l@{}}{\textit{Pre-Saturation Triggers ($q_{\rm trig} = 0.70$): Continued Climb Beyond Truncation}} \\
1L $S_5$ $aba^{-1}$ ($8/12$) & $0.6426$ & $11.85 {\pm} 1.44$ & $18.85 {\pm} 1.65$ & $65.31 {\pm} 2.09$ & $\mathbf{0.09}/77.34$ & $8.12/77.55$ & $0.70/\mathbf{79.56}$ \\
1L $S_5$ $aba^{-1}$ (45\%) ($6/6$) & $0.5863$ & $3.72 {\pm} 0.32$ & $4.65 {\pm} 0.59$ & $69.08 {\pm} 2.96$ & $\mathbf{0.12}/77.85$ & $2.98/80.59$ & $1.22/\mathbf{82.51}$ \\
2L $S_5$ $[a,b]$ $aba^{-1}$ (45\%) ($4/6$) & $0.6309$ & $4.22 {\pm} 0.94$ & $5.32 {\pm} 1.03$ & $58.66 {\pm} 5.91$ & $\mathbf{0.30}/77.80$ & $4.57/77.01$ & $2.04/\mathbf{79.80}$ \\
2L $S_5$ $ab^{-1}$ ($12/12$) & $0.0501$ & $73.82 {\pm} 6.28$ & $73.76 {\pm} 4.72$ & $73.53 {\pm} 2.75$ & $\mathbf{0.05}/97.42$ & $0.11/\mathbf{99.34}$ & $0.09/99.09$ \\
2L $S_5$ $[a,b,a]$ $aba^{-1}$ ($12/12$) & $0.4682$ & $12.55 {\pm} 3.57$ & $72.21 {\pm} 4.17$ & $12.35 {\pm} 1.68$ & $\mathbf{0.01}/81.69$ & $1.08/\mathbf{93.49}$ & $0.40/87.06$ \\
\bottomrule
\end{tabular*}
\vspace{2pt}
\begin{tabular*}{\textwidth}{@{\extracolsep{\fill}}llcccccc@{}}
\multicolumn{8}{@{}l@{}}{\textbf{Panel B: Matched-Budget Open-Loop Schedules ($S_H = \bar{S}_H^{\rm CL}$) and Risk-Constrained Frontier $\mathrm{TermAcc}_{\rm RC}(\tau_{\rm DD})$ ($q_{\rm trig} = 0.70$)}} \\
\midrule
\textbf{Suite ($n$)} & \textbf{Schedule} & \textbf{$S_H$} & \textbf{Max DD (\%)} & \textbf{Worst} & \textbf{Raw Term (\%)} & \textbf{$\le 1\%$} & \textbf{$\mathrm{TermAcc}_{\rm RC}$ ($0.5/1.0/2.0\%$)} \\
\midrule
2L $S_5$ $[a,b,a]$ ($12$) & Snapshot ($t_{\rm trig}$) & $0.0000$ & $0.00 {\pm} 0.00$ & $0.00$ & $72.78 {\pm} 0.78$ & 12/12 & $72.78$/$72.78 {\pm} 0.78$/$72.78$ \\
 & Micro-Step & $0.0200$ & $0.01 {\pm} 0.00$ & $0.03$ & $81.69 {\pm} 0.90$ & 12/12 & $81.69$/$81.69 {\pm} 0.90$/$81.69$ \\
 & OL-Front & $0.4682$ & $4.03 {\pm} 0.35$ & $6.92$ & $80.45 {\pm} 1.16$ & 0/12 & $72.78$/$72.78 {\pm} 0.78$/$72.78$ \\
 & OL-Const & $0.4682$ & $4.06 {\pm} 1.93$ & $19.64$ & $91.43 {\pm} 2.00$ & 8/12 & $87.26$/$87.26 {\pm} 3.57$/$87.26$ \\
 & CL-LR Only & $0.2874$ & $1.08 {\pm} 0.89$ & $10.83$ & $93.49 {\pm} 1.24$ & 11/12 & $91.03$/$92.28 {\pm} 2.10$/$92.28$ \\
 & \textbf{Closed-Loop} & $0.4682$ & $\mathbf{0.40 {\pm} 0.06}$ & $\mathbf{0.83}$ & $87.06 {\pm} 0.94$ & \textbf{12/12} & $83.40$/$\mathbf{87.06 {\pm} 0.94}$/$87.06$ \\
\midrule
1L $S_5$ $aba^{-1}$ ($8$) & Micro-Step & $0.0200$ & $0.09 {\pm} 0.03$ & $0.22$ & $77.34 {\pm} 0.88$ & 8/8 & $77.34$/$77.34 {\pm} 0.88$/$77.34$ \\
 & OL-Front & $0.6426$ & $10.96 {\pm} 1.61$ & $19.11$ & $67.86 {\pm} 2.03$ & 0/8 & $72.01$/$72.01 {\pm} 0.73$/$72.01$ \\
 & OL-Const & $0.6426$ & $6.83 {\pm} 0.63$ & $9.22$ & $77.56 {\pm} 0.94$ & 0/8 & $72.01$/$72.01 {\pm} 0.73$/$72.01$ \\
 & CL-LR Only & $0.6861$ & $8.12 {\pm} 0.63$ & $10.82$ & $77.55 {\pm} 0.82$ & 0/8 & $72.01$/$72.01 {\pm} 0.73$/$72.01$ \\
 & \textbf{Closed-Loop} & $0.6426$ & $\mathbf{0.70 {\pm} 0.10}$ & $\mathbf{0.99}$ & $79.56 {\pm} 0.86$ & \textbf{8/8} & $75.14$/$\mathbf{79.56 {\pm} 0.86}$/$79.56$ \\
\bottomrule
\end{tabular*}
\end{table*}

\begin{table*}[!htbp]
\centering
\scriptsize
\setlength{\tabcolsep}{2.8pt}
\renewcommand{\arraystretch}{0.88}
\caption{\textbf{Nine-Arm Post-Trigger Prefix-Pinning and Consolidation Dissection on 3-Layer $\mathbb{Z}_{113}$ Battery 9 ($2a+b$, $N=12$ paired seeds, $t_{\rm trig} = 183.3 \pm 7.1, H=1{,}500$, $623{,}232$ Params).}
Tests Definition~\ref{def:interface_drift} at $L=3$ with budget $S_H = \sum_{k=0}^{H-1} \eta_k$ and path length $\mathcal{P}_H = \sum_{k=0}^{H-1} \|\theta_{t_{\rm trig}+k+1} - \theta_{t_{\rm trig}+k}\|_2$ (terminal validation-logit RMS drift: $\|z_{\rm val}(t_{\rm trig}+H) - z_{\rm val}(t_{\rm trig})\|_F / \sqrt{|\mathcal{D}_{\rm val}|V}$). Pinning $(E, U, B_0)$ zeroes $\delta_{h_0}$ ($31.26\% \to 0.32\%$ DD), and pinning $(E, U, B_0, B_1)$ zeroes $(\delta_{h_0}, \delta_{h_1})$ ($0.04\% \pm 0.01\%$ DD; $-0.28 \pm 0.19$ pp vs.\ Pin $E{+}U{+}B_0^*$, $p_{\rm Holm}=0.0469$).}
\label{tab:app_cohort9_3l}
\begin{tabular*}{\textwidth}{@{\extracolsep{\fill}}l@{\hspace{2.5pt}}c@{\hspace{2.5pt}}c@{\hspace{2.5pt}}c@{\hspace{2.5pt}}c@{\hspace{2.5pt}}c@{\hspace{2.5pt}}c@{\hspace{2.5pt}}c@{}}
\toprule
\textbf{Policy ($L=3$)} & \textbf{$S_H$} & \textbf{Max DD (\%)} & \textbf{Min (\%)} & \textbf{Term (\%)} & \textbf{$\delta_{h_0}$} & \textbf{$\delta_{h_1}$} & \textbf{$\mathcal{P}_H$} \\
\midrule
Snapshot ($t_{\rm trig}$) & $0.000$ & $0.00 \pm 0.00$ & $99.90 \pm 0.05$ & $99.90 \pm 0.05$ & $0.000$ & $0.000$ & $0.00 \pm 0.00$ \\
Std ($\eta_0{=}10^{-3}$) & $1.500$ & $75.99 \pm 1.19$ & $24.00 \pm 1.19$ & $94.86 \pm 5.03$ & $0.984 \pm 0.003$ & $0.957 \pm 0.004$ & $89.94 \pm 3.57$ \\
AMSGrad-Only & $1.500$ & $82.12 \pm 1.03$ & $17.88 \pm 1.03$ & $91.90 \pm 5.30$ & $0.988 \pm 0.002$ & $0.980 \pm 0.002$ & $64.86 \pm 2.08$ \\
$\perp$AdamW ($\eta_0{=}10^{-3}$) & $1.500$ & $27.66 \pm 4.81$ & $72.31 \pm 4.83$ & $97.66 \pm 0.61$ & $0.985 \pm 0.001$ & $0.979 \pm 0.002$ & $145.49 \pm 1.59$ \\
Frz $E{+}U$ & $1.500$ & $31.26 \pm 3.87$ & $68.73 \pm 3.87$ & $96.63 \pm 2.21$ & $0.951 \pm 0.003$ & $0.940 \pm 0.004$ & $104.51 \pm 1.60$ \\
Pin $E{+}U{+}B_0^*$ & $1.500$ & $0.32 \pm 0.20$ & $99.63 \pm 0.21$ & $99.92 \pm 0.04$ & $\mathbf{0.000}$ & $0.788 \pm 0.014$ & $101.79 \pm 0.90$ \\
Pin $E{+}U{+}B_{0:1}^*$ & $1.500$ & $0.04 \pm 0.01$ & $99.88 \pm 0.05$ & $99.93 \pm 0.04$ & $\mathbf{0.000}$ & $\mathbf{0.000}$ & $89.36 \pm 1.33$ \\
Micro-Step & $0.020$ & $\mathbf{0.01 \pm 0.00}$ & $\mathbf{99.90 \pm 0.05}$ & $\mathbf{99.96 \pm 0.02}$ & $0.133 \pm 0.005$ & $0.169 \pm 0.008$ & $1.57 \pm 0.04$ \\
CL-LR Only & $0.0012$ & $\mathbf{0.01 \pm 0.00}$ & $\mathbf{99.90 \pm 0.05}$ & $99.94 \pm 0.03$ & $0.027 \pm 0.011$ & $0.038 \pm 0.014$ & $0.27 \pm 0.12$ \\
Closed-Loop & $0.0016$ & $\mathbf{0.01 \pm 0.00}$ & $\mathbf{99.90 \pm 0.05}$ & $99.92 \pm 0.04$ & $0.015 \pm 0.005$ & $0.023 \pm 0.008$ & $\mathbf{0.08 \pm 0.03}$ \\
\bottomrule
\end{tabular*}
\end{table*}

\subsection{Non-Abelian \texorpdfstring{$S_5$}{S5} Compositional Factorial, Conjugacy-Class, and Iso-DoF Circuit Details}
\label{app:s5_composition_details}
To isolate why two-token bidirectional models plateau on ternary conjugation $a \circ b \circ a^{-1}$ in $S_5$ ($|S_5|=120$), we first ran a validation-only development study across $N=3$ paired seeds (\texttt{500}--\texttt{502}; Table~\ref{tab:app_s5_composition} Panel A) before locking Batteries 10--11 (Tables~\ref{tab:app_s5_composition}--\ref{tab:app_s5_isodof_pw}). Because $\mathrm{Cl}(aba^{-1}) = \mathrm{Cl}(b)$ partitions $S_5$ into seven cycle-type classes, uniform intra-class prediction yields a $7/120 \approx 5.83\%$ baseline:
\begin{equation}
\label{eq:s5_unary_floor}
\sum_{C \in \mathrm{Cl}(S_5)} |C| = 1 + 10 + 15 + 20 + 20 + 24 + 30 = 120, \qquad
\sum_{C \in \mathrm{Cl}(S_5)} \frac{|C|}{120} \cdot \frac{1}{|C|} = \frac{7}{120} \approx 5.83\%.
\end{equation}
At $d_{\rm model}=128$ ($4{\times}32$), bidirectional $T=2$ $[a,b]$ Full transfer reaches $99.38\%$ cycle-type accuracy and $33.90\% \pm 6.21\%$ exact accuracy. Scrambling $a \mapsto a'$ drops exact accuracy by $-28.52 \pm 6.12$ pp to $5.38\% \pm 0.09\%$ while preserving $99.56\% \pm 0.12\%$ cycle-type accuracy ($87.90\%$ fixed-$b$ intra-class diversity). Corrupting $h_0$ at slot $0$ or slot $1$ under causal $[a,b,a]$ degrades accuracy by $-9.82 \pm 2.54$ pp and $-7.64 \pm 2.27$ pp at $4{\times}32$, and by $-19.75 \pm 3.78$ pp and $-7.37 \pm 2.93$ pp at $8{\times}32$ ($d_{\rm model}=256$). Scaling $d_{\rm model}: 128 \to 256$ also lifts $\mathrm{erank}(W_E)$ from $71.16 \pm 3.23$ to $81.85 \pm 4.78$, $\mathrm{erank}(h_0)$ from $87.38 \pm 3.15$ to $105.89 \pm 2.25$, and $\mathrm{erank}(h_1)$ from $71.32 \pm 2.42$ to $89.17 \pm 2.42$. Across all 15 Iso-DoF suites (Table~\ref{tab:app_s5_isodof_pw}), ridge-calibrated Peter--Weyl probes at slot $1$ decode $a\circ b$ near chance ($0.48\%$--$6.64\%$). Across the ten grokked ($\ge 85\%$) suites, slot $2$ refines $aba^{-1}$ directly from $h_0^{\rm slot2}$ ($68.98\%$--$91.92\%$) to $h_1^{\rm slot2}$ ($80.94\%$--$97.50\%$). Although $W_E$ irrep energy is near-isotropic in both $2{\times}64$ Cold ($28.5\%/27.8\%/27.1\%$ across $\wedge^2 V_{\rm std}, \mathrm{std}_4, \mathrm{Specht}_5$) and $2{\times}64$ $E{+}U$ ($30.7\%/25.2\%/25.3\%$), scratch co-adapts $\{W_E, B_{0:1}, W_U\}$ into a shared gauge ($\mathrm{LN}_f$ $R^2$: $\mathrm{std}_4\ 0.285, \mathrm{sgn}\ 0.972$). In contrast, $E{+}U$ requires $d_{\rm mlp}=1{,}536$ ($953\text{k}$) to align blocks onto the fixed donor chart ($\mathrm{std}_4\ 0.322, \mathrm{std{\cdot}sgn}_4\ 0.248$).

\begin{table*}[!htbp]
\centering
\scriptsize
\setlength{\tabcolsep}{2.8pt}
\renewcommand{\arraystretch}{0.84}
\caption{\textbf{Non-Abelian $S_5$ Compositional Development ($N=3$ dev seeds), Held-Out 7-Conjugacy-Class Test Decomposition, and Mechanistic Counterfactual / Effective-Rank Diagnostics ($N=12$ paired seeds, Pure FP32).}
\textbf{Panel A}: Validation-only development factorial ($N=3$ paired dev seeds, $30\text{k}$ steps, $30\%$ train) isolating 3-token $[a,b,a]$, causal prefix masking, and width scaling ($d=256$ vs.\ reference scale $|S_5|=\sum_\lambda d_\lambda^2 = 120$).
\textbf{Panel B}: Held-out test accuracy across all 7 $S_5$ conjugacy classes at peak-validation checkpoints (including $d=128$ crossover arms X1--X3; secondary trajectory-maximum accuracies $\max_t A_{\rm test}(t)$ across the ten $30\text{k}$-step $S_5$ arms in Table~\ref{tab:depth_s5_acquisition} are $34.48 \pm 6.19\%$ for $[a,b]$ Bidir $d{=}128$, $57.82 \pm 7.41\%$ for $[a,b]$ Caus.\ X3, $40.70 \pm 8.96\%$ for $[a,b,a]$ Bidir X1, $51.13 \pm 7.99\%$ for $[a,b,a]$ Caus.\ X2, $43.64 \pm 10.64\%$ for $[a,b,a]$ Caus.\ Cold $d{=}128$, $80.08 \pm 7.81\%$ for $[a,b,a]$ Caus.\ Full $d{=}128$, $42.02 \pm 13.10\%$ for $[a,b,a]$ Caus.\ $U{+}B$ $d{=}256$, $78.35 \pm 9.54\%$ for $[a,b,a]$ Caus.\ $E{+}U$ $d{=}256$, $92.86 \pm 3.32\%$ for $[a,b,a]$ Caus.\ Cold $d{=}256$, and $97.97 \pm 1.01\%$ for $[a,b,a]$ Caus.\ Full $d{=}256$).
\textbf{Panel C}: Counterfactual $a$-scrambling ($a \mapsto a'$), fixed-$b$ intra-class prediction diversity ratio, linear probe decoding of $a\circ b$ at slot 1 vs.\ $aba^{-1}$ at the final slot, and effective rank ($\mathrm{erank}$) of $W_E, h_0^{\rm final}, h_1^{\rm final}$.}
\label{tab:app_s5_composition}
\begin{tabular*}{\textwidth}{@{\extracolsep{\fill}}lcccc@{}}
\toprule
\multicolumn{5}{@{}l@{}}{\textbf{Panel A: Zero-Test-Touch Development Factorial ($N=3$ paired dev seeds, Validation Metrics Only)}} \\
\midrule
\textbf{Factorial Cell} & \textbf{Cold MaxVal (\%)} & \textbf{Full MaxVal (\%)} & \textbf{Full Cycle (\%)} & \textbf{Full Hard3 (\%)} \\
\midrule
$T{=}2\ [a,b]$ Bidir ($d{=}128, 427\text{k}$) & $5.23 \pm 0.17$ & $31.01 \pm 18.54$ & $99.20\%$ & $33.83 \pm 17.50$ \\
$T{=}2\ [b,a]$ Bidir ($d{=}128, 427\text{k}$) & $5.31 \pm 0.09$ & $30.25 \pm 15.27$ & $99.49\%$ & $35.44 \pm 15.53$ \\
$T{=}3\ [a,b,a]$ Bidir ($d{=}128, 427\text{k}$) & $5.38 \pm 0.28$ & $34.21 \pm 18.80$ & $99.54\%$ & $38.77 \pm 18.62$ \\
$T{=}3\ [a,b,a]$ Causal ($d{=}128, 427\text{k}$) & $54.35 \pm 24.07$ & $40.43 \pm 17.78$ & $99.13\%$ & $36.50 \pm 21.04$ \\
$T{=}2\ [a,b]$ Bidir ($d{=}256, 1.64\text{M}$) & $5.28 \pm 0.06$ & $68.25 \pm 31.21$ & $98.99\%$ & $67.93 \pm 31.56$ \\
\textbf{$T{=}3\ [a,b,a]$ Causal ($d{=}256, 1.64\text{M}$)} & $\mathbf{94.79 \pm 2.19}$ & $\mathbf{95.34 \pm 2.43}$ & $\mathbf{100.00\%}$ & $\mathbf{97.58 \pm 2.36}$ \\
\bottomrule
\end{tabular*}
\vspace{2pt}
\begin{tabular*}{\textwidth}{@{\extracolsep{\fill}}lccccccccc@{}}
\multicolumn{10}{@{}l@{}}{\textbf{Panel B: Held-Out Test Decomposition Across All 7 Conjugacy Classes of $S_5$ ($|S_5|=120$)}} \\
\midrule
\textbf{Model / Arm} & \textbf{Exact (\%)} & \textbf{Cycle} & \textbf{$\mathrm{id}$} & \textbf{$(1^3,2)$} & \textbf{$(1,2^2)$} & \textbf{$(1^2,3)$} & \textbf{$(2,3)$} & \textbf{$(5)$} & \textbf{$(1,4)$} \\
\midrule
C1: $T{=}2\ [a,b]$ Bidir ($d{=}128$) & $33.90 \pm 6.21$ & $99.38\%$ & $100.00$ & $58.58$ & $54.65$ & $30.63$ & $28.65$ & $21.87$ & $28.46$ \\
X1: $[a,b,a]$ Bidir ($d{=}128$, Caus.\ D.) & $40.11 \pm 8.99$ & $98.79\%$ & $99.88$ & $54.93$ & $43.52$ & $43.65$ & $41.53$ & $28.85$ & $37.28$ \\
X2: $[a,b,a]$ Caus.\ ($d{=}128$, Bidir D.) & $50.43 \pm 8.02$ & $99.67\%$ & $99.88$ & $63.49$ & $69.00$ & $50.99$ & $48.21$ & $33.60$ & $49.92$ \\
X3: $T{=}2\ [a,b]$ Caus.\ ($d{=}128$) & $57.24 \pm 7.41$ & $99.71\%$ & $100.00$ & $72.20$ & $58.12$ & $61.49$ & $61.39$ & $43.85$ & $55.38$ \\
8D1: $[a,b]$ Bidir (45\%, $n{=}4$) & $70.11 \pm 0.47$ & $99.98\%$ & $100.00$ & $98.47$ & $68.15$ & $84.81$ & $81.45$ & $41.45$ & $66.57$ \\
C4: $[a,b,a]$ Caus.\ Cold ($d{=}128$) & $43.33 \pm 10.69$ & $99.02\%$ & $99.88$ & $61.90$ & $53.21$ & $43.51$ & $42.68$ & $31.40$ & $40.15$ \\
C4: $[a,b,a]$ Caus.\ Full ($d{=}128$) & $79.98 \pm 7.83$ & $99.77\%$ & $100.00$ & $87.64$ & $76.38$ & $85.84$ & $82.54$ & $68.91$ & $81.92$ \\
C6: $[a,b,a]$ Caus.\ Cold ($d{=}256$) & $92.68 \pm 3.36$ & $99.94\%$ & $100.00$ & $99.61$ & $\mathbf{98.04}$ & $94.93$ & $95.64$ & $79.75$ & $94.44$ \\
\textbf{C6: $[a,b,a]$ Caus.\ Full ($d{=}256$)} & $\mathbf{97.89 \pm 1.02}$ & $\mathbf{99.99\%}$ & $100.00$ & $\mathbf{99.93}$ & $95.09$ & $\mathbf{99.62}$ & $\mathbf{99.63}$ & $\mathbf{93.71}$ & $\mathbf{99.57}$ \\
\bottomrule
\end{tabular*}
\vspace{2pt}
\begin{tabular*}{\textwidth}{@{\extracolsep{\fill}}lccccccc@{}}
\multicolumn{8}{@{}l@{}}{\textbf{Panel C: Counterfactual $a$-Scrambling, Fixed-$b$ Diversity, Linear Probing, and Effective Rank ($N=12$)}} \\
\midrule
\textbf{Model / Arm} & \textbf{$a$-Scram} & \textbf{Drop (pp)} & \textbf{Div\%} & \textbf{Prb S1/S2} & \textbf{$\mathrm{erank}(W_E)$} & \textbf{$\mathrm{erank}(h_0)$} & \textbf{$\mathrm{erank}(h_1)$} \\
\midrule
C1: $[a,b]$ Bidir ($128$) & $5.38{\pm}0.09$ & $-28.52{\pm}6.12$ & $87.90$ & $0.43$/$32.30$ & $77.2$ & $86.8{\pm}7.2$ & $74.2{\pm}5.9$ \\
X1: $[a,b,a]$ Bidir ($128$, C.D.) & $5.38{\pm}0.10$ & $-34.74{\pm}8.91$ & $96.21$ & $0.41$/$38.83$ & $80.7$ & $97.8{\pm}1.8$ & $83.3{\pm}1.4$ \\
X2: $[a,b,a]$ Caus.\ ($128$, B.D.) & $5.44{\pm}0.09$ & $-44.99{\pm}7.94$ & $88.35$ & $0.42$/$47.84$ & $75.5$ & $85.3{\pm}7.8$ & $70.7{\pm}5.9$ \\
X3: $[a,b]$ Caus.\ ($128$, C.D.) & $5.53{\pm}0.09$ & $-51.71{\pm}7.33$ & $97.13$ & $0.31$/$54.70$ & $76.6$ & $93.5{\pm}2.1$ & $78.8{\pm}1.8$ \\
C4: $[a,b,a]$ Caus.\ ($128$, C.D.) & $5.65{\pm}0.07$ & $-74.33{\pm}7.78$ & $99.05$ & $0.39$/$77.82$ & $71.2$ & $87.4{\pm}3.2$ & $71.3{\pm}2.4$ \\
\textbf{C6: $[a,b,a]$ Caus.\ ($256$)} & $5.79{\pm}0.05$ & $-92.10{\pm}1.01$ & $99.96$ & $0.51$/$97.55$ & $81.8$ & $105.9{\pm}2.3$ & $89.2{\pm}2.4$ \\
\bottomrule
\end{tabular*}
\end{table*}

\begin{table*}[!htbp]
\centering
\scriptsize
\setlength{\tabcolsep}{2.8pt}
\renewcommand{\arraystretch}{0.82}
\caption{\textbf{Iso-DoF 2-Layer $S_5$ Architecture Grid ($427\text{k}$--$1.64\text{M}$ Parameters, $d_m := d_{\rm mlp}$) and 15-Suite Ridge-Calibrated Two-Stage Peter--Weyl Circuit Decoding ($N=12$ paired seeds, Pure FP32).}
Columns 5--8 report held-out exact Peter--Weyl decoding accuracy (\%) at intermediate slot 1 ($h_0^{\rm slot1} \to a\circ b$) versus final readout slot 2 ($h_0^{\rm slot2}, h_1^{\rm slot2}, \mathrm{LN}_f \to aba^{-1}$) under cross-validated ridge regularization ($\lambda \in \{10^{-4}, \dots, 10^2\}$). Across all 15 suites, $h_0^{\rm slot1}$ yields near-chance $a\circ b$ decoding ($0.48\%$--$6.64\%$, $1/120=0.83\%$), whereas across the ten grokked ($\ge 85\%$) suites $h_0^{\rm slot2} \to h_1^{\rm slot2}$ directly refines $aba^{-1}$ from $68.98\%$--$91.92\%$ to $80.94\%$--$97.50\%$.}
\label{tab:app_s5_isodof_pw}
\begin{tabular*}{\textwidth}{@{\extracolsep{\fill}}lccccccc@{}}
\toprule
\textbf{Cell / Condition} & \textbf{Trig} & \textbf{$I_{\rm full}$ ($\uparrow$)} & \textbf{Val-Sel.\ (\%)} & \textbf{$h_0^{\rm s1}(ab)$} & \textbf{$h_0^{\rm s2}$ (\%)} & \textbf{$h_1^{\rm s2}$ (\%)} & \textbf{$\mathrm{LN}_f$} \\
\midrule
$4{\times}32, d_m{=}512$ ($427{,}008$) Cold & $3,0$ & $3{,}855.0{\pm}1{,}038.6$ & $43.33{\pm}10.69$ & $0.59$ & $30.58{\pm}8.15$ & $39.20{\pm}9.70$ & $35.78$ \\
$4{\times}32, d_m{=}512$ ($427{,}008$) Full & $9,5$ & $17{,}745.1{\pm}2{,}141.1$ & $79.98{\pm}7.83$ & $0.84$ & $55.14{\pm}6.93$ & $77.14{\pm}7.84$ & $76.52$ \\
$2{\times}64, d_m{=}512$ ($427{,}008$) Cold & $12,11$ & $15{,}715.5{\pm}846.0$ & $98.54{\pm}0.70$ & $0.71$ & $77.37{\pm}1.65$ & $93.12{\pm}1.03$ & $92.05$ \\
$2{\times}64, d_m{=}512$ ($427{,}008$) $U{+}B$ & $2,1$ & $4{,}498.8{\pm}1{,}889.4$ & $25.02{\pm}9.57$ & $0.48$ & $18.16{\pm}8.15$ & $24.10{\pm}9.30$ & $22.25$ \\
$2{\times}64, d_m{=}512$ ($427{,}008$) $E{+}U$ & $10,5$ & $16{,}867.0{\pm}1{,}804.1$ & $85.95{\pm}5.67$ & $0.68$ & $69.25{\pm}5.13$ & $80.94{\pm}5.73$ & $80.01$ \\
$2{\times}64, d_m{=}512$ ($427{,}008$) Full & $8,3$ & $14{,}337.8{\pm}2{,}417.8$ & $66.08{\pm}10.97$ & $0.64$ & $30.88{\pm}5.91$ & $59.42{\pm}9.82$ & $57.87$ \\
$4{\times}32, d_m{=}1536$ ($953{,}344$) Cold & $11,4$ & $8{,}703.7{\pm}980.2$ & $88.30{\pm}3.77$ & $0.66$ & $68.98{\pm}4.51$ & $82.51{\pm}3.62$ & $82.33$ \\
$4{\times}32, d_m{=}1536$ ($953{,}344$) Full & $12,6$ & $22{,}562.2{\pm}672.3$ & $93.42{\pm}1.94$ & $0.71$ & $72.20{\pm}2.29$ & $90.71{\pm}2.08$ & $90.88$ \\
$8{\times}32, d_m{=}354$ ($953{,}028$) Cold & $4,1$ & $3{,}711.7{\pm}821.0$ & $46.89{\pm}10.37$ & $0.51$ & $37.84{\pm}8.62$ & $47.01{\pm}10.36$ & $46.33$ \\
$8{\times}32, d_m{=}354$ ($953{,}028$) Full & $12,8$ & $21{,}300.8{\pm}1{,}689.5$ & $94.45{\pm}1.85$ & $5.84$ & $74.29{\pm}6.67$ & $92.59{\pm}2.43$ & $91.54$ \\
$2{\times}64, d_m{=}1536$ ($953{,}344$) Cold & $12,5$ & $17{,}527.1{\pm}576.1$ & $94.37{\pm}1.78$ & $0.73$ & $75.90{\pm}2.68$ & $88.93{\pm}1.49$ & $87.53$ \\
$2{\times}64, d_m{=}1536$ ($953{,}344$) $E{+}U$ & $12,9$ & $21{,}515.2{\pm}585.6$ & $96.97{\pm}0.95$ & $0.70$ & $79.13{\pm}2.11$ & $92.37{\pm}0.74$ & $92.46$ \\
$2{\times}64, d_m{=}1536$ ($953{,}344$) Full & $12,6$ & $21{,}424.8{\pm}934.6$ & $95.02{\pm}1.57$ & $0.71$ & $71.31{\pm}4.17$ & $92.06{\pm}1.96$ & $92.32$ \\
$8{\times}32, d_m{=}1024$ ($1{,}640{,}448$) Cold & $11,6$ & $12{,}064.7{\pm}1{,}081.1$ & $92.68{\pm}3.36$ & $0.63$ & $81.88{\pm}5.29$ & $91.98{\pm}3.56$ & $91.94$ \\
$8{\times}32, d_m{=}1024$ ($1{,}640{,}448$) Full & $12,9$ & $23{,}784.5{\pm}1{,}019.1$ & $97.89{\pm}1.02$ & $6.64$ & $91.92{\pm}2.62$ & $97.50{\pm}1.08$ & $97.50$ \\
\bottomrule
\end{tabular*}
\end{table*}

\begin{table*}[!htbp]
\centering
\scriptsize
\setlength{\tabcolsep}{2.8pt}
\renewcommand{\arraystretch}{0.82}
\caption{\textbf{Ten-Arm Post-Trigger Consolidation and Zero-Leakage Validation-Gated $\mathrm{ValRC}(\tau)$ Dissection Across Regimes A, B, and C ($N=12$ paired seeds, Pure FP32).}
On Regime A ($\mathbb{Z}_{113}$ pooled $48$ branches), PGL-Consol achieves $0.03\%$ mean DD ($0.11\%$ worst seed, $48/48$ val-feasible, $99.22\%$ $\mathrm{ValRC}(1\%)$ vs.\ $39.40\%$ Std DD); on 2L $S_5\ q_{\rm trig}=0.95$ macro, PGL-Consol reaches $98.70\% \pm 0.17\%$ $\mathrm{ValRC}(1\%)$ ($+0.47 \pm 0.07$ pp vs.\ Closed-Loop, $p_{{\rm Holm},8}=0.00391$; Std trig.\ Term $86.50\% \pm 3.84\%$, $\perp$AdamW $40.89\% \pm 3.91\%$). Table rows below detail the 5-suite 2L $S_5\ q_{\rm trig}=0.70$ Regime B macro ($58$ triggered branches; Holm $M=8$, $^{**}p_{\rm Holm}<0.01$; secondary Micro-Step shows unadjusted $p$).}
\label{tab:app_pgl_valrc}
\begin{tabular*}{\textwidth}{@{\extracolsep{\fill}}lccccccc@{}}
\toprule
\textbf{Controller Arm} & \textbf{$S_H$} & \textbf{DD\%} & \textbf{Term\%} & \textbf{RC($0.5\%$)} & \textbf{RC($1.0\%$)} & \textbf{ITT RC} & \textbf{$\Delta$RC (pp)} \\
\midrule
Std ($\eta_0, \lambda{=}1$) & $3.000$ & $20.48{\pm}2.75$ & $84.71{\pm}1.88$ & $73.30$ & $73.30{\pm}0.50$ & $72.36{\pm}0.71$ & $+11.56{\pm}0.80^{**}$ \\
$\perp$AdamW & $3.000$ & $51.84{\pm}1.22$ & $44.86{\pm}1.16$ & $73.30$ & $73.30{\pm}0.50$ & $72.36{\pm}0.71$ & $+11.56{\pm}0.80^{**}$ \\
CL-LR Only & $0.467$ & $4.86{\pm}0.31$ & $86.32{\pm}0.85$ & $79.12$ & $79.75{\pm}1.17$ & $78.55{\pm}1.21$ & $+5.10{\pm}0.91^{**}$ \\
Cautious CL-LR & $0.448$ & $4.27{\pm}0.42$ & $87.30{\pm}0.91$ & $79.80$ & $80.46{\pm}1.40$ & $79.16{\pm}1.33$ & $+4.39{\pm}0.74^{**}$ \\
Fixed-Leaky & $0.503$ & $3.99{\pm}0.48$ & $85.34{\pm}0.81$ & $77.25$ & $79.44{\pm}1.19$ & $78.20{\pm}1.17$ & $+5.42{\pm}0.81^{**}$ \\
Matched Open-Loop & $0.257$ & $2.75{\pm}0.18$ & $87.14{\pm}0.91$ & $77.50$ & $78.30{\pm}0.75$ & $77.16{\pm}0.90$ & $+6.56{\pm}0.88^{**}$ \\
Micro-Step & $0.020$ & $0.02{\pm}0.01$ & $79.90{\pm}0.54$ & $79.90$ & $79.90{\pm}0.54$ & $78.69{\pm}0.71$ & $+4.95{\pm}0.52$ ($p{=}0.0005$) \\
PGL Gate + $v^{{\max}}$ & $0.210$ & $0.18{\pm}0.01$ & $82.47{\pm}0.54$ & $80.98$ & $82.47{\pm}0.54$ & $81.18{\pm}0.80$ & $+2.39{\pm}0.42^{**}$ \\
Closed-Loop ($v^{{\max}}$) & $0.522$ & $1.06{\pm}0.20$ & $85.34{\pm}0.61$ & $76.99$ & $82.70{\pm}0.92$ & $81.44{\pm}1.19$ & $+2.15{\pm}0.51^{**}$ \\
\textbf{PGL-Consol (Ours)} & $0.258$ & $\mathbf{0.24{\pm}0.02}$ & $84.85{\pm}0.89$ & $80.93$ & $\mathbf{84.85{\pm}0.89}$ & $83.43{\pm}0.93$ & Ref.\ ($58/58$) \\
\bottomrule
\end{tabular*}
\end{table*}

\begin{table*}[!htbp]
\centering
\scriptsize
\setlength{\tabcolsep}{2.8pt}
\renewcommand{\arraystretch}{0.82}
\caption{\textbf{Real-Sensory MNIST Supervised Cluster Separability vs.\ Unsupervised CRT Harmonic Pruning ($\mathrm{Op1}:\ (2y_1+y_2)\bmod 10$) and 10-Arm Consolidation Dissection ($N=12$ paired seeds, $60$ branches).}
\textbf{Panel A} separates supervised $S_{10}$-permutation-invariant metrics (10-class ridge digit probe accuracy and 9-mode Fourier $R^2$) from cyclic-order-sensitive unsupervised digit-centroid DFT energy fractions ($k\in\{1,3\}$ odd modes vs.\ CRT modes $\{k=2,4,5\}$ and terminal $k=5$ parity). \textbf{Panel B} reports the 5-suite zero-retuning consolidation macro ($H=2{,}000, q_{\rm trig}=0.82, \eta_0=1.5\times 10^{-3}$; Holm $M=8$, secondary Micro-Step shows unadjusted $p$).}
\label{tab:app_mnist_full}
\begin{tabular*}{\textwidth}{@{\extracolsep{\fill}}lccccccc@{}}
\toprule
\multicolumn{8}{@{}l@{}}{\textbf{Panel A: Supervised Cluster Separability ($t=0$) vs.\ Unsupervised Centroid DFT CRT Pruning ($\mathrm{Op1}$)}} \\
\midrule
\textbf{Arm (Scale)} & \textbf{Dig$_0$ (\%)} & \textbf{$R^2_{\mathbb{Z}_{10},0}$} & \textbf{Odd$_0$ (\%)} & \textbf{Odd$_{\rm pk}$ (\%)} & \textbf{CRT $\{2,4,5\}$} & \textbf{Std $k{=}5$} & \textbf{PGL $k{=}5$} \\
\midrule
Cold ($1.0\times n_c$) & $88.57{\pm}0.25$ & $0.570{\pm}0.003$ & $29.95{\pm}0.50$ & $5.33{\pm}0.86$ & $94.67{\pm}0.86$ & $8.68{\pm}0.88$ & $15.26{\pm}0.69$ \\
$U{+}B$ (Cold $E_\phi$) & $88.57{\pm}0.25$ & $0.570{\pm}0.003$ & $29.95{\pm}0.50$ & $3.70{\pm}0.80$ & $96.30{\pm}0.80$ & $8.06{\pm}0.53$ & $13.32{\pm}0.65$ \\
$E{+}U$ ($1.0\times n_c$) & $97.26{\pm}0.07$ & $0.808{\pm}0.002$ & $13.59{\pm}1.67$ & $4.10{\pm}0.88$ & $95.90{\pm}0.88$ & $8.87{\pm}0.64$ & $14.85{\pm}0.49$ \\
\textbf{Full ($1.0\times n_c$)} & $97.26{\pm}0.07$ & $0.808{\pm}0.002$ & $13.59{\pm}1.67$ & $5.42{\pm}1.36$ & $94.58{\pm}1.36$ & $8.84{\pm}0.54$ & $18.03{\pm}0.72$ \\
\bottomrule
\end{tabular*}
\vspace{2pt}
\begin{tabular*}{\textwidth}{@{\extracolsep{\fill}}lccccccc@{}}
\multicolumn{8}{@{}l@{}}{\textbf{Panel B: Ten-Arm Post-Trigger Consolidation Macro Across 5 Real-Sensory MNIST Suites ($60$ branches)}} \\
\midrule
\textbf{Controller Arm} & \textbf{$S_H$} & \textbf{Test DD} & \textbf{Val DD} & \textbf{Term\%} & \textbf{RC($0.5\%$)} & \textbf{RC($1.0\%$)} & \textbf{$\Delta$RC($1\%$)} \\
\midrule
Std ($\eta_0, \lambda{=}0.8$) & $3.000$ & $25.02{\pm}3.69$ & $25.00{\pm}3.64$ & $86.96{\pm}0.27$ & $87.55$ & $87.55{\pm}0.18$ & $+2.87{\pm}0.12$ pp$^{**}$ \\
$\perp$AdamW & $3.000$ & $13.66{\pm}0.86$ & $13.88{\pm}0.86$ & $83.66{\pm}0.23$ & $87.55$ & $87.55{\pm}0.18$ & $+2.87{\pm}0.12$ pp$^{**}$ \\
CL-LR Only & $0.302$ & $1.28{\pm}0.11$ & $1.27{\pm}0.11$ & $90.59{\pm}0.14$ & $88.81$ & $89.41{\pm}0.22$ & $+1.01{\pm}0.19$ pp$^{**}$ \\
Cautious CL-LR & $0.300$ & $0.84{\pm}0.08$ & $0.90{\pm}0.10$ & $90.62{\pm}0.13$ & $89.17$ & $89.91{\pm}0.28$ & $+0.52{\pm}0.21$ pp$^{*}$ \\
Fixed-Leaky (No Gate) & $0.318$ & $0.81{\pm}0.06$ & $0.86{\pm}0.07$ & $89.83{\pm}0.15$ & $88.48$ & $89.21{\pm}0.24$ & $+1.21{\pm}0.15$ pp$^{**}$ \\
Matched Open-Loop & $0.061$ & $0.23{\pm}0.02$ & $0.33{\pm}0.02$ & $91.30{\pm}0.12$ & $90.90$ & $91.26{\pm}0.15$ & $-0.84{\pm}0.07$ pp$^{**}$ \\
Micro-Step ($S_H{=}0.02$) & $0.020$ & $0.02{\pm}0.00$ & $0.04{\pm}0.01$ & $90.47{\pm}0.12$ & $90.47$ & $90.47{\pm}0.12$ & $-0.04{\pm}0.02$ ($p{=}0.0317$) \\
PGL Gate + Hard $v^{{\max}}$ & $0.066$ & $0.11{\pm}0.01$ & $0.16{\pm}0.01$ & $90.35{\pm}0.13$ & $90.35$ & $90.35{\pm}0.13$ & $+0.07{\pm}0.01$ pp$^{**}$ \\
Closed-Loop ($v^{{\max}}$) & $0.323$ & $1.03{\pm}0.06$ & $1.11{\pm}0.06$ & $89.44{\pm}0.17$ & $87.87$ & $88.76{\pm}0.26$ & $+1.67{\pm}0.15$ pp$^{**}$ \\
\textbf{PGL-Consol (Ours)} & $0.061$ & $0.09{\pm}0.01$ & $0.14{\pm}0.01$ & $90.42{\pm}0.13$ & $90.42$ & $90.42{\pm}0.13$ & Ref.\ ($60/60$ Feas.) \\
\bottomrule
\end{tabular*}
\end{table*}

\clearpage
\section{Model Architectures, Parameter Census, and Initialization Details}
\label{app:arch_details}

This appendix details the parameter census, state-dictionary keys, and transfer rules across symbolic Batteries 1--12 and real-sensory MNIST Battery 13.

\begin{table*}[!htbp]
\centering
\scriptsize
\setlength{\tabcolsep}{2.8pt}
\renewcommand{\arraystretch}{0.78}
\caption{\textbf{Parameter Census Across Symbolic Pre-LN Transformers ($\mathbb{Z}_{113}, S_5$ Iso-DoF Grid) and Real-Sensory MNIST \texttt{SensoryGrokTransformer}.} Component assignments match the factorial decomposition ($E / E_\phi$: Token Embedding or Conv2d Perceptual Trunk, $U$: Readout Head, $B$: Internal Transformer Blocks; \texttt{F} denotes \texttt{bias=False}). Symbolic donor transfer rescales transferred tensors to $1.0\times$ cold Frobenius norm $n_c$; on MNIST, donor arms ($E_{\rm only}, E{+}U, U{+}B$, Full) use $1.0\times n_c$, while RandOrth matches trained donor norms $n_d$ ($1.2\times$--$5.4\times n_c$).}
\label{tab:model_parameters}
\begin{tabular*}{\textwidth}{@{\extracolsep{\fill}}llcccc@{}}
\toprule
\textbf{Parameter Tensor} & \textbf{Module Type} & \textbf{Shape} & \textbf{Params} & \textbf{Init} & \textbf{Transfer Policy} \\
\midrule
\texttt{token\_emb} & \texttt{Emb(115, 128)} & $[115, 128]$ & $14{,}720$ & $\mathcal{N}(0, 0.02^2)$ & Trans.\ ($E, 1{\times}n_c$) \\
\texttt{pos\_emb} & \texttt{Emb(4, 128)} & $[4, 128]$ & $512$ & $\mathcal{N}(0, 0.02^2)$ & \textbf{Cold} / \textbf{Frob.} ($S_5$) \\
\texttt{blocks.$\ell$.ln1,2} & \texttt{2$\times$LN(128)} & $4{\times}[128]$ & $512$ & $\mathbf{1}, \mathbf{0}$ & \textbf{Cold} / \textbf{Frob.} ($S_5$, MN) \\
\texttt{blocks.$\ell$.attn} & \texttt{Lin(128, 384/128, F)} & $[384/128, 128]$ & $65{,}536$ & $\mathcal{N}(0, 0.02^2)$ & Trans.\ ($B_\ell, 1{\times}n_c$) \\
\texttt{blocks.$\ell$.mlp} & \texttt{Lin(128$\leftrightarrow$512)} & $+640$ bias & $131{,}712$ & $\mathcal{N}(0, 0.02^2)$ & Trans.\ ($B_\ell, 1{\times}n_c$) \\
\texttt{ln\_f + head} & \texttt{LN+Lin(128, 113, F)} & $[113, 128]{+}256$ & $14{,}720$ & $\mathbf{1},\mathbf{0} / \mathcal{N}$ & Trans.\ ($U, 1{\times}n_c$) \\
\texttt{sensory\_stem ($E_\phi$)} & \texttt{2xConv+Pool+Lin} & Trunk & $125{,}824$ & Kaiming & Trans.\ ($E_\phi, 1{\times}n_c$) \\
\midrule
\multicolumn{6}{@{}p{\textwidth}@{}}{\textbf{$\mathbb{Z}_{113}$ Totals ($d{=}128$):} $\mathbf{227{,}712}$ (1L) $\mid$ $\mathbf{425{,}472}$ (2L) $\mid$ $\mathbf{623{,}232}$ (3L: $619{,}520$ 2D + $3{,}712$ 1D)} \\
\multicolumn{6}{@{}p{\textwidth}@{}}{\textbf{$S_5$ Iso-DoF Totals ($V{=}120$):} $\mathbf{229{,}248}$ (1L) $\mid$ $\mathbf{427{,}008}$ (2L $4{\times}32$ \& $2{\times}64$, $d_{\rm model}{=}128, d_{\rm mlp}{=}512$) $\mid$ $\mathbf{953{,}344}$ (2L $4{\times}32/2{\times}64$, $d_{\rm model}{=}128, d_{\rm mlp}{=}1{,}536$) \& $\mathbf{953{,}028}$ (2L $8{\times}32$, $d_{\rm model}{=}256, d_{\rm mlp}{=}354$, $0.033\%$-matched) $\mid$ $\mathbf{1{,}640{,}448}$ (2L $8{\times}32, d_{\rm model}{=}256, d_{\rm mlp}{=}1{,}024$)} \\
\multicolumn{6}{@{}p{\textwidth}@{}}{\textbf{Real-Sensory MNIST Total ($V{=}10$):} $\mathbf{523{,}136}$ ($E_\phi$: $125{,}824$ + $\texttt{pos\_emb}$: $256$ + 2L Transformer $B$ ($d{=}128, 4{\times}32, d_{\rm mlp}{=}512$): $395{,}520$ + $\mathrm{LN}_f$: $256$ + $U$: $1{,}280$)} \\
\bottomrule
\end{tabular*}
\end{table*}

\subsection{Architectural Census Across Depths, Iso-DoF \texorpdfstring{$S_5$}{S5}, and Real-Sensory MNIST}
At base width ($d_{\rm model}=128, d_{\rm mlp}=512$) with GELU activations \citep{Hendrycks2016_GELU}, each block $B_\ell$ has $197{,}760$ parameters. On $\mathbb{Z}_{113}$ ($V=115$), total parameters equal $\mathbf{227{,}712}$ at 1L, $\mathbf{425{,}472}$ at 2L, and $\mathbf{623{,}232}$ at 3L; on $S_5$ ($V=120$), totals equal $\mathbf{229{,}248}$ at 1L and $\mathbf{427{,}008}$ at 2L across both $4{\times}32$ and $2{\times}64$. In Battery 11, three $0.033\%$-matched $953\text{k}$ 2-layer $S_5$ cells pair $\mathbf{953{,}344}$ parameters on $4{\times}32$ and $2{\times}64$ ($d_{\rm model}=128, d_{\rm mlp}=1{,}536$) with $\mathbf{953{,}028}$ parameters on $8{\times}32$ ($d_{\rm model}=256, d_{\rm mlp}=354$; $-316$ parameters due to integer $d_{\rm mlp}$). These bridge $427\text{k}$ to the $\mathbf{1{,}640{,}448}$-parameter $8{\times}32$ ($d_{\rm mlp}=1{,}024$) model, whose $T=2$ donors qualify in $\sFiveDonorDTwoFiveSixCausalSteps$ steps versus $\sFiveDonorDOneTwoEightCausalSteps$ causal and $\sFiveDonorDOneTwoEightBidirSteps$ bidirectional steps at $d=128$. In Battery 13, the $\mathbf{523{,}136}$-parameter \texttt{SensoryGrokTransformer} encodes each $28\times 28$ image via weight-shared $E_\phi$ ($125{,}824$ params): two $5\times 5$ stride-2 \texttt{Conv2d}--GELU stages ($1\to 32\to 64$), $2\times 2$ average pooling, and bias-free linear projection $576\to 128$. The two slot embeddings plus $256$-parameter $W_{\rm pos}$ pass through two pre-LN Transformer blocks ($395{,}520$ params), terminal $\mathrm{LN}_f$ ($256$ params), and readout $U \in \mathbb{R}^{10\times 128}$ ($1{,}280$ params). Under $1.0\times n_c$ norm matching, $E_{\rm only}$ transfers $125{,}824$ parameters ($E_\phi$), $E{+}U$ transfers $127{,}104$ ($E_\phi, U$), $U{+}B$ transfers $397{,}312$ ($W_{\rm pos}, B, \mathrm{LN}_f, U$), and Full transfers all $523{,}136$.

\end{document}